\documentclass[10pt,twocolumn,letterpaper]{article}

\usepackage[pagenumbers]{cvpr} % To force page numbers, e.g. for an arXiv version
\usepackage{enumitem}
\definecolor{cvprblue}{rgb}{0.21,0.49,0.74}
\usepackage[pagebackref,breaklinks,colorlinks,allcolors=cvprblue]{hyperref}
\usepackage[table]{xcolor}
\def\confName{CVPR}
\def\confYear{2027}

\title{Gradient-Aware Optimal Transport for Flow Matching}

\author{Eungyeol Han \qquad Hoseong Tae \qquad Jong-Seok Lee\thanks{Corresponding author.}\\[0.3em]
School of Integrated Technology, Yonsei University\\
{\tt\small \{eungyeol,hoseong.tae,jong-seok.lee\}@yonsei.ac.kr}
}

\makeatletter
\appto\@maketitle{\noindent
\begin{minipage}{\textwidth}
    \centering
    \captionsetup{type=figure, font=small}
    \begin{subfigure}[t]{0.48\linewidth}
        \centering
        \includegraphics[width=\linewidth,height=0.70\linewidth,keepaspectratio,
            trim=8bp 33bp 10bp 0bp,clip]{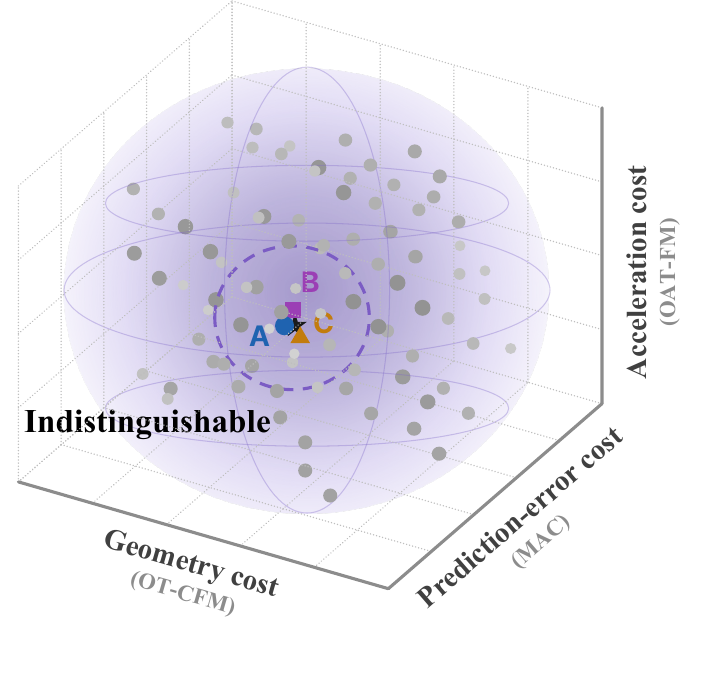}
        \caption{Coupling cost space}
        \label{fig:problem_a}
    \end{subfigure}
    \hfill
    \begin{subfigure}[t]{0.48\linewidth}
        \centering
        \includegraphics[width=\linewidth,height=0.70\linewidth,keepaspectratio,
            trim=0bp 5bp 7bp 0bp,clip]{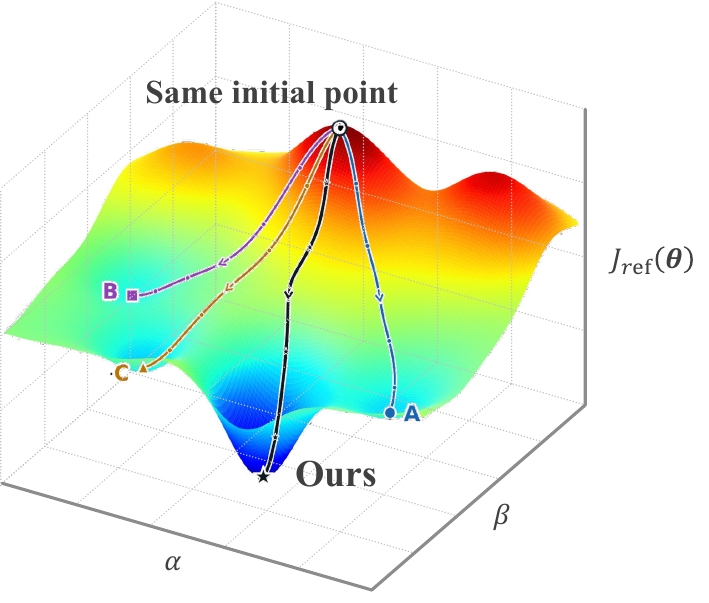}
        \caption{Parameter-update landscape}
        \label{fig:problem_b}
    \end{subfigure}
    \par\vspace{0.4em}
    \caption{\textbf{Similar Costs, Different Learning Utility.} (a) Different couplings can have similar costs under existing criteria. (b) From the same initial model, these couplings can induce different parameter updates and changes in a shared reference loss.}
    \label{fig:problem}
\end{minipage}
\par\vspace{1em}
}
\makeatother

\begin{document}
\renewcommand{\thefootnote}{\fnsymbol{footnote}}
\setcounter{footnote}{1} % Use a dagger for the corresponding-author note.
\maketitle

\begin{abstract}
Flow Matching (FM) trains on source-target couplings, which define its training paths and velocity targets. Independent coupling can lead to curved trajectories, while minibatch Optimal Transport (OT) encourages straighter flows by minimizing squared Euclidean transport costs. Recent methods use the model's prediction error to guide coupling selection. However, neither criterion directly assesses how a coupling contributes to model learning. We present counterexamples in which two couplings have identical marginals, geometric costs, and prediction errors at every flow time, yet lead to different learning effects. To address this, we propose \textbf{GOT-FM}, a gradient-aware coupling framework whose transport cost measures how changing the coupling affects a FM loss after an optimizer update. To make this cost tractable, we differentiate the post-update loss with respect to the coupling, which yields one cost per source--target pair, and use it to rank a finite pool of feasible couplings. We experimentally demonstrate that GOT-FM consistently improves image quality in few-step generation.
\end{abstract} 
\vspace{-2.5em}
\section{Introduction}
\label{sec:intro}

Flow-based models have recently achieved remarkable progress in image synthesis \cite{lipman2022flow, labs2025flux, liu2022flow, ma2024sit}. Normalizing flows \cite{rezende2015variational, papamakarios2021normalizing} generate data by transforming samples from a simple source distribution, such as Gaussian noise, into the target data distribution through invertible mappings. Continuous Normalizing Flows (CNFs) \cite{chen2018neural, grathwohl2018ffjord} extend this idea to continuous time by transporting samples along a learned time-dependent velocity field.
However, likelihood-based CNF training is computationally expensive because it requires solving the ordinary differential equations (ODEs) during training. Flow Matching (FM) \cite{lipman2022flow, liu2022flow, albergo2022building} avoids this cost by regressing the velocity field toward conditional target velocities along predefined paths, so that no ODE simulation is needed during training. In the commonly used linear-path setting, a source sample $\mathbf{z}$ and a target sample $\mathbf{y}$ define the path $\mathbf{x}_t=(1-t)\mathbf{z}+t\mathbf{y}$, and the model is trained to predict its velocity $\mathbf{u}=\mathbf{y}-\mathbf{z}$ at each flow time $t$. Therefore, the way source and target samples are paired, which we refer to as the coupling, determines the training paths and velocity targets that the model learns from.

The coupling also plays a key role in an important goal of FM-based models, generating high-quality samples with only a few sampling steps \cite{geng2026mean, liu2023instaflow}. A common way to reach this goal is to learn straighter trajectories, since a straight trajectory can be integrated accurately with large step sizes \cite{lee2023minimizing, kim2025simple}. However, vanilla FM pairs source and target samples randomly, so paths from different pairs often cross, which makes its trajectories curved \cite{lee2023minimizing}. Two main lines of work address this problem. The first line straightens the trajectories of a trained model. Rectified Flow \cite{liu2022flow} retrains the model on source--target pairs generated by the model itself, making the trajectories straighter with each round, but this requires multiple rounds of costly retraining. The second line improves the coupling itself, so that the model learns straighter paths from the start. Optimal Transport (OT)-based FM methods \cite{pooladian2023multisample, tong2023improving} pair samples within each mini-batch by minimizing the squared Euclidean transport cost. The resulting pairs induce shorter distances and cross less often, encouraging straighter learned flows. Our work follows this second line.

%An important goal of FM-based models is to generate high-quality samples with only a few sampling steps \cite{salimans2022progressive, song2023consistency, geng2026mean, liu2023instaflow}. Much prior work has focused on learning straighter trajectories since straight trajectories can be integrated accurately with large step sizes. However, the vanilla FM pairs source and target samples at random. Paths from different pairs frequently cross one another, an the model learns to average their conflicting velocities, resulting in curved trajectories. Rectified Flow \cite{liu2022flow} addresses this problem by repeatedly retraining the model on source-target pairs generated by the model itself. Another work improves the coupling directly by adopting Optimal Transport (OT). OT-based FM methods \cite{pooladian2023multisample, tong2023improving} pair samples within each mini-batch by minimizing the squared Euclidean transport cost, which reduces path crossings, and corresponds to the minimum-energy transport between the two distributions \cite{benamou2000computational}.

Although OT-based couplings reduce path crossings, they are determined only by the geometric distance. More recent studies have further improved coupling design by considering additional information \cite{meral2026aligning, lee2026geometry, pierret2026tessellations, han2026beyond, yue2025oat, chiang2026cot, lin2026beyond,davtyan2025faster,kong2026alignflow}. For example, OAT-FM \cite{yue2025oat} incorporates velocity information into transport and MAC \cite{lin2026beyond} uses the model's velocity prediction error to evaluate candidate pairs. These methods show that coupling design can benefit from information beyond geometric distance alone.

However, these coupling methods evaluate a coupling \emph{before} the model learns from it, using properties such as geometric distance \cite{tong2023improving, pooladian2023multisample} or prediction error \cite{lin2026beyond}. What ultimately matters for training is how much the model improves \emph{after} learning from the coupling, which we call its \emph{learning utility}. During training, a source--target pair affects the model only through the gradient it produces. The optimizer turns this gradient into a parameter update, and this update determines how much the model improves. Therefore, the gradient of a pair carries the information needed to estimate its learning utility \cite{koh2017understanding, pruthi2020estimating, xia2024less, zheng2024intriguing}. Unlike prediction error, which measures how well the model currently fits a source--target pair, the gradient also tells us in which direction the update will move the model. As a result, couplings which look similar under existing criteria can lead to different learning outcomes, as shown in Figure \ref{fig:problem}. This observation motivates us to use gradients to identify couplings that are useful for training.

In this work, we propose \textbf{GOT-FM}, a gradient-aware optimal transport framework for Flow Matching. Our key idea is to transform the local learning effect of a coupling into a pairwise transport cost. Specifically, we differentiate the post-update FM loss with respect to the coupling, accounting for both pairwise gradients and the optimizer's response. The resulting cost enables gradient-aware selection among feasible coupling plans while preserving both empirical marginals. In this way, our method changes how samples are paired while retaining the standard FM loss formulation.

Our main contributions are summarized as follows:
\begin{enumerate}[label=\arabic*., leftmargin=1.5em, itemsep=3pt]
    \item We show that couplings with identical geometric costs and prediction errors can yield different learning utility through counterexamples.
    \item We propose \textbf{GOT-FM}, a gradient-aware coupling framework. We derive a gradient-aware transport cost that turns the local effect of coupling changes on a post-update FM objective into pairwise edge costs. The cost separates pair-specific gradients from a sensitivity shared by all pairs, which enables gradient-aware coupling selection under fixed empirical marginals without storing per-pair gradients.
    \item We demonstrate that GOT-FM improves few-step image generation quality over the evaluated FM baselines. Through ablation studies, we further highlight the importance of the optimizer-conditioned learning utility for coupling selection.
    
    %We further show that incorporating the optimizer response improves coupling selection and that gradient-aware selection outperforms random and prediction-error-based alternatives.
    
    %and examine one-step utility prediction, optimizer response, candidate-pool construction, reference objective choice, and sparse coupling selection through ablation studies.
    %Our analysis further confirms that the proposed cost accurately predicts the one-step utility of each coupling, which directly supprots the observation in the first contribution.
\end{enumerate}

\vspace{-3mm}
\section{Related Work}
\label{sec:rework}

\subsection{Optimal Transport in Flow Matching}
Standard FM \cite{albergo2022building, liu2022flow, lipman2022flow} pairs source and target samples independently. Multisample Flow Matching \cite{pooladian2023multisample} and OT-CFM \cite{tong2023improving} instead build minibatch OT couplings that minimize the geometric transport cost, which leads to straighter trajectories. Later work improves how these couplings are constructed. LOOM-CFM \cite{davtyan2025faster} extends the optimization across minibatches, AlignFlow \cite{kong2026alignflow} aligns the noise distribution with data points through semi-discrete OT, and OAT-FM \cite{yue2025oat} incorporates velocity information through acceleration-based transport. These methods define the coupling by properties of the samples, whereas we define it by its effect on model training.

%요약 : Flow Matching의 few-step generation의 성능 향상을 위해서 source-target coupling이 학습에 큰 영향을 준다는 것과, 이 성능을 높이기 위해서 optimal transport가 활발히 사용된다는 것은 기존 연구에서 지속해서 보여왔다.

%The choice of source-target coupling plays an importance role in Flow Matching (FM). Conventional FM methods \cite{albergo2022building, liu2022flow, lipman2022flow} commonly use independent source and target samples to construct training paths. To improve these pairings, Multisample Flow Matching \cite{pooladian2023multisample} and OT-CFM \cite{tong2023improving} introduce Optimal Transport (OT) to reduce geometric transport costs. By constructing more structured couplings, these methods aim to reduce training complexity and obtain straighter generation trajectories. 

%Recent studies have further improved how OT couplings are constructed. LOOM-CFM \cite{davtyan2025faster} extends coupling optimization across minibatches. OAT-FM \cite{yue2025oat} incorporates velocity information through acceleration-based transport, and COT-FM \cite{chiang2026cot} constructs cluster-specific source distributions for local OT. These studies show that coupling design is important for both FM training and generation. Our work builds on this direction but focuses on a different criterion: the learning utility of a coupling under the current model and optimizer state.

\vspace{-0.35em}
\subsection{Model-Aware Coupling for Flow Matching}
Geometrically optimal paths are not necessarily easy for a network to learn. Not-So-Optimal Transport Flows \cite{hui2025not} shows in 3D point cloud generation that straighter flows can be harder to learn. MAC \cite{lin2026beyond} selects and upweights source--target pairs with low prediction errors, while its MAC-full constructs a minibatch coupling by optimizing prediction-error costs. MAC thus judges a pair by how well the current model already fits it, whereas our GOT-FM instead scores feasible coupling changes through their local effect on a post-update FM objective.

\vspace{-0.35em}
\subsection{Gradient-Based Learning Utility}
Gradients are widely used to estimate how useful a training sample is for learning \cite{ren2018learning, killamsetty2021grad, wang2024greats, min2026gist}. If the gradient of a training example aligns with that of a target objective, one update step on it is expected to reduce the target loss. However, gradient alignment alone does not determine how the model changes because adaptive optimizers such as Adam use accumulated gradient statistics to compute parameter updates. Therefore, recent work measures learning utility in the optimizer-induced update space \cite{xia2024less, chang2025scalable, wang2026opus, fan2026optiselect, wang2026filter}. This motivates evaluating training signals based on both their gradients and the optimizer's response.

Unlike existing gradient-based data selection methods, which select or reweight predefined training examples, we evaluate optimizer-aware learning utility over source-target assignments and use it within OT. Therefore, the coupling favors useful pairs while preserving both marginals.

\section{Problem Statements}
\label{sec:problem}

%기존 model-aware information을 추가로 반영하여 coupling cost matrix를 작성하는 경우에, 기존 조건을 만족시키는 pairing/batch 내에서도 gradient의 방향이 달라지는 경우가 많고, 이 것이 실제로 

\subsection{Preliminaries}
\label{sec:preliminaries}
\paragraph{Coupling in Flow Matching.}
We consider FM between a source distribution $\mu$ and a target distribution $\nu$. Let $\{\mathbf{z}_i\}^n_{i=1}$ and $\{{\mathbf{y}_j\}}_{j=1}^{m}$ denote samples from these distributions, with empirical weights $a_i$ and $b_j$, respectively. Each source-target pair $(\mathbf{z}_i, \mathbf{y}_j)$ defines a training path. A coupling $\boldsymbol{\pi}$ specifies how much probability mass is distributed across these pairs.

We consider couplings that preserve the empirical source and target marginals, so that the learned flow still transports $\mu$ to $\nu$ regardless of which pairs are selected:
\begin{equation}
\begin{aligned}
\label{eq:preliminaries_marginal}
\Pi(\mathbf{a},\mathbf{b}) = \{\boldsymbol{\pi}\in \mathbb{R}_+^{n\times m} : \ & \boldsymbol{\pi}\mathbf{1}_m=\mathbf{a}, \ \boldsymbol{\pi}^\top\mathbf{1}_n=\mathbf{b}\}, \\
& \text{where } \textstyle\sum_i a_i=\sum_j b_j=1.
\end{aligned}
\end{equation}

In linear-path FM, a source-target pair defines an interpolation and target velocity at each flow time $t\in[0,1]$:
\begin{equation}
\label{eq:target-velocity}
\mathbf{x}_{ij}^t=(1-t)\mathbf{z}_i+t\mathbf{y}_j, \quad \mathbf{u}_{ij}=\mathbf{y}_j-\mathbf{z}_i.
\end{equation}

The model
$v_{\boldsymbol{\theta}}:\mathbb{R}^d\times[0,1]
\to\mathbb{R}^d$
predicts the target velocity. For each source--target
pair, we define the instantaneous loss at flow time $t$ as
\begin{equation}
\label{eq:instantaneous-loss}
\ell_{ij}(\boldsymbol{\theta},t)
=
\frac{1}{d}
\left\|
v_{\boldsymbol{\theta}}(\mathbf{x}_{ij}^t,t)
-\mathbf{u}_{ij}
\right\|_2^2.
\end{equation}
Averaging over flow time gives the per-pair FM loss,
and the coupling determines the overall training objective:
\begin{equation}
\label{eq:time-averaged-fm-loss}
\begin{aligned}
\ell_{ij}(\boldsymbol{\theta})
=
\mathbb{E}_{t\sim\mathcal{U}[0,1]}
[\ell_{ij}(\boldsymbol{\theta},t)], \quad
\mathcal{L}(\boldsymbol{\theta};\boldsymbol{\pi})
=
\sum_{i,j}\pi_{ij}\ell_{ij}(\boldsymbol{\theta}).
\end{aligned}
\end{equation}
Although all couplings in $\Pi(\mathbf{a},\mathbf{b})$ preserve the same empirical marginals, they can assign different amounts of mass to individual source–target pairs. Consequently, they can produce different training paths and parameter gradients.

\vspace{-1.0em}
\paragraph{Optimal transport couplings.}
OT-based FM methods select the coupling by solving an optimal transport problem within each minibatch \cite{pooladian2023multisample, tong2023improving}. Given a cost matrix $\mathbf{C}\in\mathbb{R}^{n\times m}$, whose entry $C_{ij}$ is the edge cost of pairing $\mathbf{z}_i$ with $\mathbf{y}_j$, the OT coupling is
\begin{equation}
\label{eq:preliminaries_optimal_transport}
\boldsymbol{\pi}^\star\in\operatorname*{arg\,min}_{\boldsymbol{\pi}\in\Pi(\mathbf{a},\mathbf{b})}\ \langle \boldsymbol{\pi}, \mathbf{C}\rangle_F,
\qquad
\langle \boldsymbol{\pi}, \mathbf{C}\rangle_F=\sum_{i,j}\pi_{ij}C_{ij}.
\end{equation}
Standard OT-based FM uses the squared Euclidean cost $C^{\mathrm{geo}}_{ij}=\|\mathbf{z}_i-\mathbf{y}_j\|_2^2$, which pairs nearby samples and reduces path crossings. A model-aware alternative uses the current prediction error of each pair as the edge cost, $C^{\mathrm{pred}}_{ij}(\boldsymbol{\theta})=\ell_{ij}(\boldsymbol{\theta})$, so that $\langle \boldsymbol{\pi}, \mathbf{C}^{\mathrm{pred}}(\boldsymbol{\theta})\rangle_F=\mathcal{L}(\boldsymbol{\theta};\boldsymbol{\pi})$. %With uniform weights and $n=m$, as in minibatch training, an optimal coupling can always be chosen as a permutation matrix scaled by $1/n$ \cite{gabriel2019computational}. Solving the OT problem therefore amounts to finding the best one-to-one matching between source and target samples.

%\paragraph{Coupling criteria as edge costs.}
%Different coupling criteria correspond to different cost matrices in this OT problem. The geometric criterion uses $C^{\mathrm{geo}}$, which gives $C_{\mathrm{geo}}(\pi)=\langle \pi, C^{\mathrm{geo}}\rangle_F$. A prediction-error criterion instead uses the current per-pair FM loss, $C^{\mathrm{pred}}_{ij}(\boldsymbol{\theta})=\ell_{ij}(\boldsymbol{\theta})$, which gives $C_{\mathrm{pred}}(\pi;\boldsymbol{\theta})=\langle \pi, C^{\mathrm{pred}}(\boldsymbol{\theta})\rangle_F=\mathcal{L}(\boldsymbol{\theta};\pi)$. GOT-FM keeps the same OT problem but replaces the cost matrix with one that reflects the learning utility of each pair (Section~\ref{sec:method}).

\subsection{Counterexample}
\label{sec:counterexample}

We provide a simple example showing that identical
transport costs and prediction errors do not necessarily
lead to the same learning utility. Higher-dimensional examples are provided in Appendix \ref{sec:sup_counterexample}.

\noindent\textbf{Coupling construction.}
Consider two source samples and two target samples, each with probability mass $1/2$:
\begin{equation}
\mathbf{z}_1=(-1,0), \ \mathbf{z}_2=(1,0), \ \mathbf{y}_1=(0,-1), \ \mathbf{y}_2=(0,1).
\end{equation}
We compare two couplings with $\mathbf{a}=\mathbf{b}=(\frac{1}{2},\frac{1}{2})$:
\begin{equation}
\label{eq:counterexample-couplings}
\boldsymbol{\pi}_A=\frac12
\begin{pmatrix}
1&0\\
0&1
\end{pmatrix},
\qquad
\boldsymbol{\pi}_B=\frac12
\begin{pmatrix}
0&1\\
1&0
\end{pmatrix}.
\end{equation}

Every source--target pair has the \textbf{same squared Euclidean distance}:
\begin{equation}
\|\mathbf{z}_i-\mathbf{y}_j\|_2^2=2 \ \text{for all} \ (i,j).
\end{equation}
Thus, the two couplings have the same geometric transport cost $ \langle \boldsymbol{\pi}_A,\mathbf{C}^\text{geo}\rangle_F=  \langle \boldsymbol{\pi}_B, \mathbf{C}^\text{geo}\rangle_F=2$.

\noindent\textbf{Identical prediction errors.}
Consider a two-layer linear velocity model
$v_{\boldsymbol{\theta}}(\mathbf{x},t)
=\mathbf{W}_2\mathbf{W}_1\mathbf{x}$. Let
\begin{equation}
\label{eq:counterexample-model}
\mathbf{W}_{1,0}=
\begin{pmatrix}
1&1\\
0&1
\end{pmatrix},
\qquad
\mathbf{W}_{2,0}=
\begin{pmatrix}
-1&1\\
0&2
\end{pmatrix}.
\end{equation}
The velocity field is $v_{\boldsymbol{\theta}_0}(\mathbf{x}_{ij}^t,t)=(-(1-t)z_{i,1},\,2t y_{j,2}).$

%$v_{\boldsymbol{\theta}_0}(\mathbf{x},t)=(-x_1,2x_2)$. 

For every source--target pair, the current FM
prediction error at flow time $t$ is
\begin{equation}
\label{eq:counterexample-error}
\ell_{ij}(\boldsymbol{\theta}_0,t)
=
\frac12\left[t^2+(2t-1)^2\right].
\end{equation}
Consequently, the two couplings have \textbf{identical
prediction errors} at every flow time $t\in[0,1]$:
\begin{equation}
\mathcal{L}(\boldsymbol{\theta}_0;\boldsymbol{\pi}_A)=\mathcal{L}(\boldsymbol{\theta}_0;\boldsymbol{\pi}_B)=\frac13
\end{equation}

\noindent\textbf{Different learning utility.}
Despite having identical current losses, the two
couplings induce different parameter gradients.
To compare their learning effects, we use the same
standard FM objective:
\begin{equation}
\label{eq:counterexample-reference}
J_{\mathrm{IID}}(\boldsymbol{\theta})
=
\frac14\sum_{i,j}
\int_0^1
\ell_{ij}(\boldsymbol{\theta},t)\,dt.
\end{equation}
This objective evaluates the model updated with $\boldsymbol{\pi}_A$ and $\boldsymbol{\pi}_B$ under the same source--target distribution.

Starting from the same model parameters, we apply
one SGD update with learning rate $\eta=0.1$: $\boldsymbol{\theta}^+(\boldsymbol{\pi})=\boldsymbol{\theta}_0-\eta \nabla_{\boldsymbol{\theta}}\mathcal{L}(\boldsymbol{\theta};\boldsymbol{\pi})|_{\boldsymbol{\theta}=\boldsymbol{\theta}_0}$.
The resulting decreases in the common FM loss $U_\text{IID}(\boldsymbol{\pi})=J_\text{IID}(\boldsymbol{\theta}_0)-J_\text{IID}(\boldsymbol{\theta}^+(\boldsymbol{\pi}))$ are
\begin{equation}
\label{eq:counterexample-utility}
\begin{aligned}
U_{\mathrm{IID}}(\boldsymbol{\pi}_A)&=0.023495,\\
U_{\mathrm{IID}}(\boldsymbol{\pi}_B)&=0.011782.
\end{aligned}
\end{equation}
Thus, although both couplings have identical transport costs and current prediction errors, they produce different learning utility under the same FM objective. This example shows that these existing coupling criteria do not fully characterize the effect of a coupling on model learning. Detailed computation is provided in Appendix \ref{sec:sup_counterexample}.

This motivates selecting couplings based on their effects on learning after an optimizer update. However, different couplings assign different weights to source--target pairs, making their training losses unsuitable for comparing learning utility directly. A common reference objective allows such comparisons, but evaluating every candidate through a separate optimizer update is computationally expensive. In the next section, we introduce \textbf{GOT-FM}, which addresses both issues while preserving the empirical marginals by deriving a pairwise transport cost from the post-update loss.

\vspace{-1.0em}
\begin{figure*}[t]
  \centering
  \includegraphics[width=\textwidth]{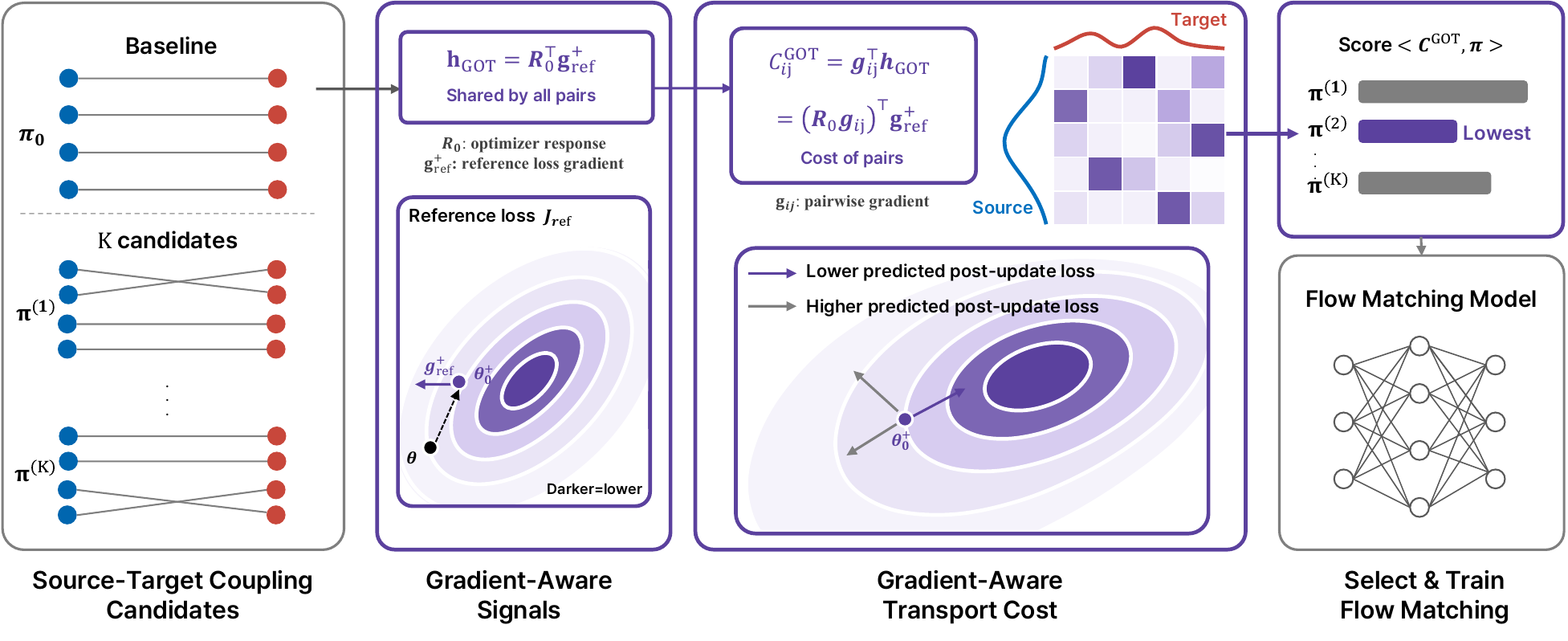}
  \caption{\textbf{Overview of GOT-FM.}
  We select the source--target coupling for Flow Matching (FM) by its estimated effect on learning, while keeping the source and target marginals fixed.
  \textbf{Source--Target Coupling Candidates.} Samples from the source and target distributions form candidate pairs, and each pair defines a training path and a velocity target.
  \textbf{Gradient-Aware Signals.} The optimizer response $\mathbf{R}_0$ and the reference-loss gradient $\mathbf{g}_{\mathrm{ref}}^+$ define a sensitivity shared by all pairs: $\mathbf{h}_{\mathrm{GOT}}=\mathbf{R}_0^\top\mathbf{g}_{\mathrm{ref}}^+$.
  \textbf{Gradient-Aware Transport Cost.} We compute edge costs through directional derivatives of the pairwise FM losses along $\mathbf{h}_{\mathrm{GOT}}$, without explicitly forming per-pair parameter-gradient vectors. Darker cells indicate lower cost. We then score the feasible candidate couplings and select the one with the lowest linearized cost.
  \textbf{Train Flow Matching.} We select the coupling given by the transport plan and use it to train the FM model.}
  \vspace{-0.5em}
  \label{fig:main}
\end{figure*}

\section{Method}
\label{sec:method}
\vspace{-0.5em}

We propose \textbf{GOT-FM}, a gradient-aware coupling framework that selects source--target pairs by their effect on learning, as shown in Figure \ref{fig:main}. Ideally, we would train on the coupling whose optimizer update most reduces a common FM loss. However, this goal raises three difficulties. First, each coupling defines its own training loss, so these losses cannot be compared directly. Second, the post-update loss depends on the whole coupling through the aggregate gradient and the optimizer, so it does not decompose into a weighted sum of fixed pairwise costs. Third, any change to the coupling must preserve both empirical marginals, so pairs cannot be chosen one by one.

We address these difficulties as follows:
\begin{enumerate}[label=\arabic*., leftmargin=1.5em, itemsep=3pt]
    \item We compare all candidates on a shared reference objective in Section \ref{sec:method_reference}.
    \item We derive a pairwise transport cost by differentiating the post-update reference loss with respect to the coupling in Section \ref{sec:method_got}.
    \item We efficiently compute the costs using a shared optimizer-conditioned sensitivity and rank feasible couplings while preserving both empirical marginals in Sections \ref{sec:method_selection} and \ref{sec:method-coupling-selection}.
\end{enumerate}
%We propose \textbf{GOT-FM}, a gradient-aware coupling framework for Flow Matching. Our goal is to select source-target couplings based on how they affect model learning after an optimizer update. %Evaㅕluating this effect for every feasible coupling is computationally expensive, therefore we construct a transport cost from the local effect of coupling changes on a shared reference objective.

%Directly evaluating this effect for every feasible coupling is impractical. Therefore, we construct a transport cost from the local effect of coupling changes on a shared reference objective.

%Our method has three main components: (i) a common objective for comparing learning progress in Section \ref{sec:method_reference}, (ii) an optimizer-conditioned transport cost in Section \ref{sec:method_got}, and (iii) a practical procedure for selecting feasible couplings in Sections \ref{sec:method_selection} and \ref{sec:method-coupling-selection}. 

\subsection{Objective and Shared Reference}
\label{sec:method_reference} 
\textbf{A common measure of learning progress.} Standard OT selects a coupling by minimizing a predefined edge cost. In contrast, GOT-FM seeks a coupling whose optimizer update improves model learning. Since different couplings assign different weights to source--target pairs, their post-update training losses are not directly comparable. Therefore, we evaluate all candidate updates using a shared FM reference objective, denoted by $J_\text{ref}$.

%Different couplings define different training distributions over source-target pairs. Therefore, comparing their current training losses does not directly tell us which coupling produces a more useful model update. Instead, we evaluate the updated model using a common FM reference objective, denoted by $J_\text{ref}$.

To construct $J_\text{ref}$, we consider two coupling distributions: independent coupling (IID), commonly used in standard FM training, and Euclidean OT, which favors shorter transport paths. We combine their FM risks with equal weights to evaluate learning progress without relying on either coupling criterion alone:
\begin{equation}
\label{eq:meta-objective}
J_{\mathrm{ref}}(\boldsymbol{\theta})
=
\frac{1}{2}J_{\mathrm{IID}}(\boldsymbol{\theta})
+
\frac{1}{2}J_{\mathrm{OT}}(\boldsymbol{\theta}).
\end{equation}
$J_\text{IID}$ uses independently coupled source and target samples. $J_\text{OT}$ uses Euclidean OT couplings constructed on reference batches of size $B_\text{ref}$. %In practice, we estimate $J_{\mathrm{ref}}$ using reference samples and a specified flow-time evaluation rule. We denote this finite evaluation by $\widehat J_{\mathrm{ref}}$ and hold its samples and evaluation conditions fixed within each coupling-selection event. 
We compare this choice with alternative reference
objectives in Section \ref{sec:exp_ablation}, with further
details and theoretical analysis provided in Appendices \ref{sec:sup_ablation} and \ref{sec:sup_objective}, respectively.

\vspace{4.0pt}
\noindent\textbf{Learning objective.} Let $\mathbf{g}_{\boldsymbol{\pi}}$ be the FM training gradient induced by coupling $\boldsymbol{\pi}$. We define $\mathcal {A}_s(\boldsymbol{\theta},\mathbf {g})$ as the parameter-update map of the training optimizer. Given the current model parameters $\boldsymbol{\theta}$ and pre-update optimizer state $s$, we write the updated parameters as
\begin{equation}
\label{eq:candidate-update}
\boldsymbol{\theta}^{+}(\boldsymbol{\pi})
=
\mathcal{A}_s(\boldsymbol{\theta},\mathbf{g}_{\boldsymbol{\pi}}).
\end{equation}
We evaluate this update through 
\begin{equation}
\label{eq:glac-post-update-loss}
F_s(\boldsymbol{\pi})
=
J_{\mathrm{ref}}
\left(\boldsymbol{\theta}^{+}(\boldsymbol{\pi})\right).
\end{equation}
The corresponding learning utility is
\begin{equation}
\label{eq:learning-utility}
U_s(\boldsymbol{\pi})
=
J_{\mathrm{ref}}(\boldsymbol{\theta})
-
F_s(\boldsymbol{\pi}).
\end{equation}
Therefore, a lower post-update loss $F_s(\boldsymbol{\pi})$ means a larger learning utility. %Our ideal selection objective is to minimize $F_s(\boldsymbol{\pi})$ over the feasible coupling set $\mathcal{K} \subseteq \Pi(\mathbf{a}, \mathbf{b})$, while keeping the starting model, optimizer state, and reference fixed. 

%However, the optimizer update and reference loss generally depend nonlinearly on the coupling. Evaluating the post-update loss of each candidate separately would require $K$ virtual optimizer updates and reference-loss evaluations. To address both challenges, we linearize the post-update objective around a baseline coupling, transforming its local learning effect into a pairwise transport cost.

%Reference evaluation. We estimate the reference risks using sampled reference batches and a specified flow-time evaluation rule. The reference samples, time points, and scoring randomness are held fixed within a coupling-selection event. All derivatives below refer to the objective evaluated under these fixed conditions, rather than an exactly computed population risk. The META reference is used to construct the selection cost. Separate AUDIT evaluations measure the learning effect of the selected coupling and are not used to select it. The reference construction, OT batch size, flow-time evaluation, loss normalization, and META/AUDIT sampling roles must therefore be specified alongside the experimental settings.

\subsection{Gradient-Aware Transport Cost}
\label{sec:method_got}
\noindent\textbf{Pairwise Gradient.} For each source--target pair $(\mathbf{z}_i,\mathbf{y}_j)$, we define its FM training gradient as
\begin{equation}
\mathbf{g}_{ij}
=
\nabla_{\boldsymbol{\theta}}
\ell_{ij}(\boldsymbol{\theta}).
\end{equation}
The training gradient induced by a coupling $\boldsymbol{\pi}$ is then
\begin{equation}
\mathbf{g}_{\boldsymbol{\pi}}
=
\sum_{i,j}\pi_{ij}\mathbf{g}_{ij}.
\end{equation}

Although $\mathbf{g}_{\boldsymbol{\pi}}$ is linear in the coupling, the post-update objective $F_s(\boldsymbol{\pi})$ in Eq.~\eqref{eq:glac-post-update-loss} is generally nonlinear. Therefore, we construct a local approximation around the Euclidean OT coupling $\boldsymbol{\pi}_0$ of the current minibatch. The corresponding virtual update is
\begin{equation}
\mathbf{g}_0=\mathbf{g}_{\boldsymbol{\pi}_0},
\qquad
\boldsymbol{\theta}_0^+
=
\mathcal{A}_s(\boldsymbol{\theta},\mathbf{g}_0).
\end{equation}
This virtual update does not modify the actual training state.

\noindent\textbf{Optimizer Response.} Rather than relying on raw gradient alignment, we account for how the optimizer transforms training gradients into parameter updates. We compute the optimizer's sensitivity to the training gradient:
\begin{equation}
\label{eq:optimizer_response}
\mathbf{R}_0
=
\left.
D_{\mathbf{g}}
\mathcal{A}_s(\boldsymbol{\theta},\mathbf{g})
\right|_{\mathbf{g}=\mathbf{g}_0}.
\end{equation}
We then compute the gradient of the reference loss at the virtually updated parameters:
\begin{equation}
\mathbf{g}_{\mathrm{ref}}^+
=
\left.
\nabla_{\boldsymbol{\vartheta}}
J_{\mathrm{ref}}(\boldsymbol{\vartheta})
\right|_{\boldsymbol{\vartheta}=
\boldsymbol{\theta}_0^+}.
\end{equation}

\noindent\textbf{Gradient-Aware Edge Cost.} Combining the optimizer response with the
pairwise and reference gradients, the chain rule yields our transport cost:
\begin{equation}
\label{eq:got_cost}
C_{ij}^{\mathrm{GOT}}
=
\mathbf{g}_{ij}^{\top}
\mathbf{R}_0^{\top}
\mathbf{g}_{\mathrm{ref}}^+
=
\left.
\frac{\partial F_s(\boldsymbol{\pi})}
{\partial \pi_{ij}}
\right|_{\boldsymbol{\pi}=\boldsymbol{\pi}_0}.
\end{equation}
For finite coupling changes, we use a first-order approximation of the post-update reference loss around $\boldsymbol{\pi}_0$:
\begin{equation}
F_s(\boldsymbol{\pi})
\approx
F_s(\boldsymbol{\pi}_0)
+
\sum_{i,j}
C_{ij}^{\mathrm{GOT}}
(\pi_{ij}-\pi_{0,ij}).
\end{equation}
Equivalently, the predicted utility improvement over the baseline coupling is
\begin{equation}
\label{eq:predicted-utility}
U_s(\boldsymbol{\pi})-U_s(\boldsymbol{\pi}_0)
\approx
-\sum_{i,j}
C_{ij}^{\mathrm{GOT}}
(\pi_{ij}-\pi_{0,ij}).
\end{equation}
This expresses the local learning effect as a pairwise transport cost for ranking
feasible coupling plans.

\noindent\textbf{Shared Sensitivity for Edge Scoring.} A direct implementation of the gradient-aware edge cost in Eq.~\eqref{eq:got_cost} would require computing a full parameter gradient for every scored pair, which is computationally expensive. We avoid this by exploiting a shared direction in Eq.~\eqref{eq:got_shared}. Within a selection event, the optimizer response and the reference gradient are the same for all pairs, so they form one shared direction:
\begin{equation}
\label{eq:got_shared}
\mathbf{h}_{\mathrm{GOT}}
=
\mathbf{R}_0^\top\mathbf{g}_{\mathrm{ref}}^+,
\qquad
C_{ij}^{\mathrm{GOT}}
=
D_{\boldsymbol{\theta}}
\ell_{ij}(\boldsymbol{\theta})
[\mathbf{h}_{\mathrm{GOT}}].
\end{equation}
We compute $\mathbf{h}_{\mathrm{GOT}}$ once using
a vector--Jacobian product through the optimizer,
then obtain the required edge costs through
pair-loss directional derivatives.
This avoids storing a full parameter-gradient
vector for every scored pair. Further theoretical analysis, including the local derivative, approximation error, and optimizer response is provided in Appendix \ref{sec:sup_theory}.

\subsection{Coupling Candidate Construction}
\label{sec:method_selection}

We construct candidate couplings within $\Pi(\mathbf{a},\mathbf{b})$, defined in Eq.~\eqref{eq:preliminaries_marginal}, ensuring that both empirical source and target marginals are preserved.

%We construct candidate couplings that preserve the empirical source and target marginals. Given source
%probabilities $\mathbf{a}$ and target probabilities $\mathbf{b}$, the set of valid couplings is
%\begin{equation}
%\label{eq:marginal-constraints}
%\Pi(\mathbf{a},\mathbf{b})
%=
%\left\{
%\boldsymbol{\pi}\geq 0:
%\boldsymbol{\pi}\mathbf{1}_m=\mathbf{a},
%\;
%\boldsymbol{\pi}^\top\mathbf{1}_n=\mathbf{b}
%\right\}.
%\end{equation}

A low cost on one pair does not by itself define a valid coupling. Moving mass onto one pair must be balanced by changes on other pairs, so that both empirical marginals in Eq.~\eqref{eq:preliminaries_marginal} are preserved. For this reason, we cannot simply keep or re-weight favorable pairs, as data selection methods do. Instead, we construct complete feasible couplings and rank them with $\mathbf{C}^\text{GOT}$.

Using the Euclidean OT coupling $\boldsymbol{\pi}_0$ as the baseline, we construct feasible coupling candidates by solving linear programs with (i) geometry/endpoint-based costs, including linear combinations of normalized geometric cost and endpoint prediction error, and (ii) independent Gaussian random-cost matrices. The endpoint error is the mean FM prediction mean squared error (MSE) at $t=0$ and $t=1$. All candidates preserve the empirical source and target marginals. We compare candidate-pool compositions in Section \ref{sec:exp_ablation}. Detailed candidate construction procedures are provided in Appendix \ref{sec:sup_coupling}.

Let $\mathcal{K}\subseteq\Pi(\mathbf{a},\mathbf{b})$ denote the set of couplings satisfying the specified transport-cost and movement constraints. We retain proposals in $\mathcal{K}$.
The resulting candidate pool is
\begin{equation}
\label{eq:candidate-pool}
\mathcal{P}
=
\{\boldsymbol{\pi}^{(1)},\ldots,\boldsymbol{\pi}^{(K)}\}
\subseteq\mathcal{K}.
\end{equation}

%We then select a coupling from $\mathcal{P}$ using the gradient-aware cost, as described in Section~\ref{sec:method-coupling-selection}.

\subsection{Gradient-Aware Coupling Selection}
\label{sec:method-coupling-selection}

Given the candidate set $\mathcal{P}$, we select the
coupling that minimizes the gradient-aware transport cost:
\begin{equation}
\label{eq:pool-selection}
\boldsymbol{\pi}_{\mathrm{sel}}
\in
\operatorname*{arg\,min}_{\boldsymbol{\pi}\in\mathcal{P}}
\sum_{i,j}\pi_{ij}C_{ij}^\text{GOT}.
\end{equation}
As shown in Eq.~\eqref{eq:predicted-utility},
minimizing this cost maximizes the predicted
one-step utility improvement relative to the
baseline coupling $\boldsymbol{\pi}_0$.

\vspace{4.0pt}
\noindent\textbf{Efficient edge scoring.}
To efficiently compare the candidates, we evaluate the
gradient-aware costs only on the union of their
supports and the baseline support:
\begin{equation}
\label{eq:support-union}
\Omega_{\mathrm{score}}
=
\operatorname{supp}(\boldsymbol{\pi}_0)
\cup
\bigcup_{k=1}^{K}
\operatorname{supp}(\boldsymbol{\pi}^{(k)}).
\end{equation}

We compute $\mathbf{h}_\text{GOT}$ once per selection event and score only the edges in $\Omega_\text{score}$ to reduce computation. Rather than solving an additional transport problem over this support, we rank the pre-generated feasible coupling candidates. Appendix \ref{sec:sup_relation_to_direct} compares finite-pool selection with direct optimization over the full feasible coupling set.

\subsection{Training}
\label{sec:method_training}
\textbf{Applying gradient-aware selection at selected steps.} Candidate-restricted scoring reduces the computational cost. To further reduce its frequency, we apply gradient-aware selection only at designated training steps. Additional analysis regarding selection frequency and timing is provided in Appendix \ref{sec:sup_ablation}.

Let $r$ index optimizer updates, and let $\mathcal{T}_\text{sel}$ denote the steps at which selection is performed. At each step, $\boldsymbol{\pi}_{0,r}$ is the Euclidean OT coupling of the current minibatch. The training coupling is

\begin{equation}
\label{eq:sparse-schedule}
\boldsymbol{\pi}_r
=
\begin{cases}
\boldsymbol{\pi}_{\mathrm{sel},r},
& r\in\mathcal{T}_\text{sel}, \\[3pt]
\boldsymbol{\pi}_{0,r},
& r\notin\mathcal{T}_\text{sel}.
\end{cases}
\end{equation}

At a selected step, we construct candidates for the current minibatch and compute the gradient-aware cost using the current model and optimizer state. At other steps, we use the current minibatch’s Euclidean OT coupling. We do not carry a previously selected coupling over to new source and target samples.

\vspace{4.0pt}
\noindent\textbf{Updating the FM model.} Once $\boldsymbol{\pi}_r$ is selected, we train the model with the standard FM objective under that coupling. Let
\begin{equation}
\label{eq:training-gradient}
\widehat{\mathbf{g}}^{\,\mathrm{train}}_{\boldsymbol{\pi}_r}
=
\nabla_{\boldsymbol{\theta}}
\widehat{\mathcal{L}}(\boldsymbol{\theta}_r;\boldsymbol{\pi}_r)
\end{equation}
denote the gradient computed using the training-time samples and randomness. The actual update is
\begin{equation}
\label{eq:actual-update}
\boldsymbol{\theta}_{r+1}
=
\mathcal{A}_{s_r}
\left(
\boldsymbol{\theta}_r,
\widehat{\mathbf{g}}^{\,\mathrm{train}}_{\boldsymbol{\pi}_r}
\right),
\end{equation}
with the optimizer state updated in the usual way.

\section{Experiments}
\label{sec:exp}

\begin{figure*}[t]
  \centering
  \vspace{-3mm}
  \includegraphics[width=\textwidth]{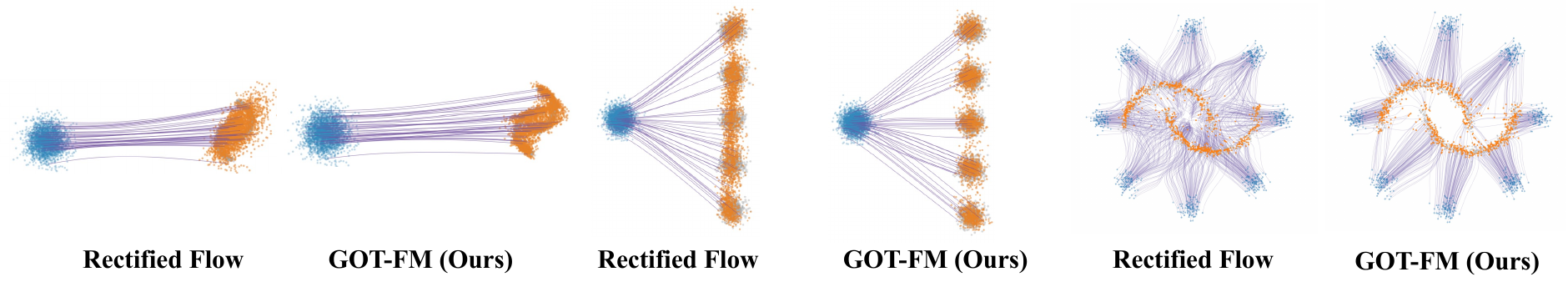}
  \caption{\textbf{Visualization of transport on 2D synthetic distributions.} We compare Rectified Flow \cite{liu2022flow} and GOT-FM (Ours) on Two Moons, a five-component Gaussian Mixture, and eight Gaussians to a two-moons distribution. Blue points denote the source distribution, gray points the target distribution, and orange points the generated samples. Transport lines indicate the corresponding sample movements.}
  \vspace{-2mm}
  \label{fig:toy}
\end{figure*}

We evaluate GOT-FM on unconditional and conditional image generation. We further examine its one-step utility prediction, candidate-pool construction, reference objective, intervention settings, optimizer response and inference efficiency. Additional ablation analyses regarding frequency of gradient-aware selection, timing of gradient-aware selection, and reference batch sizes are provided in Appendix \ref{sec:sup_ablation}.

%We evaluate GOT-FM to answer three questions. First, does GOT-FM improve few-step generation quality on unconditional and conditional image generation tasks? Second, does gradient-aware coupling selection provide benefits over random and prediction-error-based selection under the evaluated sparse-intervention settings? Third, how do the frequency and timing of gradient-aware interventions and the choice of reference objective affect generation performance?

%We evaluate our method to understand whether gradient-aware learning utility provides a useful criterion for coupling selection in Flow Matching. Our experiments focus on three questions. First, does the proposed GOT-FM criterion select couplings with better one-step learning utility than previous criteria? Second, Sparse GOT-FM improve few-step generation while reducing the number of expensive gradient-aware coupling selections? Third, how do the choice of reference objective, the timing of sparse interventions, and other components affect the resulting performance?

\subsection{Datasets and Baselines}
\label{sec:exp_baselines}
We evaluate \textbf{GOT-FM} on three two-dimensional synthetic distributions (Two Moons, a five-component Gaussian mixture, and eight Gaussian modes to a two-moons distribution) to visualize the learned transport behavior. We further conduct unconditional image generation on CIFAR-10 \cite{krizhevsky2009learning} and conditional generation on ImageNet $256 \times 256$ \cite{deng2009imagenet}.
%to evaluate performance in higher-dimensional settings.

We organize our comparisons by FM training framework. On CIFAR-10 \cite{krizhevsky2009learning}, we first compare GOT-FM with the standard independent coupling and existing coupling methods under Rectified Flow \cite{liu2022flow}: OT-CFM \cite{tong2023improving}, LOOM-CFM \cite{davtyan2025faster}, OAT-FM \cite{yue2025oat}, AlignFlow \cite{kong2026alignflow}, and MAC-full \cite{lin2026beyond}. We also apply GOT-FM to MeanFlow \cite{geng2026mean}, which uses a different training objective, and compare it with standard MeanFlow. On ImageNet $256\times256$ \cite{deng2009imagenet}, we evaluate Rectified Flow \cite{liu2022flow} with the SiT \cite{ma2024sit} B/4 backbone, comparing independent coupling, OT-CFM \cite{tong2023improving}, and GOT-FM.

\subsection{Implementation Details}
\label{sec:exp_implementation}
For a fair comparison on CIFAR-10 \cite{krizhevsky2009learning}, we use a TorchCFM U-Net \cite{dhariwal2021diffusion,tong2023improving}, and keep the model initialization, training data, FM objective, optimizer, and number of training updates fixed across coupling variants. Although the original OAT-FM \cite{yue2025oat} paper reports results at NFE=135, we evaluate it at the same NFEs as the other baselines for a consistent comparison. For MAC-full \cite{lin2026beyond}, we evaluate its coupling criterion within our Rectified Flow \cite{liu2022flow} setup, rather than using the Shortcut-based results reported in the original paper. For class-conditional ImageNet $256\times256$ \cite{deng2009imagenet}, we evaluate the methods using the same SiT-B/4 \cite{ma2024sit} backbone. We evaluate generation quality using Fréchet Inception Distance (FID) \cite{heusel2017gans} at different NFEs. Detailed training configurations, checkpoint initialization, baseline implementations, and computational costs are provided in Appendices \ref{sec:sup_implementation} and \ref{sec:sup_computational}. Extensions to MeanFlow \cite{geng2026mean} and conditional image generation are also provided in Appendices \ref{sec:sup_meanflow} and \ref{sec:sup_conditional}, respectively.

%main table
\begin{table}[t]
  \centering
  \vspace{-2mm}
  \caption{\textbf{Unconditional image generation on CIFAR-10~\cite{krizhevsky2009learning}.} We report FID~\cite{heusel2017gans} (lower is better) at different numbers of function evaluations (NFEs). Each indented row applies the method to the base model above it, Rectified Flow~\cite{liu2022flow} or MeanFlow~\cite{geng2026mean}. Since MeanFlow is designed for one-step generation, we report its results only at NFE=1. $^\dagger$ Implementation details for OAT-FM \cite{yue2025oat} and MAC-full \cite{lin2026beyond} are provided in Section \ref{sec:exp_implementation}.}
  \vspace{-0.5em}
  \label{tab:ot_main_cifar}
  \small
  \setlength{\tabcolsep}{2.0pt}
  \renewcommand{\arraystretch}{1.10}
  \begin{tabular*}{\columnwidth}{@{\extracolsep{\fill}}lccccc@{}}
    \toprule
    Method & NFE=1 & NFE=4 & NFE=8 & NFE=16 & NFE=50 \\
    \midrule
    Rectified Flow~\cite{liu2022flow}                & 353.3 & 58.2 & 23.6 & 15.3 & 11.4 \\
    \quad  OT-CFM~\cite{tong2023improving}           & 246.8 & 39.8 & 21.2 & 14.7 & 10.5 \\
    \quad  LOOM-CFM~\cite{davtyan2025faster}         & 280.3 & 36.8 & 20.2 & 14.1 & 10.2 \\
    \quad  OAT-FM$^\dagger$ \cite{yue2025oat}                 & 313.8 & 102.6 & 58.2 & 44.3 & 40.0 \\
    \quad  AlignFlow~\cite{kong2026alignflow}        & 278.2 & 38.2 & 21.7 & 18.4 & 11.3 \\
    \quad  MAC-full$^\dagger$~\cite{lin2026beyond}                  & 319.1 & 42.2 & 20.3 & 16.7 & 11.5 \\
    \quad  \textbf{GOT-FM (Ours)}                    & $\mathbf{242.1}$ & $\mathbf{36.0}$ & $\mathbf{17.6}$ & $\mathbf{12.1}$ & $\mathbf{8.1}$ \\
    \midrule
    MeanFlow~\cite{geng2026mean}                     & 6.6 & - & - & - & - \\
    \quad  \textbf{GOT-FM (Ours)}                    & $\mathbf{6.3}$ & - & - & - & - \\
    \bottomrule
  \end{tabular*}
  \vspace{-0.5em}
\end{table}

%Imagenet
\begin{table}
  \centering
  \caption{\textbf{Class-conditional image generation on ImageNet 256$\times$256~\cite{deng2009imagenet}.} We report FID~\cite{heusel2017gans} (lower is better) at different numbers of function evaluations (NFEs). Each indented row applies the method to the base model, Rectified Flow~\cite{liu2022flow}.}
  \label{tab:imagenet_cond}
  \vspace{-1.0em}
  \small
  \setlength{\tabcolsep}{2.0pt}
  \renewcommand{\arraystretch}{1.10}
  \begin{tabular*}{\columnwidth}{@{\extracolsep{\fill}}lccccc@{}}
    \toprule
    Method & NFE=1 & NFE=4 & NFE=8 & NFE=16 & NFE=50 \\
    \midrule
    Rectified Flow~\cite{liu2022flow}       & 307.5 & 125.5 & 76.3 & 59.0 & 45.2 \\
    \quad OT-CFM~\cite{tong2023improving}   & 312.0 & 122.2 & 71.4 & 55.1 & 48.8 \\
    \quad \textbf{GOT-FM (Ours)}            & $\mathbf{298.3}$ & $\mathbf{119.3}$ & $\mathbf{68.2}$ & $\mathbf{52.1}$ & $\mathbf{42.7}$ \\
    \bottomrule
  \end{tabular*}
  \vspace{-1.7em}
\end{table}

\subsection{Main Experimental Results}
\label{sec:exp_main_results}
\noindent\textbf{2D Synthetic Distributions.} We first visualize the learned transport behavior on three 2D synthetic datasets. These datasets allow us to examine how different coupling strategies transport source samples toward target distributions with different geometric structures. Figure \ref{fig:toy} compares Rectified Flow and GOT-FM on these datasets. The visualization demonstrates that our generated 2D point cloud is much closer to the original data distributions and shows straighter trajectories.

\noindent\textbf{Unconditional Generation on CIFAR-10 \cite{krizhevsky2009learning}.} We evaluate unconditional image generation on CIFAR-10 \cite{krizhevsky2009learning} using different numbers of function evaluations (NFEs). Table \ref{tab:ot_main_cifar} shows that GOT-FM achieves the lowest FID compared to the other FM baselines. These results indicate that gradient-aware coupling selection can improve the generation quality beyond geometric or model-aware coupling alone.

\noindent\textbf{Conditional Generation on ImageNet \cite{deng2009imagenet}.} We further evaluate GOT-FM on class-conditional ImageNet $256\times256$ generation using the SiT-B/4 \cite{ma2024sit} backbone. As shown in Table \ref{tab:imagenet_cond}, GOT-FM achieves lower FID than the other FM baselines across all sampling budgets. These results extend the observed benefits of GOT-FM to conditional image generation with Transformer-based flow models.

%\noindent\textbf{Qualitative Results.} Figure \ref{fig:qualitative} presents generated CIFAR-10 samples from Rectified Flow \cite{liu2022flow}, OT-CFM \cite{tong2023improving}, and GOT-FM. This comparison provides a visual complement to the quantitative FID results in Table \ref{tab:ot_main_cifar} and GOT-FM indicates higher quality image samples.

%\begin{figure}[t]
  %\centering
  %\includegraphics[width=\linewidth]{fig/04_qualitative_ver2.pdf}
  %\caption{\textbf{Qualitative comparison on CIFAR-10 \cite{krizhevsky2009learning}.} The first row shows Rectified Flow \cite{liu2022flow}, the second row OT-CFM \cite{tong2023improving}, the third row GOT-FM (Ours) under 50 sampling steps. GOT-FM improves overall image quality.}
  %\vspace{-3mm}
  %\label{fig:qualitative}
%\end{figure}

\begin{figure}[t]
  \centering
  \begin{subfigure}[t]{0.48\linewidth}
    \centering
    \includegraphics[width=\linewidth]{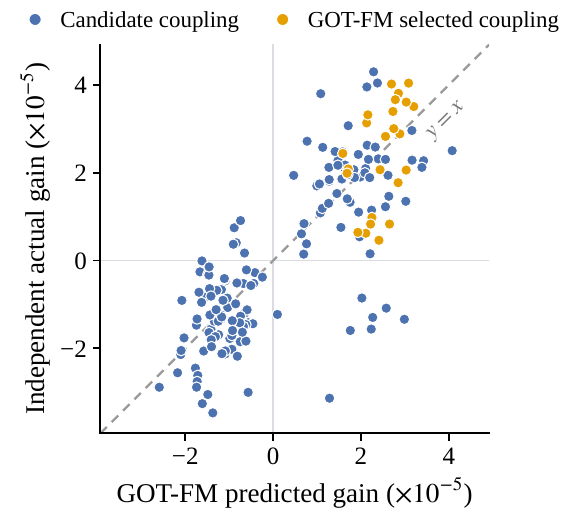}
    \caption{Effect of Learning Utility}
    \label{fig:gotfm_gain_scatter_a}
  \end{subfigure}
  \hfill
  \begin{subfigure}[t]{0.48\linewidth}
    \centering
    \includegraphics[width=\linewidth]{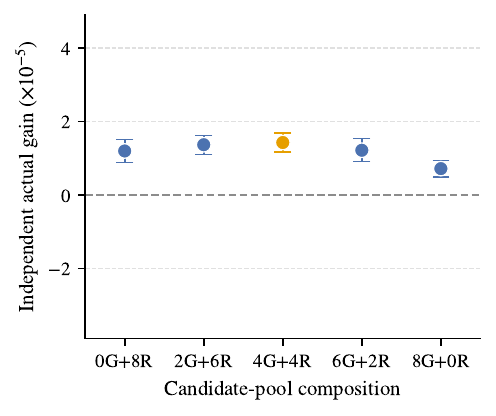}
    \caption{Effect of Candidate Pool}
    \label{fig:gotfm_gain_scatter_b}
  \end{subfigure}

  \caption{\textbf{Effect of One-step Utility and Candidate Pool.} (a) GOT-FM predicted gain vs. independent actual gain. (b) Mean independent actual gain across candidate pools with different ratios of geometry/endpoint (G) to random-cost (R) candidates. Positive gains indicate improvement over the baseline coupling update. Error bars indicate $95\%$ confidence intervals across three seeds.}
  \label{fig:gotfm_gain_scatter}
\end{figure}
\vspace{0.5em}

\subsection{Ablation Analysis}
\label{sec:exp_ablation}
\noindent\textbf{Effect of Learning Utility.} We examine whether GOT-FM's predicted learning utility reflects the actual improvement measured on an independent evaluation set. As shown in Figure \ref{fig:gotfm_gain_scatter_a}, predicted and actual gains are positively correlated (Spearman correlation coefficient=0.828). Notably, all couplings selected by GOT-FM yield positive actual gains (yellow dots). These results suggest that the proposed cost provides a useful signal for coupling selection. Evaluation details are given in Appendix \ref{sec:sup_ablation}.

%We first examine whether the proposed cost accurately predicts one-step learning utility and selects beneficial couplings. Under matched model and optimizer states, we compare the predicted gain with the exact virtual-update gain on the same reference objective $J_\text{ref}$, and separately evaluate the selected updates on an independent set.

\noindent\textbf{Effect of Candidate Pool.} Since GOT-FM selects couplings from a finite candidate pool, its effectiveness may depend on which couplings are available. We vary the ratio of geometry/endpoint (G) and random-cost (R) pools while keeping the candidate pool size and feasibility constraints fixed. As shown in Figure \ref{fig:gotfm_gain_scatter_b}, GOT-FM yields positive mean learning gains across all candidate pools. This result shows that its benefits are not limited to a specific proposal mixture. More detailed comparisons are provided in Appendix \ref{sec:sup_ablation}.

\begin{table}[t]
  \centering
  \caption{\textbf{Effect of reference objective
  $J_{\mathrm{ref}}$.}
  Stability denotes the coupling-selection agreement
  across independent reference draws, reported as mean $\pm$ standard deviation across three seeds. Gain Corr. ($\rho_s$) denotes the Spearman correlation between predicted and independent actual gain.
  We report FID \cite{heusel2017gans} for unconditional CIFAR-10 \cite{krizhevsky2009learning} generation at 4 and 8 NFEs.}
  \vspace{-0.5em}
  \label{tab:ot_ablation_meta}
  \footnotesize
  \setlength{\tabcolsep}{4pt}
  \renewcommand{\arraystretch}{1.10}
  \begin{tabular}{@{}lcccc@{}}
    \toprule
    $J_{\mathrm{ref}}$
    & Stability (\%) $\uparrow$
    & \shortstack{Gain Corr. ($\rho_s$) $\uparrow$}
    & FID$_4$ $\downarrow$
    & FID$_8$ $\downarrow$ \\
    \midrule
    IID
    & $86.48 \pm 8.10$
    & 0.520 & 37.35 & 18.30 \\
    OT
    & $\mathbf{99.22 \pm 0.10}$
    & 0.385 & 38.50 & 19.00 \\
    MAC \cite{lin2026beyond} & $58.3\pm1.27$ & 0.520 & 42.99 & 23.71 \\
    AlignFlow \cite{kong2026alignflow} & $91.7 \pm 0.82 $ & -0.128 & 43.33 & 23.41 \\
    IID/OT (1:3) & $95.30 \pm 0.28$ & 0.548 & 43.20 & 23.37\\
    IID/OT (3:1) & $96.25\pm 1.20$ & 0.553 & 43.71 & 24.02 \\
    \textbf{IID/OT (1:1)}
    & $97.40 \pm 1.47$
    & $\mathbf{0.828}$
    & $\mathbf{36.04}$
    & $\mathbf{17.55}$ \\
    \bottomrule
  \end{tabular}
  \vspace{-1.5em}
\end{table}

\noindent\textbf{Effect of the Reference Objective.} Since the gradient-aware criterion is the main source of improvement, we next examine the reference objective used to compute it. We compare seven reference objectives: independent (IID) coupling, Euclidean OT coupling, MAC \cite{lin2026beyond} coupling, AlignFlow \cite{kong2026alignflow} coupling, and IID/OT mixtures with ratios of 1:3, 3:1 and 1:1. As shown in Table \ref{tab:ot_ablation_meta}, the equal mixture of IID/OT achieves the lowest FID at 4 and 8 NFEs and highest gain-rank correlation. Interestingly, the OT reference gives the highest selection agreement across reference draws, i.e., its selected coupling changes the least, but it does not give the best FID. This suggests that a more consistent selection does not necessarily lead to better generation quality.

\noindent\textbf{Effect of Sparse Coupling Selection Criterion.} The improvement may arise simply from occasionally replacing Euclidean OT with another coupling, regardless of how that coupling is selected. To examine whether the selection criterion itself matters, we keep the intervention frequency, timing, and candidate pool fixed, varying only the selection criterion. Sparse-Random randomly selects a coupling from the candidate pool, while Sparse-MAC uses a MAC-style prediction-error criterion~\cite{lin2026beyond}. All methods use Euclidean OT at non-intervention steps. As shown in Table~\ref{tab:ot_ablation_policy}, GOT-FM achieves the lowest FID across the evaluated NFEs, outperforming both variants and Euclidean OT. These results suggest that gradient-aware coupling selection provides benefits beyond simply replacing Euclidean OT or selecting couplings based on prediction error. Additional analyses of intervention frequency and timing are provided in Appendix~\ref{sec:sup_ablation}.

%The frequency and timing experiments in Appendix \ref{sec:sup_ablation} determine how often and when selection is applied, but not whether the selection criterion itself matters. The gain could also come from simply departing from Euclidean OT at some updates, regardless of which coupling is chosen. To test this, we keep the intervention frequency, timing, and candidate pool the same as GOT-FM and change only the selection criterion. Sparse-Random selects a coupling from the candidate pool at random, and Sparse-MAC selects one using a MAC \cite{lin2026beyond}-style prediction-error criterion. All variants use Euclidean OT at non-intervention steps. As shown in Table \ref{tab:ot_ablation_policy}, GOT-FM achieves the lowest FID scores, outperforming both variants and Euclidean OT. This supports the conclusion that the improvement comes from using gradient information to select couplings, not from simply departing from Euclidean OT or from selecting by prediction error.

\begin{figure}[t]
  \centering
  \begin{subfigure}[b]{0.48\linewidth}
    \centering
    \includegraphics[width=\linewidth]{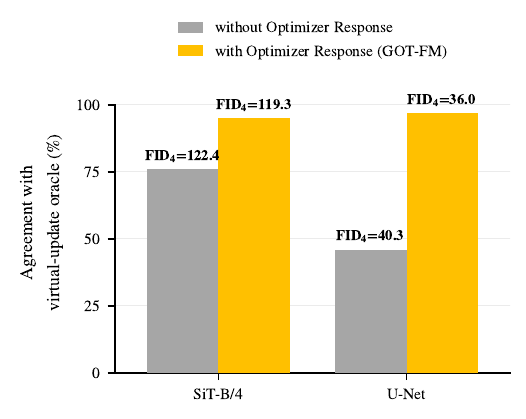}
    \caption{Effect of Optimizer Response}
    \label{fig:optimizer}
  \end{subfigure}\hfill
  \begin{subfigure}[b]{0.48\linewidth}
    \centering
    \includegraphics[width=\linewidth]{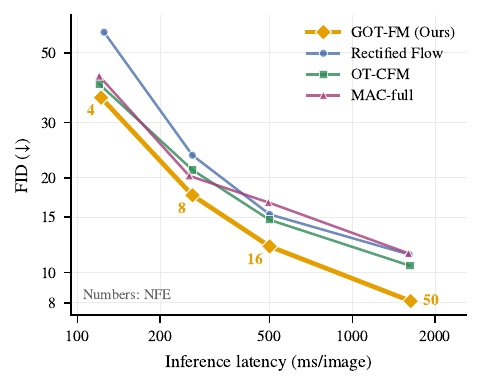}
    \caption{Inference Latency}
    \label{fig:latency}
  \end{subfigure}
  \caption{\textbf{Effect of the optimizer response and inference latency.} (a) Agreement with virtual update oracle and FID at NFEs=4 with and without the optimizer response $\mathbf{R}_0$. (b) FID \cite{heusel2017gans} versus inference latency on CIFAR-10 \cite{krizhevsky2009learning} against Rectified Flow \cite{liu2022flow}, OT-CFM \cite{tong2023improving}, and MAC-full \cite{lin2026beyond} at different NFEs.}
  \vspace{-1.5em}
  \label{fig:r0_ablation}
\end{figure}

\noindent\textbf{Effect of Optimizer Response.} We examine whether incorporating the optimizer response (Eq.~\eqref{eq:optimizer_response}) improves coupling selection compared with raw gradient alignment. We compare GOT-FM with a variant that omits $\mathbf{R}_0$: $C_{ij}^\text{Align}=-\mathbf{g}_{ij}^{\top}\mathbf{g}_{\mathrm{ref}}^+$. We define the virtual-update oracle as the candidate that achieves the lowest $J_{\mathrm{ref}}$ after a separate one-step optimizer update from the same model and optimizer state. Figure \ref{fig:optimizer} reports how often each method selects the same coupling as this oracle within the candidate pool. Including $\mathbf{R}_0$ improves oracle agreement, suggesting that accounting for the optimizer response improves the accuracy of one-step coupling selection. We further evaluate the effect of the optimizer response on generation quality by comparing FID with and without $\mathbf{R}_0$. Incorporating $\mathbf{R}_0$ yields lower FID, suggesting that its benefits extend beyond oracle agreement to downstream generation performance.

%from
%approximately $75\%$ to $95\%$ on SiT \cite{ma2024sit} B/4 and from $45\%$ to $97\%$on U-Net \cite{dhariwal2021diffusion,tong2023improving}.

%accounting for the optimizer response in Eq.~\eqref{eq:optimizer_response} rather than raw gradient alignment improves coupling selection by comparing GOT-FM with a variant that omits $\mathbf{R}_0$: $C_{ij}^\text{Align}=-\mathbf{g}_{ij}^{\top}\mathbf{g}_{\mathrm{ref}}^+$. We define the virtual-update oracle as the candidate that achieves the lowest $J_{\mathrm{ref}}$ after a separate one-step optimizer update from the same
%model and optimizer state. Figure \ref{fig:optimizer} reports how often each method selects the same coupling as this oracle within the candidate pool. Including $\mathbf{R}_0$ improves oracle agreement, indicating that GOT-FM achieves more accurate one-step coupling selection.

%compares agreement with virtual update oracle in Eq.~\eqref{eq:glac-post-update-loss}. These results show that including the optimizer response improves agreement with the oracle, suggesting that it provides a more accurate estimate of one-step learning utility.

\noindent\textbf{Generation Quality at Matched Inference Cost.} GOT-FM modifies only the training-time coupling and introduces no additional inference operations. Figure \ref{fig:latency} shows that GOT-FM achieves the lowest FID among the evaluated methods while maintaining comparable inference latency at matched NFEs. This shows that the improvements in generation quality do not come at the cost of increased inference computation. 

%Figure~\ref{fig:latency} compares FID against measured inference latency on CIFAR-10~\cite{krizhevsky2009learning}. Across all evaluated NFEs, GOT-FM achieves the lowest FID while also exhibiting lower measured inference latency than the baselines at matched NFEs. These results demonstrate a favorable quality--latency trade-off without modifying the inference architecture. Further computational costs are reported in Appendix~\ref{sec:sup_computational}.

%Figure~\ref{fig:latency} compares FID against measured inference latency on CIFAR-10~\cite{krizhevsky2009learning}. At matched NFEs, all methods exhibit similar inference latency, while GOT-FM achieves lower FID. GOT-FM achieves the lowest FID at every evaluated NFE while maintaining comparable inference latency. Further computational costs are reported in Appendix~\ref{sec:sup_computational}.

%Finally, we examine whether the improved generation quality also leads to faster sampling. Figure \ref{fig:latency} reports inference latency at matched NFEs, showing comparable latency across methods. Since GOT-FM changes only the training procedure, this saving comes from requiring fewer NFEs for the same quality. Cost comparisons with other baselines are provided in Appendix~\ref{sec:sup_computational}.

%causal alternatives for the repeated sparse policy
\begin{table}[t]
  \centering
  \vspace{-3mm}
  \caption{\textbf{Effect of sparse-intervention controls.}
  We compare training policies under matched intervention
  frequency, timing, and candidate pool
  for the sparse variants. We report FID \cite{heusel2017gans} at 4, 8, and 16 NFEs.}
  \vspace{-0.5em}
  \label{tab:ot_ablation_policy}
  \footnotesize
  \setlength{\tabcolsep}{3pt}
  \renewcommand{\arraystretch}{1.10}
  \begin{tabular*}{\columnwidth}
    {@{\extracolsep{\fill}}lccc@{}}
    \toprule
    Training policy
    & $\mathrm{FID}_4 \downarrow$
    & $\mathrm{FID}_8 \downarrow$
    & $\mathrm{FID}_{16} \downarrow$ \\
    \midrule
    OT-only       & 39.8 & 21.2 & 14.7 \\
    Sparse-Random & 40.1 & 22.5 & 12.9 \\
    Sparse-MAC \cite{lin2026beyond}  & 39.5 & 22.0 & 12.6 \\
    \textbf{GOT-FM (Ours)}  & $\mathbf{36.0}$ & $\mathbf{17.6}$ & $\mathbf{12.1}$ \\
    \bottomrule
  \end{tabular*}
  \vspace{-1mm}
\end{table}

%\begin{table}[t]
 % \centering
  %\caption{\textbf{Effect of Optimizer Response.}}
  %\vspace{-0.5em}
  %\label{tab:optimizer_response}
 % \footnotesize
 % \setlength{\tabcolsep}{3pt}
 % \renewcommand{\arraystretch}{1.10}
 % \begin{tabular*}{\columnwidth}{@{\extracolsep{\fill}}lcc@{}}
 %   \toprule
 %   Criterion & \shortstack{Actual Gain $\uparrow$} & \shortstack{Oracle Regret $\downarrow$} \\
 %   \midrule
  %  Lookahead Alignment (w/o $\mathbf{R}_0$) & --- & --- \\
   % \textbf{GOT-FM} (w/ $\mathbf{R}_0$) & --- & --- \\
   % \bottomrule
  %\end{tabular*}
  %\vspace{-2mm}
%\end{table}

%\begin{table}[t]
%  \centering
 % \caption{\textbf{Inference latency at different NFEs.}
 % We compare GOT-FM with Vanilla Rectified Flow, OT-CFM, and MAC on CIFAR-10 \cite{krizhevsky2009learning},
 % reporting latency in ms/image (lower is better).}
 % \vspace{-0.5em}
 % \label{tab:inference_latency}
 % \footnotesize
 % \setlength{\tabcolsep}{2pt}
 % \renewcommand{\arraystretch}{1.10}
 % \begin{tabular*}{\columnwidth}{@{\extracolsep{\fill}}lccc@{}}
 %   \toprule
  %  Method & NFE=1 & NFE=10 & NFE=50 \\
   % \midrule
  %  Rectified Flow \cite{liu2022flow} & $13.04 \pm 0.02$ & $128.65 \pm 1.97$ & $641.22 \pm 4.20$ \\
  %  OT-CFM \cite{tong2023improving} & $13.03 \pm 0.02$ & $128.48 \pm 1.69$ & $640.25 \pm 1.94$ \\
  %  MAC-full \cite{lin2026beyond} & $13.04 \pm 0.02$ & $128.73 \pm 2.17$ & $641.36 \pm 4.87$ \\
  %  \textbf{GOT-FM (Ours)} & $13.09 \pm 0.31$ & $128.56 \pm 1.66$ & $640.72 \pm 3.89$ \\
  %  \bottomrule
  %\end{tabular*}
  %\vspace{-2mm}
%\end{table}

%\begin{figure}[t]
  %\centering
  %\includegraphics[width=\linewidth]{fig/efficiency.pdf}
  %\caption{\textbf{Training and inference efficiency on CIFAR-10~\cite{krizhevsky2009learning}.}}
  %\label{fig:efficiency}
%\end{figure}

\section{Conclusion}
\label{sec:conclusion}
We propose \textbf{GOT-FM}, a gradient-aware coupling method for Flow Matching. Since GOT-FM only changes how samples are paired, it can be applied to existing FM models such as Rectified Flow and MeanFlow without changing the model architecture. Experiments on CIFAR-10 and ImageNet $256\times256$ showed that GOT-FM consistently improves FID over FM baselines across sampling steps. Ablation studies further highlight the importance of optimizer-conditioned learning utility in selecting couplings. These results suggest that selecting couplings according to their estimated effect on learning is an effective strategy for improving FM. However, GOT-FM relies on a local one-step approximation and a finite candidate pool, which may limit its effectiveness over longer training horizons. Future work will explore more efficient coupling search and learning-aware criteria that account for global optimization effects.

%\clearpage
%\input{sec/7_LLM.tex}

{
    \small
    \bibliographystyle{ieeenat_fullname}
    \bibliography{main}
}

% WARNING: do not forget to delete the supplementary pages from your submission 
\clearpage
\setcounter{page}{1}
\maketitlesupplementary

\vspace{2em}
% \begin{center}
% \Large\textbf{Appendix}
% \end{center}
\vspace{1.5em}
\begin{center}
    \large\textbf{Supplementary Contents}
\end{center}

\begingroup
\footnotesize
\color{cvprblue}
% Omit the CVPR heading prefix when reusing titles in the contents.
\renewcommand{\texorpdfstring}[2]{#2}

\newcommand{\suppsection}[1]{%
    \noindent
    {\small\hypersetup{linkcolor=black}\color{black}%
    \textbf{\ref{#1}.\ \nameref{#1}}}%
    \nobreak\hspace{0.5em}\dotfill\nobreak\hspace{0.5em}\pageref{#1}\par
    \vspace{0.5em}
}

\newcommand{\suppsubsection}[1]{%
    \noindent\hspace*{1.5em}%
    {\hypersetup{linkcolor=black}\color{black}%
    \ref{#1}.\ \nameref{#1}}%
    \nobreak\hspace{0.5em}\dotfill\nobreak\hspace{0.5em}\pageref{#1}\par
    \vspace{0.2em}
}

% Contents
\suppsection{sec:sup_ablation}
\suppsubsection{sec:sup_ablation_frequency}
\suppsubsection{sec:sup_ablation_timing}
\suppsubsection{sec:sup_ablation_batch}
\suppsubsection{sec:sup_ablation_utility}
\suppsubsection{sec:sup_ablation_candidate}
\suppsubsection{sec:sup_ablation_optimizer}

\suppsection{sec:sup_counterexample}
\suppsubsection{sec:sup_counterexample_2}
\suppsubsection{sec:sup_counterexample_high}
\suppsubsection{sec:sup_empirical_counterexample}

\suppsection{sec:sup_objective}
\suppsubsection{sec:sup_objective_1}
\suppsubsection{sec:sup_objective_2}
\suppsubsection{sec:sup_objective_3}

\suppsection{sec:sup_theory}
\suppsubsection{sec:sup_theory_1}
\suppsubsection{sec:sup_theory_2}
\suppsubsection{sec:sup_theory_3}
\suppsubsection{sec:sup_theory_marginal}
\suppsubsection{sec:sup_theory_4}

\suppsection{sec:sup_coupling}
\suppsubsection{sec:sup_coupling_1}
\suppsubsection{sec:sup_coupling_2}
\suppsubsection{sec:sup_coupling_3}
\suppsubsection{sec:sup_coupling_4}

\suppsection{sec:sup_meanflow}

\suppsection{sec:sup_conditional}

\suppsection{sec:sup_implementation}

\suppsection{sec:sup_computational}
\suppsubsection{sec:sup_computational_2}
\suppsubsection{sec:sup_computational_1}

\suppsection{sec:sup_qualitative}

\suppsection{sec:sup_discussion}
\suppsubsection{sec:sup_discussion_1}
\suppsubsection{sec:sup_discussion_2}
\suppsubsection{sec:sup_discussion_3}
\endgroup

\section{Additional Ablation Analysis}
\label{sec:sup_ablation}
\subsection{Frequency of Gradient-Aware Selection}
\label{sec:sup_ablation_frequency}
We first examine how often gradient-aware selection needs to be applied during training. We denote by $q$ the fraction of training updates at which gradient-aware selection is performed. Figure \ref{fig:hyper_frequency} varies the selection frequency $q$. With $q=1/8$, GOT-FM achieves the lowest FID at 4 and 8 NFEs, while reducing the number of selection calls. This shows that gradient-aware selection does not need to be applied at every update.
\subsection{Timing of Gradient-Aware Selection}
\label{sec:sup_ablation_timing}
\begin{figure}[h]
  \centering
  % Crop the PDFs' empty bottom margins to keep subcaptions close to the plots.
  \begin{subfigure}[b]{0.48\linewidth}
    \centering
    \includegraphics[width=\linewidth,trim=0 20bp 0 0,clip]{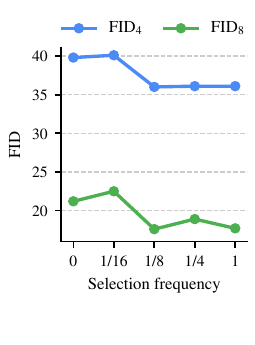}
    \caption{Selection frequency}
    \label{fig:hyper_frequency}
  \end{subfigure}
  \hfill
  \begin{subfigure}[b]{0.48\linewidth}
    \centering
    \includegraphics[width=\linewidth,trim=0 20bp 0 0,clip]{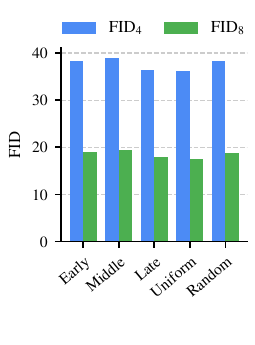}
    \caption{Intervention timing}
    \label{fig:intervention_timing}
  \end{subfigure}
  \vspace{-1mm}
  \caption{\textbf{Effects of selection frequency and intervention timing on CIFAR-10 \cite{krizhevsky2009learning}.} (a) Generation FID \cite{heusel2017gans} at 4 and 8 NFEs under different selection frequencies $q$. (b) Generation FID under different intervention schedules with the same number of gradient-aware selection calls.}
  \vspace{-0.5em}
  \label{fig:hyper_ablation}
\end{figure}

Figure~\ref{fig:intervention_timing} compares five schedules with
the same budget of $H$ selection calls. Let $T$ denote the number
of continuation updates after the common branch point, indexed
from $1$ to $T$. Early, Middle, and Late each place all $H$ calls
in a contiguous window. Early uses updates $1,\ldots,H$.
Middle uses updates $s+1,\ldots,s+H$, where $s=\lfloor(T-H)/2\rfloor$,
and Late uses updates $T-H+1,\ldots,T$. Random selects $H$ update indices
uniformly without replacement from $\{1,\ldots,T\}$.
Uniform applies selection periodically at the configured interval
of eight updates throughout the continuation.
Among these schedules, Uniform achieves the lowest FID at both
4 and 8 NFEs under the same selection-call budget.

\vspace{-0.5em}
\subsection{Effect of Reference Batch Size}
\label{sec:sup_ablation_batch}

Since GOT-FM estimates learning utility using a finite reference bank,
the reliability of coupling selection depends on the gradient information
provided by the reference objective. We investigate this dependence by
varying the OT reference matching size $B_{\mathrm{ref}}$, which determines
the minibatch size used to construct the OT reference coupling, while
keeping the reference mixture ratio and other training configurations fixed.

We evaluate coupling-selection stability by resampling the reference bank
while holding the model parameters, optimizer state, and candidate pool
fixed. Stability measures selection agreement across independent reference
draws, while gain correlation measures the Spearman rank correlation
between predicted gains and actual gains measured on an independent
evaluation set.

As shown in Table~\ref{tab:reference_batch_size}, increasing
$B_{\mathrm{ref}}$ consistently improves both metrics over the evaluated
range. As $B_{\mathrm{ref}}$ increases from 8 to 128, selection agreement
rises from 56.25\% to 97.40\%, and gain correlation increases from 0.433
to 0.828. These results provide empirical evidence that larger OT
reference matchings yield more stable coupling decisions and more
reliable learning-utility estimates in this setting.

Although downstream FID does not vary monotonically with
$B_{\mathrm{ref}}$, the configuration with the highest stability and gain
correlation, $B_{\mathrm{ref}}=128$, also achieves the lowest FID across
all evaluated NFEs. These observations complement the analysis of
approximate scoring in Appendix~\ref{sec:sup_theory_marginal}, which
characterizes coupling-selection stability in terms of scoring errors
relative to the score gap between competing candidates.

\begin{table}[t]
    \centering
    \caption{\textbf{Effect of OT reference batch size
    $B_{\mathrm{ref}}$ on CIFAR-10~\cite{krizhevsky2009learning}.}
    Stability denotes coupling-selection agreement
    across independent reference draws.
    Gain Corr. ($\rho_s$) denotes the Spearman
    correlation between predicted and independent
    actual gains.
    FID~\cite{heusel2017gans} is reported at 1, 4, 8, 16, and 50 NFEs.}
    \label{tab:reference_batch_size}
    \vspace{-0.5em}
    \footnotesize
    \setlength{\tabcolsep}{3pt}
    \renewcommand{\arraystretch}{1.15}

    \resizebox{\columnwidth}{!}{%
    \begin{tabular}{@{}lccccccc@{}}
        \toprule
        $B_{\mathrm{ref}}$
        & \shortstack{Stability\\(\%) ($\uparrow$)}
        & \shortstack{Gain Corr.\\($\rho_s$) ($\uparrow$)}
        & FID$_1$($\downarrow$)
        & FID$_4$($\downarrow$)
        & FID$_8$($\downarrow$)
        & FID$_{16}$($\downarrow$)
        & FID$_{50}$($\downarrow$) \\
        \midrule
        8 & 56.25 & 0.433 & 255.2 & 37.5 & 20.1 & 16.1 & 13.4 \\
        16 & 59.38 & 0.547 & 246.3 & 38.1 & 21.1 & 17.1 & 14.4 \\
        32 & 78.13 & 0.569 & 248.5 & 38.2 & 20.6 & 16.5 & 12.9 \\
        64 & 90.63 & 0.634 & 244.4 & 37.5 & 20.4 & 14.6 & 10.0 \\
        \textbf{128 (GOT-FM)} & \textbf{97.40} & \textbf{0.828} & \textbf{242.1} & \textbf{36.0} & \textbf{17.6} & \textbf{12.1} & \textbf{8.1} \\
        \bottomrule
    \end{tabular}%
    }
    \vspace{-0.5em}
\end{table}

\begin{table}
    \centering
    \caption{\textbf{Effect of one-step learning utility.}
    We compare predicted and measured improvements
    over the Euclidean OT baseline update.
    Virtual updates are evaluated using the scoring
    reference or an independent evaluation set.
    Positive rate denotes the fraction of evaluated
    events with positive gain.}
    \label{tab:one_step_utility}
    \vspace{-0.5em}
    \footnotesize
    \setlength{\tabcolsep}{3pt}
    \renewcommand{\arraystretch}{1.12}
    \begin{tabular}{@{}lcc@{}}
        \toprule
        Evaluation
        & Mean gain ($\uparrow$)
        & Positive (\%) ($\uparrow$) \\
        \midrule
        Linearized predicted gain
        & $+9.955\times10^{-6}$ & 97.9 \\
        Virtual update (reference)
        & $+9.608\times10^{-6}$ & 97.9 \\
        Virtual update (independent)
        & $+8.583\times10^{-6}$ & 93.6 \\
        Executed update (independent)
        & $+8.945\times10^{-6}$ & 94.1 \\
        \bottomrule
    \end{tabular}
    \vspace{-1em}
\end{table}

\subsection{Evaluation of One-Step Learning Utility}
\label{sec:sup_ablation_utility}
We further evaluate the one-step learning utility examined in Section \ref{sec:exp_ablation}. 
Table~\ref{tab:one_step_utility} compares four measures of improvement relative to the Euclidean OT baseline update. 
Linearized prediction estimates the gain using GOT-FM's gradient-aware cost, while virtual update (reference) measures the exact gain after applying separate optimizer updates and evaluating them on the shared reference objective. Virtual update (independent) evaluates the same updated models on an independent evaluation set, whereas executed update (independent) measures the gain after a stochastic training update with newly sampled randomness. We report the mean gain and positive rate for each setting. The linearized prediction is close to the exact virtual-update gain. The independently evaluated gains remain positive on average, although they are smaller than the gain measured on the scoring reference.

\subsection{Effect of Candidate Pool}
\label{sec:sup_ablation_candidate}
We further examine the effect of candidate-pool
composition beyond the analysis in
Section \ref{sec:exp_ablation}.
GOT-FM selects a coupling from a finite pool,
so its performance may depend on the available
candidates.
We vary the ratio of geometry/endpoint-based
(G) and random-cost (R) proposals while keeping
the configured number of alternative proposals at eight and the
feasibility constraints fixed. Every final selection pool also includes
the Euclidean OT baseline.
The ordered geometry/endpoint proposal library and random-cost objectives
are defined in Appendix~\ref{sec:sup_coupling_1}.
This analysis examines whether the benefits of
gradient-aware selection depend on a particular
candidate-generation strategy. Table~\ref{tab:candidate_pool_fid}
shows comparable FID across all five pool compositions, including
the all-random (0G + 8R) and all-geometry/endpoint (8G + 0R) pools.
At any fixed NFE, the reported FID varies by at most 1.0 point
across compositions, and FID$_{50}$ ranges from 8.1 to 8.3.
These results indicate that GOT-FM is relatively insensitive to
the tested candidate-pool constructions and does not rely on
a particular mixture of geometry/endpoint and random-cost proposals.

\begin{table}[t]
    \centering
    \caption{\textbf{Effect of candidate-pool composition
    on CIFAR-10~\cite{krizhevsky2009learning}.}
    We vary the ratio of geometry/endpoint-based
    (G) and random-cost (R) proposals while
    keeping eight configured alternative proposals plus the baseline.
    FID~\cite{heusel2017gans} is reported at 1, 4, 8, 16, and 50 NFEs.
    Lower FID is better.}
    \label{tab:candidate_pool_fid}
    \vspace{-0.5em}
    \footnotesize
    \setlength{\tabcolsep}{2.8pt}
    \renewcommand{\arraystretch}{1.12}
    \begin{tabular}{@{}lccccc@{}}
        \toprule
        Pool
        & FID$_1$($\downarrow$)
        & FID$_4$($\downarrow$)
        & FID$_8$($\downarrow$)
        & FID$_{16}$($\downarrow$)
        & FID$_{50}$($\downarrow$) \\
        \midrule
        0G + 8R & 242.7 & 36.4 & 18.5 & 12.4 & 8.3 \\
        2G + 6R & 242.6 & 36.5 & 18.6 & 12.4 & 8.3 \\
        \textbf{4G + 4R (GOT-FM)} & \textbf{242.1} & \textbf{36.0} & \textbf{17.6} & \textbf{12.1} & \textbf{8.1} \\
        6G + 2R & 242.6 & 36.5 & 18.6 & 12.4 & 8.3 \\
        8G + 0R & 242.7 & 36.5 & 18.5 & 12.4 & 8.3 \\
        \bottomrule
    \end{tabular}
    \vspace{-0.5em}
\end{table}

\subsection{Effect of Optimizer Response}
\label{sec:sup_ablation_optimizer}
We extend the optimizer-response ablation
in Figure~\ref{fig:optimizer} by reporting
generation FID at additional NFEs.
We compare GOT-FM with a gradient-alignment
variant that excludes the optimizer response
$\mathbf R_0$:
$C_{ij}^{\mathrm{Align}}
=-\mathbf g_{ij}^{\top}\mathbf g_{\mathrm{ref}}^+$.
GOT-FM and the gradient-alignment variant use the same candidate-generation
and coupling-selection settings.

\begin{table}[t]
    \centering
    \caption{\textbf{Effect of optimizer response
    on CIFAR-10~\cite{krizhevsky2009learning} generation.}
    We compare gradient-alignment selection
    without $\mathbf R_0$ against GOT-FM with $\mathbf R_0$.
    FID~\cite{heusel2017gans} is reported at 1, 4, 8, 16, and 50 NFEs.
    Lower FID is better.}
    \label{tab:optimizer_response_fid}
    \vspace{-0.5em}
    \footnotesize
    \setlength{\tabcolsep}{3pt}
    \renewcommand{\arraystretch}{1.12}
    \begin{tabular}{@{}lccccc@{}}
        \toprule
        Method
        & FID$_1$($\downarrow$)
        & FID$_4$($\downarrow$)
        & FID$_8$($\downarrow$)
        & FID$_{16}$($\downarrow$)
        & FID$_{50}$($\downarrow$) \\
        \midrule
        w/o $\mathbf R_0$
        & 271.8 & 40.3 & 20.0 & 14.7 & 10.8 \\
        \textbf{w/ $\mathbf R_0$ (GOT-FM)}
        & \textbf{242.1} & \textbf{36.0}  & \textbf{17.6} & \textbf{12.1} & \textbf{8.1} \\
        \bottomrule
    \end{tabular}
    \vspace{-1em}
\end{table}

As shown in Section~\ref{sec:exp_ablation}, incorporating
the optimizer response improves agreement
with the virtual-update oracle.
Table~\ref{tab:optimizer_response_fid} examines whether
this improvement in coupling selection
also benefits downstream generation quality.
GOT-FM with $\mathbf R_0$ achieves lower FID than the variant
without $\mathbf R_0$ at every evaluated NFE.
%Together, these evaluations help distinguish optimizer-conditioned coupling selection from gradient alignment alone.

\section{Detailed Counterexample and Its Extension}
\label{sec:sup_counterexample}
\subsection{Detailed Two-Dimensional Counterexample}
\label{sec:sup_counterexample_2}
We provide the calculations for the two-layer linear model in Section \ref{sec:counterexample}. Both weight matrices are trainable, and
all losses below use uniform flow time $t\sim\mathcal U[0,1]$.

\noindent\textbf{Initial prediction error.}
Write the source and target endpoints as
$\mathbf z=(\sigma,0)^\top$ and
$\mathbf y=(0,\tau)^\top$, where $\sigma,\tau\in\{-1,1\}$.
Their interpolated input and target velocity are
\begin{equation}
\label{eq:app-2d-path}
\mathbf x^t=\bigl((1-t)\sigma,t\tau\bigr)^\top,
\qquad
\mathbf u=(-\sigma,\tau)^\top.
\end{equation}
With the initial weights
$\mathbf W_{1,0}=\bigl(\begin{smallmatrix}1&1\\0&1\end{smallmatrix}\bigr)$
and
$\mathbf W_{2,0}=\bigl(\begin{smallmatrix}-1&1\\0&2\end{smallmatrix}\bigr)$,
we have
$\mathbf W_{2,0}\mathbf W_{1,0}
=\operatorname{diag}(-1,2)$.
The prediction residual is therefore
\begin{equation}
\label{eq:app-2d-residual}
\begin{aligned}
\mathbf r_t
&=\mathbf W_{2,0}\mathbf W_{1,0}\mathbf x^t-\mathbf u\\
&=\bigl(t\sigma,(2t-1)\tau\bigr)^\top.
\end{aligned}
\end{equation}
Since the data dimension is $d=2$, the instantaneous FM
loss is $\ell(\boldsymbol\theta_0,t)
=\tfrac12\|\mathbf r_t\|_2^2
=\tfrac12[t^2+(2t-1)^2]$, independently of the pair.
Thus, each coupling has the same initial full-time loss,
\begin{equation}
\label{eq:app-2d-initial-loss}
\begin{aligned}
\mathcal L(\boldsymbol\theta_0;\boldsymbol\pi_A)
&=\mathcal L(\boldsymbol\theta_0;\boldsymbol\pi_B)\\
&=\frac12\int_0^1[t^2+(2t-1)^2]\,dt
=\frac13.
\end{aligned}
\end{equation}

\noindent\textbf{Coupling-dependent parameter gradients.}
For the factorized velocity model
$\mathbf v_{\boldsymbol\theta}(\mathbf x,t)=
\mathbf W_2\mathbf W_1\mathbf x$, the gradients of the
per-pair loss $\frac12\|\mathbf W_2\mathbf W_1\mathbf x^t-
\mathbf u\|_2^2$ at initialization are
\begin{equation}
\label{eq:app-2d-pair-gradients}
\begin{gathered}
\nabla_{\mathbf W_2}\ell
=\mathbf r_t(\mathbf W_{1,0}\mathbf x^t)^\top,\\
\nabla_{\mathbf W_1}\ell
=\mathbf W_{2,0}^\top\mathbf r_t(\mathbf x^t)^\top.
\end{gathered}
\end{equation}
Define $\kappa=\sigma\tau\in\{-1,1\}$.
The diagonal coupling $\boldsymbol\pi_A$ uses
$\kappa=+1$, whereas the off-diagonal coupling
$\boldsymbol\pi_B$ uses $\kappa=-1$.
The two pairs within each coupling give identical
parameter gradients, because replacing
$(\sigma,\tau)$ by $(-\sigma,-\tau)$ changes the signs
of both $\mathbf r_t$ and $\mathbf x^t$.
Integrating Eq.~\eqref{eq:app-2d-pair-gradients}
over $t\in[0,1]$ gives
\begin{equation}
\label{eq:app-2d-gradient-matrices}
\begin{gathered}
\mathbf G_{1,\kappa}
:=\nabla_{\mathbf W_1}\mathcal L
 =\begin{pmatrix}
 -\frac16 & -\frac\kappa3\\[2pt]
 \frac{1-2\kappa}{6} & \frac{1+\kappa}{3}
 \end{pmatrix},\\[5pt]
\mathbf G_{2,\kappa}
:=\nabla_{\mathbf W_2}\mathcal L
 =\begin{pmatrix}
 \frac{1+2\kappa}{6} & \frac\kappa3\\[2pt]
 \frac{1-\kappa}{6} & \frac16
 \end{pmatrix}.
\end{gathered}
\end{equation}
Here, $\mathcal L$ is evaluated at $\boldsymbol\theta_0$
and the coupling specified by $\kappa$.
Hence, the two couplings induce different gradients
even though their initial pairwise prediction errors agree.

\noindent\textbf{One-step SGD updates.}
Starting from the \emph{same} initial weights, one SGD
step updates both matrices simultaneously:
\begin{equation}
\label{eq:app-2d-SGD}
\mathbf W_{k,\kappa}^+
=\mathbf W_{k,0}-\eta\mathbf G_{k,\kappa},
\qquad k\in\{1,2\}.
\end{equation}
For example, substituting $\kappa=+1$ and $\kappa=-1$
into Eq.~\eqref{eq:app-2d-gradient-matrices} gives
\begin{equation}
\label{eq:app-2d-updates-A}
\begin{gathered}
\mathbf G_{1,A}=
\begin{pmatrix}-\frac16&-\frac13\\-\frac16&\frac23\end{pmatrix},\quad
\mathbf G_{2,A}=
\begin{pmatrix}\frac12&\frac13\\0&\frac16\end{pmatrix},\\[4pt]
\mathbf G_{1,B}=
\begin{pmatrix}-\frac16&\frac13\\\frac12&0\end{pmatrix},\quad
\mathbf G_{2,B}=
\begin{pmatrix}-\frac16&-\frac13\\\frac13&\frac16\end{pmatrix}.
\end{gathered}
\end{equation}
The post-update effective velocity matrices are
$\mathbf V_\kappa^+
=\mathbf W_{2,\kappa}^+\mathbf W_{1,\kappa}^+$.
Importantly, the product is formed \emph{after} updating
each factor. A single-matrix SGD update on
$\mathbf W_2\mathbf W_1$ is not equivalent.

\noindent\textbf{Evaluation on a common IID FM objective.}
To compare the two updates using the same criterion,
we evaluate both on the independent coupling of the
four endpoint pairs:
\begin{equation}
\label{eq:app-2d-IID-definition}
\begin{aligned}
J_{\mathrm{IID}}(\boldsymbol\theta)
&=\frac14\sum_{\sigma,\tau\in\{-1,1\}}\\
&\quad\int_0^1\frac12\left\|
\mathbf W_2\mathbf W_1\mathbf x^t-\mathbf u
\right\|_2^2\,dt.
\end{aligned}
\end{equation}
Let $\mathbf V=\mathbf W_2\mathbf W_1$
with entries $V_{ab}$. Using
$\mathbb E[\sigma\tau]=0$ and
$\mathbb E[\sigma^2]=\mathbb E[\tau^2]=1$,
Eq.~\eqref{eq:app-2d-IID-definition} reduces to
\begin{equation}
\label{eq:app-2d-IID-closed}
J_{\mathrm{IID}}(\mathbf V)
=1+\frac{V_{11}-V_{22}}{2}
+\frac16\|\mathbf V\|_F^2.
\end{equation}
At initialization, $\mathbf V_0=\operatorname{diag}(-1,2)$,
so $J_{\mathrm{IID}}(\boldsymbol\theta_0)=1/3$.
Substituting the separate SGD-updated factors from
Eq.~\eqref{eq:app-2d-SGD} yields
\begin{equation}
\label{eq:app-2d-IID-after-A}
\begin{aligned}
J_{\mathrm{IID}}(\boldsymbol\theta_A^+)
&=\frac13-\frac{\eta}{3}
+\frac{107\eta^2}{108}\\
&\quad{}-\frac{5\eta^3}{72}
+\frac{23\eta^4}{3888},
\end{aligned}
\end{equation}
\begin{equation}
\label{eq:app-2d-IID-after-B}
\begin{aligned}
J_{\mathrm{IID}}(\boldsymbol\theta_B^+)
&=\frac13-\frac{\eta}{6}
+\frac{53\eta^2}{108}\\
&\quad{}-\frac{5\eta^3}{216}
+\frac{23\eta^4}{3888}.
\end{aligned}
\end{equation}
For the learning rate $\eta=0.1$ used in the main text,
these are $0.3098385545$ and $0.3215515175$, respectively.
Thus, the \emph{exact} one-step improvements under the
shared IID reference are
\begin{equation}
\label{eq:app-2d-exact-utility}
\begin{gathered}
U_{\mathrm{IID}}(\boldsymbol\pi_A)
=\frac13-J_{\mathrm{IID}}(\boldsymbol\theta_A^+)
\approx 0.023495,\\
U_{\mathrm{IID}}(\boldsymbol\pi_B)
=\frac13-J_{\mathrm{IID}}(\boldsymbol\theta_B^+)
\approx 0.011782.
\end{gathered}
\end{equation}
These values follow from the explicitly computed SGD
updates. They are not first-order Taylor estimates.
The example shows that equal current velocity-prediction
errors do not determine one-step learning utility, because
the parameter-space gradients depend on the factorized
model and on the selected coupling. It does not imply
that one coupling is preferred under every reference
objective or model parameterization.

\subsection{A Full-Dimensional Counterexample}
\label{sec:sup_counterexample_high}

We construct a separate counterexample in arbitrarily large even dimensions in which both endpoint distributions have full-rank covariance. This construction uses a scalar-parameter velocity model initialized at zero output, distinct from the factorized two-layer model in Section~\ref{sec:counterexample}. Two distinct couplings minimize the Euclidean transport cost and have identical initial FM prediction error at every flow time, yet produce different one-step learning utilities under the \emph{same independent-coupling FM reference}. This common reference evaluates both candidate updates without selecting either OT optimum as the evaluation law.

\paragraph{Endpoint distributions and couplings.}
Let $d=2m$ and define
\begin{equation}
\label{eq:app-full-s-matrix}
\begin{gathered}
\mathbf S=\mathbf I_m\otimes
\begin{pmatrix}0&-1\\1&0\end{pmatrix},\\
\mathbf S^\top=-\mathbf S,\quad \mathbf S^2=-\mathbf I_d.
\end{gathered}
\end{equation}
Consider the uniformly weighted empirical supports
\begin{equation}
\label{eq:app-full-supports}
\begin{gathered}
\mathcal Z_d=\{\pm\sqrt d\,\mathbf e_i\}_{i=1}^d,\\
\mathcal Y_d=(\mathbf I_d+\mathbf S)\mathcal Z_d,
\end{gathered}
\end{equation}
with probability $1/(2d)$ per atom, and let $\mu_d$ and $\nu_d$
denote the corresponding empirical distributions. Both are centered,
and their covariance matrices satisfy
\begin{equation}
\label{eq:app-full-covariances}
\operatorname{Cov}(\mathbf Z)=\mathbf I_d,
\qquad
\operatorname{Cov}(\mathbf Y)=2\mathbf I_d.
\end{equation}
Thus, both empirical marginals span $\mathbb R^d$ affinely.
Define two deterministic couplings by
\begin{equation}
\label{eq:app-full-two-plans}
\begin{gathered}
\boldsymbol\pi_A:\;\mathbf y=(\mathbf I_d+\mathbf S)\mathbf z,\\
\boldsymbol\pi_B:\;\mathbf y=(\mathbf I_d-\mathbf S)\mathbf z.
\end{gathered}
\end{equation}
Since $-\mathbf S$ permutes $\mathcal Z_d$ and
$(\mathbf I_d+\mathbf S)(-\mathbf S)=\mathbf I_d-\mathbf S$,
both mappings are bijections onto $\mathcal Y_d$. Hence
$\boldsymbol\pi_A,\boldsymbol\pi_B\in\Pi(\mu_d,\nu_d)$.

\paragraph{Equal optimal transport costs.}
Every source atom has squared norm $d$, every target atom has
squared norm $2d$, and
$\mathbf z_i^\top\mathbf y_j\in\{-d,0,d\}$. Therefore,
$\|\mathbf y_j-\mathbf z_i\|_2^2\ge d$ for all pairs.
Every edge used by either coupling satisfies
$\mathbf y-\mathbf z=\pm\mathbf S\mathbf z$ and consequently has
squared distance $d$. Thus, with
$G_{ij}=\|\mathbf y_j-\mathbf z_i\|_2^2$,
\begin{equation}
\label{eq:app-full-equal-ot}
\begin{aligned}
\langle\mathbf G,\boldsymbol\pi_A\rangle_F
&=\langle\mathbf G,\boldsymbol\pi_B\rangle_F=d\\
&=\min_{\boldsymbol\pi\in\Pi(\mu_d,\nu_d)}
\langle\mathbf G,\boldsymbol\pi\rangle_F.
\end{aligned}
\end{equation}
Therefore, both plans are global Euclidean OT minimizers.

\paragraph{Identical initial prediction errors, different gradients.}
Consider linear-path FM with
$\mathbf x^t=(1-t)\mathbf z+t\mathbf y$,
$\mathbf u=\mathbf y-\mathbf z$, and $t\sim\mathcal U[0,1]$.
Use the scalar-parameter velocity model to isolate the effect of coupling on learning gradients and enable exact one-step utility calculations
\begin{equation}
\label{eq:app-full-model}
\mathbf v_\theta(\mathbf x,t)
=\frac{\theta}{2}(\mathbf I_d+\mathbf S)\mathbf x,
\qquad \theta_0=0,
\end{equation}
with the usual per-coordinate FM loss
$\ell(\theta;\mathbf z,\mathbf y,t)
=d^{-1}\|\mathbf v_\theta(\mathbf x^t,t)-\mathbf u\|_2^2$.
Write $\varepsilon=+1$ for $\boldsymbol\pi_A$ and
$\varepsilon=-1$ for $\boldsymbol\pi_B$. On either plan,
$\mathbf y=(\mathbf I_d+\varepsilon\mathbf S)\mathbf z$ and
$\mathbf x^t=\mathbf z+\varepsilon t\mathbf S\mathbf z$.
Using $\mathbf z^\top\mathbf S\mathbf z=0$ and
$\|\mathbf S\mathbf z\|_2^2=\|\mathbf z\|_2^2=d$ gives
\begin{equation}
\label{eq:app-full-time-loss}
\ell_\varepsilon(\theta,t)
=1-(\varepsilon+t)\theta
+\frac{1+t^2}{2}\theta^2.
\end{equation}
In particular, $\ell_A(0,t)=\ell_B(0,t)=1$ for
\emph{every} $t\in[0,1]$. For any other edge, the initial
velocity is zero, so its prediction error is
$\|\mathbf y_j-\mathbf z_i\|_2^2/d\ge1$.
Both couplings also globally minimize the initial full-time FM
prediction-error objective. Nevertheless, integrating over time yields
\begin{equation}
\label{eq:app-full-training-losses}
\begin{gathered}
\mathcal L(\theta;\boldsymbol\pi_A)
=1-\frac32\theta+\frac23\theta^2,\\
\mathcal L(\theta;\boldsymbol\pi_B)
=1+\frac12\theta+\frac23\theta^2,
\end{gathered}
\end{equation}
with distinct gradients at $\theta_0$:
$\nabla_\theta\mathcal L(0;\boldsymbol\pi_A)=-3/2$
and $\nabla_\theta\mathcal L(0;\boldsymbol\pi_B)=+1/2$.

\paragraph{Different one-step utilities under a common IID reference.}
We evaluate both updated models using the same standard FM risk.
Specifically, $(\mathbf Z,\mathbf Y)\sim\mu_d\otimes\nu_d$ and
$T\sim\mathcal U[0,1]$ are sampled independently.
Writing $\mathbf M=\mathbf I_d+\mathbf S$, we have
$\mathbf M^\top\mathbf M=2\mathbf I_d$ and, for fixed $t$,
\begin{equation}
\label{eq:app-full-iid-moments}
\begin{gathered}
\frac1d\mathbb E\|\mathbf Y-\mathbf Z\|_2^2=3,\\
\frac1d\mathbb E[(\mathbf Y-\mathbf Z)^\top
\mathbf M\mathbf X^t]=3t-1,\\
\frac1d\mathbb E\|\mathbf M\mathbf X^t\|_2^2
=2[(1-t)^2+2t^2].
\end{gathered}
\end{equation}
Integrating the resulting FM risk over uniform flow time gives
\begin{equation}
\label{eq:app-full-iid-reference}
\begin{aligned}
J_{\mathrm{IID}}(\theta)
&=\mathbb E[\ell(\theta;\mathbf Z,\mathbf Y,T)]\\
&=3-\frac12\theta+\frac12\theta^2.
\end{aligned}
\end{equation}
This reference is determined by the product of the endpoint
marginals, independently of which candidate coupling is evaluated.
A single full-gradient SGD update with learning rate $\eta$ from
$\theta_0=0$ gives
$\theta_A^+=3\eta/2$ and $\theta_B^+=-\eta/2$.
Defining
$U_{\mathrm{IID}}(\boldsymbol\pi)
=J_{\mathrm{IID}}(0)-J_{\mathrm{IID}}(\theta_{\boldsymbol\pi}^+)$,
we obtain the \emph{exact} one-step utilities
\begin{equation}
\label{eq:app-full-utilities}
\begin{gathered}
U_{\mathrm{IID}}(\boldsymbol\pi_A)
=\frac34\eta-\frac98\eta^2,\\
U_{\mathrm{IID}}(\boldsymbol\pi_B)
=-\frac14\eta-\frac18\eta^2.
\end{gathered}
\end{equation}
At $\eta=0.1$, these are $+0.06375$ and $-0.02625$,
respectively. Their difference is $\eta(1-\eta)$.
These values are computed by evaluating the reference after the
SGD update, not by a first-order Taylor approximation. In
particular, the utilities have opposite signs for $0<\eta<2/3$.

\paragraph{Robustness to stochastic flow-time sampling.}
The sign contrast does not require an exact time-integrated
training gradient. If a fresh $T\sim\mathcal U[0,1]$ is used for
one SGD step, the gradients are
$g_A(T)=-(1+T)$ and $g_B(T)=1-T$.
For $0<\eta<1/2$, the resulting updates give positive IID utility
for $A$ at every $T$ and negative utility for $B$ except at $T=1$.
Evaluating both updates on the same exact IID reference gives
\begin{equation}
\label{eq:app-full-stochastic-utilities}
\begin{gathered}
\mathbb E_T U_{\mathrm{IID}}(\boldsymbol\pi_A;T)
=\frac34\eta-\frac76\eta^2,\\
\mathbb E_T U_{\mathrm{IID}}(\boldsymbol\pi_B;T)
=-\frac14\eta-\frac16\eta^2.
\end{gathered}
\end{equation}
These are distinct from the full-gradient utilities in
Eq.~\eqref{eq:app-full-utilities}.

\paragraph{Connection to the IID/OT reference.}
For comparison with the reference family used in GOT-FM, one may
choose the symmetric Euclidean OT optimum
$Q_{\mathrm{OT,sym}}=(\boldsymbol\pi_A+\boldsymbol\pi_B)/2$,
which does not favor either candidate through OT tie-breaking.
It induces
$J_{\mathrm{OT,sym}}(\theta)=1-\theta/2+2\theta^2/3$.
The equal-weight reference is therefore
\begin{equation}
\label{eq:app-full-symmetric-reference}
\begin{aligned}
J_{\mathrm{mix,sym}}(\theta)
&=\tfrac12J_{\mathrm{IID}}(\theta)
+\tfrac12J_{\mathrm{OT,sym}}(\theta)\\
&=2-\frac12\theta+\frac7{12}\theta^2.
\end{aligned}
\end{equation}
For the same SGD updates at $\eta=0.1$, the corresponding utilities
are $U_{\mathrm{mix,sym}}(\boldsymbol\pi_A)=+0.061875$ and
$U_{\mathrm{mix,sym}}(\boldsymbol\pi_B)\approx-0.0264583$.
This symmetric reference is an analytical example, not a claim that
the finite OT reference used in the image experiments follows
this particular tie-breaking rule.

%\paragraph{Scope.}
%The construction works for arbitrary even $d$ and shows that fixed marginals, globally optimal Euclidean transport cost, and identical initial full-time FM prediction error do not determine one-step learning utility under a common IID reference. It uses discrete full-rank empirical marginals, a zero-output scalar-parameter model at initialization, non-unique OT optima, and a specified reference and optimizer step. It does not establish an invariant ranking under every reference or guarantee long-horizon or image-generation improvements. The model and parameterization differ from the two-layer $2\times2$ counterexample in the main text.

\subsection{Empirical Analysis of Cost-Matched Couplings}
\label{sec:sup_empirical_counterexample}

The counterexamples in
Appendices~\ref{sec:sup_counterexample_2}
and~\ref{sec:sup_counterexample_high} establish that identical
geometric costs and current prediction errors can coexist with
different one-step learning utilities. We examine the corresponding
empirical question in trained neural networks: whether nearly
cost-matched couplings exhibit measured utility differences, and
which criteria identify candidates with higher utility.

\paragraph{Experimental setup.}
We use two trained U-Net checkpoints and four minibatch banks
per checkpoint, giving eight comparison events with eight
candidate couplings each. All candidates preserve both empirical
marginals. We form matched coupling pairs before measuring their
actual learning utilities. We evaluate the full-time FM prediction
error using 32-node time integration.
Here, full-time error refers to this numerical approximation to
the time integral, rather than an exact analytical integral or
the two-endpoint proposal cost in
Eq.~\eqref{eq:app-endpoint-error-proposal}.

Within each event $e$, we apply separate one-step Adam updates
from the same model parameters $\boldsymbol\theta_e$ and
pre-update optimizer state $s_e$. We evaluate the resulting
models on a common independent CIFAR-10 test set, denoting
its FM objective by $\widehat J_{\mathrm{eval},e}$.
The measured utility is
\begin{equation}
\label{eq:app_cost_matched_eval_utility}
\begin{aligned}
U_e^{\mathrm{eval}}(\boldsymbol\pi)
&=\widehat J_{\mathrm{eval},e}(\boldsymbol\theta_e)\\
&\quad-\widehat J_{\mathrm{eval},e}\!\left(
\mathcal A_{s_e}(\boldsymbol\theta_e,
\mathbf g_{\boldsymbol\pi,e})\right).
\end{aligned}
\end{equation}
Pairwise utility differences are unchanged if utilities are
instead expressed relative to the same baseline coupling update,
as in Appendix~\ref{sec:sup_ablation_utility}.
The separate candidate updates are diagnostic evaluations.
During training, GOT-FM selects the candidate with the lowest
linearized cost according to Eq.~\eqref{eq:app-pool-selection}.

\paragraph{Learning utility of cost-matched couplings.}
Table~\ref{tab:matched_utility_gap} reports the absolute utility
differences between the matched couplings. The measured gaps
have a mean of $1.2111\times10^{-5}$
and a range of $1.7818\times10^{-6}$ to $2.0695\times10^{-5}$.
These observations show utility differences among trained-network
couplings whose geometric costs and current prediction errors
are closely matched. 

\begin{table}[t]
    \centering
    \caption{\textbf{Learning utility differences between
    cost-matched couplings.}
    Both empirical marginals are preserved. Full-time prediction error
    is approximated using 32-node time integration. Absolute
    utility gaps are measured on an independent CIFAR-10 test
    set after one Adam update.}
    \label{tab:matched_utility_gap}
    \footnotesize
    \setlength{\tabcolsep}{4pt}
    \begin{tabular}{@{}lc@{}}
        \toprule
        Metric & Value \\
        \midrule
        Mean absolute utility gap & $1.2111\times10^{-5}$ \\
        Median absolute utility gap & $1.2373\times10^{-5}$ \\
        Minimum absolute utility gap & $1.7818\times10^{-6}$ \\
        Maximum absolute utility gap & $2.0695\times10^{-5}$ \\
        \bottomrule
    \end{tabular}
\end{table}

\paragraph{Comparison of coupling criteria.}
% Oracle-regret values are retained as supplied. Their aggregation rule
% requires run-record confirmation. No event-wise mean is asserted here.
Table~\ref{tab:cost_matched_utility} compares utility-oriented
Spearman rank correlations with independently measured utilities and reports oracle regret for each criterion.
The geometric, prediction-error, and acceleration costs have
weak correlations in this diagnostic. Both gradient-based
criteria provide stronger ranking signals. GOT-FM
has the highest pooled rank correlation ($0.828$), while GOT-FM
also has the lowest reported oracle regret ($5.96\times10^{-6}$).
These results support an additional selection signal from
gradient information, with the optimizer-aware GOT cost yielding
the lowest reported regret among the evaluated criteria.

\begin{table}[t]
    \centering
    \caption{\textbf{Comparison of coupling-selection criteria
    in the cost-matched diagnostic.}
    Utility-oriented Spearman correlation is computed over
    all 64 candidates. Oracle regret is based on independently
    measured one-step utilities. Higher correlation and lower
    regret are better.}
    \label{tab:cost_matched_utility}
    \footnotesize
    \setlength{\tabcolsep}{7pt}
    \begin{tabular}{lcc}
        \toprule
        Selection criterion & $\rho_s$ ($\uparrow$)
        & Oracle regret ($\times10^{-6}$) ($\downarrow$) \\
        \midrule
        Euclidean cost & 0.139 & 9.69 \\
        Full-time prediction error & 0.125 & 12.50 \\
        Acceleration cost & $-0.051$ & 12.50 \\
        Raw gradient alignment & 0.458 & 7.78 \\
        \midrule
        \rowcolor{purple!8}
        \textbf{GOT cost (Ours)} & \textbf{0.828} & \textbf{5.96} \\
        \bottomrule
    \end{tabular}
\end{table}

These observations complement the fixed-marginal analysis in
Appendix~\ref{sec:sup_theory_marginal}, which characterizes
additional first-order information along directions that preserve
geometry and prediction error exactly. The empirical results
concern approximate cost matching and finite optimizer updates,
whereas the theory concerns exact equalities and local directions under
stated feasibility assumptions.

\section{Reference Objective and Problem Formulation}
\label{sec:sup_objective}

\subsection{Fixed-Marginal Coupling Gradients}
\label{sec:sup_objective_1}

We follow the fixed-marginal coupling constraints
and FM objective defined in
Section~\ref{sec:preliminaries}.
For the theoretical analysis, we use the linear
dependence of the training gradient on the coupling:
\begin{equation}
\label{eq:app-coupling-gradient}
\begin{gathered}
\mathbf g_{\boldsymbol{\pi}}
=\sum_{i,j}\pi_{ij}\mathbf g_{ij},\\
\mathbf g_{ij}
=\nabla_{\boldsymbol\theta}\ell_{ij}(\boldsymbol\theta).
\end{gathered}
\end{equation}
This relation holds when the pairwise losses,
time evaluations, and scoring randomness are
fixed while comparing candidate couplings.
It allows us to analyze how changing the joint
pair weights affects the optimizer update
without changing the empirical marginals.

\subsection{Common Reference Objective}
\label{sec:sup_objective_2}

Different coupling plans assign different weights
to source--target pairs.
To compare their resulting parameter updates,
we evaluate them using a common FM objective
that is independent of the selected coupling.

Let $Q_{\mathrm{IID}}=\mu\otimes\nu$ denote the
independent source--target law, and let
$Q_{\mathrm{OT},B_{\mathrm{ref}}}$ denote the
coupling law induced by Euclidean minibatch OT
with reference batch size $B_{\mathrm{ref}}$.
We define the reference coupling law as
\begin{equation}
\label{eq:app-reference-law}
\begin{gathered}
Q_{\mathrm{ref}}
=\lambda Q_{\mathrm{IID}}
+(1-\lambda)Q_{\mathrm{OT},B_{\mathrm{ref}}},\\
0\leq\lambda\leq1.
\end{gathered}
\end{equation}

Since FM risk is linear in the coupling law,
the corresponding reference objective is
\begin{equation}
\label{eq:app-reference-mixture}
\begin{aligned}
J_{\mathrm{ref}}(\boldsymbol\theta)
&=
\lambda J_{\mathrm{IID}}(\boldsymbol\theta)\\
&\quad{}+(1-\lambda)
J_{\mathrm{OT},B_{\mathrm{ref}}}
(\boldsymbol\theta).
\end{aligned}
\end{equation}

We use $\lambda=1/2$ to assign equal weight
to the IID and OT reference risks.
This choice does not guarantee that both
risks improve simultaneously or that the
mixture is optimal for generation quality.

Importantly, the reference law $Q_{\mathrm{ref}}$
is distinct from the baseline coupling
$\boldsymbol{\pi}_0$.
The reference defines how candidate updates
are evaluated, whereas $\boldsymbol{\pi}_0$
is the Euclidean OT plan of the current
training minibatch used for local linearization.

\subsection{Finite Reference Evaluation}
\label{sec:sup_objective_3}

In practice, the reference objective
$J_{\mathrm{ref}}$ is evaluated using a finite
set of reference samples and flow times.
We denote this evaluation by
$\widehat J_{\mathrm{ref}}$.
Within each coupling-selection event,
the reference samples, time-evaluation rule,
and scoring randomness are fixed across
all candidate couplings.

Given the current model parameters
$\boldsymbol\theta$ and optimizer state $s$,
the finite post-update objective is
\begin{equation}
\label{eq:app-finite-update-objective}
\widehat F_s(\boldsymbol{\pi})
=
\widehat J_{\mathrm{ref}}
\left(
\mathcal A_s
(\boldsymbol\theta,\mathbf g_{\boldsymbol{\pi}})
\right).
\end{equation}
The one-step improvement relative to the
baseline coupling $\boldsymbol{\pi}_0$ is
\begin{equation}
\label{eq:app-reference-utilities}
\widehat\Delta_s
(\boldsymbol{\pi};\boldsymbol{\pi}_0)
=
\widehat F_s(\boldsymbol{\pi}_0)
-
\widehat F_s(\boldsymbol{\pi}).
\end{equation}

For the optimizer-response ablation,
we define the virtual-update oracle as
the candidate with the lowest post-update
reference loss:
\begin{equation}
\label{eq:app-virtual-oracle}
\boldsymbol{\pi}_{\mathrm{oracle}}
\in
\operatorname*{arg\,min}_{\boldsymbol{\pi}\in\mathcal P^+}
\widehat F_s(\boldsymbol{\pi}).
\end{equation}
Unlike GOT-FM's linearized approximation,
the oracle evaluates a separate optimizer
update for every candidate, starting from
the same model parameters and optimizer state.

We additionally evaluate the resulting
updates on an independent reference set
that is not used for coupling selection.
This allows us to distinguish the predicted
one-step improvement from the improvement
measured on independent evaluation samples.
These one-step measurements do not directly
establish improvements over multiple
training updates or in generation quality.

\section{Theoretical Analysis of Gradient-Aware Coupling}
\label{sec:sup_theory}
\subsection{Exact Local Coupling Derivative}
\label{sec:sup_theory_1}
Here, $J_{\mathrm{ref}}$ denotes the population reference objective,
and $\widehat J_{\mathrm{ref}}$ denotes its fixed finite evaluation
defined in Appendix~\ref{sec:sup_objective_3}.
We evaluate the sensitivity of the finite post-update objective
$\widehat F_s$ to a change in the training coupling. Let
$\boldsymbol{\pi}_0$ denote the current minibatch's baseline Euclidean OT plan,
with $\mathbf g_0=\mathbf g_{\boldsymbol{\pi}_0}$. Its virtual update is
\begin{equation}
\label{eq:app-virtual-baseline}
\boldsymbol\theta_0^+
=\mathcal A_s(\boldsymbol\theta,\mathbf g_0).
\end{equation}
The subscript $0$ identifies the baseline plan, not training step
zero. We define the optimizer sensitivity and the reference gradient
at this virtual update as
\begin{equation}
\label{eq:app-optimizer-response}
\begin{gathered}
\mathbf{R}_0=D_{\mathbf g}\mathcal A_s
(\boldsymbol\theta,\mathbf g_0),\\
\mathbf{g}_\text{ref}^+=\nabla_{\boldsymbol\vartheta}
\widehat J_{\mathrm{ref}}(\boldsymbol\vartheta)
\big|_{\boldsymbol\vartheta=\boldsymbol\theta_0^+}.
\end{gathered}
\end{equation}
When $\mathcal A_s$ and the finite reference objective $\widehat J_{\mathrm{ref}}$ are
differentiable at this point, the chain rule and
Eq.~\eqref{eq:app-coupling-gradient} yield
\begin{equation}
\label{eq:app-exact-edge-derivative}
\begin{aligned}
\left.\frac{\partial\widehat F_s(\boldsymbol{\pi})}
{\partial\pi_{ij}}\right|_{\boldsymbol{\pi}=\boldsymbol{\pi}_0}
&=(\mathbf{g}_\text{ref}^+)^\top \mathbf{R}_0
\frac{\partial\mathbf g_{\boldsymbol{\pi}}}{\partial\pi_{ij}}\\
&=\mathbf g_{ij}^\top \mathbf{R}_0^\top\mathbf{g}_\text{ref}^+
=:C_{ij}^\text{GOT}.
\end{aligned}
\end{equation}
Thus, $C_{ij}^\text{GOT}$ is the exact local derivative of the
specified finite post-update objective $\widehat F_s$, which uses
$\widehat J_{\mathrm{ref}}$. The equality
does not require the parameter update to be infinitesimal. It does
not, however, make the first-order prediction exact for an arbitrary
finite change of coupling.

Because couplings lie in a constrained transport polytope, the
meaningful derivative is its action on a feasible direction
$\mathbf D$ satisfying the active coupling constraints. In
particular, if $\mathbf D\mathbf1_m=\mathbf0$ and
$\mathbf D^\top\mathbf1_n=\mathbf0$, then
\begin{equation}
\label{eq:app-feasible-directional-derivative}
D\widehat F_s(\boldsymbol{\pi}_0)[\mathbf D]
=\langle\mathbf{C}^\text{GOT},\mathbf D\rangle.
\end{equation}
A negative value identifies a locally improving direction only when
that direction is feasible from $\boldsymbol{\pi}_0$. For a hard OT plan on the
boundary, an unconstrained negative-gradient direction may not be
feasible because it could require negative mass on a zero entry.

\subsection{Optimizer Response and Edge Scoring}
\label{sec:sup_theory_2}
The pre-update state $s$ contains the information used by the
training optimizer, such as momentum, second-moment estimates, and
step counters. We hold $s$ fixed when comparing candidate couplings.
This does not detach the dependence of the \emph{new} optimizer
moments on the supplied gradient. Moreover, $\mathbf{R}_0$ differentiates
only the current one-step update, not the preceding training history
or any future updates.

For SGD without momentum, $\mathcal A_s(\boldsymbol\theta,\mathbf g)
=\boldsymbol\theta-\eta\mathbf g$ and $\mathbf{R}_0=-\eta\mathbf I$.
For an adaptive optimizer, the response is generally different. For
example, consider an unclipped AdamW update with one parameter group,
without AMSGrad. For coordinate $p$ and update counter $k$, define
\begin{equation}
\label{eq:app-adam-moments}
\begin{gathered}
\widehat m_p=\frac{\beta_1m_p^-+(1-\beta_1)g_p}
{1-\beta_1^k},\\
\widehat v_p=\frac{\beta_2v_p^-+(1-\beta_2)g_p^2}
{1-\beta_2^k},\\
d_p=\sqrt{\widehat v_p}+\epsilon.
\end{gathered}
\end{equation}
Let $a_k=(1-\beta_1)/(1-\beta_1^k)$ and
$b_k=(1-\beta_2)/(1-\beta_2^k)$. The parameter update is
$\mathcal A_{s,p}=(1-\eta\lambda_{\mathrm{wd}})\theta_p
-\eta\widehat m_p/d_p$, with decoupled weight-decay coefficient
$\lambda_{\mathrm{wd}}$. For $\widehat v_p>0$,
direct differentiation gives
\begin{equation}
\label{eq:app-adam-response}
\frac{\partial\mathcal A_{s,p}}{\partial g_p}
=-\eta\biggl[
\frac{a_k}{d_p}
-\frac{\widehat m_p b_k g_p}
{\sqrt{\widehat v_p}\,d_p^2}
\biggr].
\end{equation}
This expression retains the dependence of the new second moment on
$g_p$. Freezing the adaptive denominator yields a different
approximation. Decoupled weight decay changes the virtual parameter
location and thus the evaluated reference gradient, even though it
does not add a direct $D_{\mathbf g}$ term when the pre-update
parameters are fixed. An Adam~\cite{kingma2014adam} update with coupled weight decay must
be handled according to its actual update rule.

If gradient clipping is active, the derivative must also include the
clipping map wherever the implemented map is differentiable. At a
clipping boundary, a single ordinary Jacobian may not describe all
feasible perturbation directions. The formula above is an
illustration for the stated unclipped AdamW assumptions. It does not
assert that every experiment used these optimizer settings.

We compute the shared parameter-space direction
\begin{equation}
\label{eq:app-shared-direction}
\mathbf{h}_\text{GOT}=\mathbf{R}_0^\top\mathbf{g}_\text{ref}^+.
\end{equation}
For each required edge, its score is then the directional derivative
\begin{equation}
\label{eq:app-edge-jvp}
C_{ij}^\text{GOT}
=\mathbf g_{ij}^\top\mathbf{h}_\text{GOT}
=D_{\boldsymbol\theta}\ell_{ij}(\boldsymbol\theta)
[\mathbf{h}_\text{GOT}].
\end{equation}
This computation can be expressed using a vector--Jacobian product
through the optimizer map followed by pair-loss directional
derivatives. In this formulation, we need not store every pair's
full parameter-gradient vector. The common direction is reused
across candidates within one selection event, but it must be
recomputed as the model or optimizer state changes.

\subsection{Local Approximation Error and Improvement Conditions}
\label{sec:sup_theory_3}
The exact derivative provides a linear approximation for finite
coupling changes. Let
\begin{equation}
\label{eq:app-gradient-perturbation}
\begin{gathered}
\mathbf d_{\boldsymbol{\pi}}=\mathbf g_{\boldsymbol{\pi}}-\mathbf g_0,\\
q_s(\mathbf g)=\widehat J_{\mathrm{ref}}
\bigl(\mathcal A_s(\boldsymbol\theta,\mathbf g)\bigr).
\end{gathered}
\end{equation}
Suppose $q_s$ has an $L_s$-Lipschitz gradient along the segment
between $\mathbf g_0$ and $\mathbf g_{\boldsymbol{\pi}}$. The standard
first-order remainder bound gives
\begin{equation}
\label{eq:app-taylor-remainder}
\begin{gathered}
\begin{aligned}
\widehat F_s(\boldsymbol{\pi})
&=\widehat F_s(\boldsymbol{\pi}_0)\\
&\quad{}+\langle\mathbf{C}^\text{GOT},\boldsymbol{\pi}-\boldsymbol{\pi}_0\rangle
+r_s(\boldsymbol{\pi}),
\end{aligned}\\[2pt]
|r_s(\boldsymbol{\pi})|\leq\frac{L_s}{2}\|\mathbf d_{\boldsymbol{\pi}}\|_2^2.
\end{gathered}
\end{equation}
We define the linearized advantage over the baseline as
\begin{equation}
\label{eq:app-predicted-gain}
\widehat\Delta_\text{GOT}(\boldsymbol{\pi};\boldsymbol{\pi}_0)
=-\langle\mathbf{C}^\text{GOT},\boldsymbol{\pi}-\boldsymbol{\pi}_0\rangle.
\end{equation}
It follows that
\begin{equation}
\label{eq:app-one-step-improvement-bound}
\begin{aligned}
\widehat F_s(\boldsymbol{\pi}_0)-\widehat F_s(\boldsymbol{\pi})
&\geq\widehat\Delta_\text{GOT}(\boldsymbol{\pi};\boldsymbol{\pi}_0)\\
&\quad{}-\frac{L_s}{2}\|\mathbf d_{\boldsymbol{\pi}}\|_2^2.
\end{aligned}
\end{equation}
Therefore, a positive predicted gain alone is insufficient for an
actual one-step improvement: a sufficient condition is that the
predicted margin exceeds the remainder term. This is a conditional
bound relative to the baseline update, not a guarantee of absolute
improvement over the original model. No numerical bound on $L_s$ is
assumed to have been established for the image experiments.

If $\|\mathbf g_{ij}\|_2\leq B_g$ for all relevant pairs, then
\begin{equation}
\label{eq:app-l1-gradient-bound}
\|\mathbf d_{\boldsymbol{\pi}}\|_2
\leq B_g\|\boldsymbol{\pi}-\boldsymbol{\pi}_0\|_1.
\end{equation}
Here, $\|\boldsymbol{\pi}-\boldsymbol{\pi}_0\|_1
=\sum_{i,j}|\pi_{ij}-(\boldsymbol{\pi}_0)_{ij}|$
is the entrywise $\ell_1$ distance between the coupling weights.
This bound requires the stated gradient-norm control. A small
geometric transport-cost gap does not by itself imply a small
gradient-space perturbation. Likewise,
clipping boundaries or a highly sensitive optimizer response can
invalidate the smoothness assumptions needed above.

Finally, fixed marginals can make parts of an edge cost irrelevant
to the coupling decision. If
$C'_{ij}=C_{ij}+\alpha_i+\beta_j$, then for every
$\boldsymbol{\pi}\in\Pi(\mathbf a,\mathbf b)$,
\begin{equation}
\label{eq:app-row-column-invariance}
\langle\mathbf C'-\mathbf C,\boldsymbol{\pi}\rangle
=\sum_i\alpha_i a_i+\sum_j\beta_j b_j,
\end{equation}
which is constant across feasible plans. Thus, different raw edge
scores need not imply different coupling decisions. The relevant
signal is the score difference along feasible transport contrasts,
not the magnitude of individual edge costs.

\subsection{Fixed-Marginal Structure of Gradient-Aware Coupling}
\label{sec:sup_theory_marginal}

We analyze which parts of the gradient-aware cost can affect
coupling selection under fixed marginals. Throughout this
subsection, the model parameters, pre-update optimizer state,
endpoint samples, reference evaluation, and pairwise scoring
randomness are fixed. We use the exact local cost
$C_{ij}^{\mathrm{GOT}}=\mathbf h_{\mathrm{GOT}}^\top\mathbf g_{ij}$
from Eqs.~\eqref{eq:app-exact-edge-derivative}
and~\eqref{eq:app-shared-direction}.

\paragraph{Feasible transport contrasts.}
Eq.~\eqref{eq:app-row-column-invariance} shows that
source-only and target-only cost terms cancel in comparisons
between plans with the same marginals. To illustrate the
remaining signal, let $\mathbf E_{ij}\in\mathbb R^{n\times m}$
have a unit entry at $(i,j)$ and zeros elsewhere. For distinct
source indices $i,k$, distinct target indices $j,l$, and $\delta>0$,
define the $2\times2$ exchange
\begin{equation}
\label{eq:app_transport_exchange}
\mathbf D_\delta
=\delta(\mathbf E_{ij}+\mathbf E_{kl}
-\mathbf E_{il}-\mathbf E_{kj}).
\end{equation}
Both row and column sums of $\mathbf D_\delta$ are zero, so
$\boldsymbol\pi_0+\mathbf D_\delta$ preserves the marginals.
Nonnegativity additionally requires
$\delta\leq\min\{\pi_{0,il},\pi_{0,kj}\}$, and the exchange
must satisfy any support, transport-cost, and movement constraints.
Its exact change in the linearized objective is
\begin{equation}
\label{eq:app_transport_contrast}
\begin{aligned}
\langle\mathbf C^{\mathrm{GOT}},\mathbf D_\delta\rangle_F
&=\delta(C_{ij}^{\mathrm{GOT}}+C_{kl}^{\mathrm{GOT}}
-C_{il}^{\mathrm{GOT}}-C_{kj}^{\mathrm{GOT}})\\
&=\delta\mathbf h_{\mathrm{GOT}}^\top
(\mathbf g_{ij}+\mathbf g_{kl}-\mathbf g_{il}-\mathbf g_{kj}).
\end{aligned}
\end{equation}
A negative value predicts an improvement over the baseline
update, and its negative is the predicted gain. Thus, individual edge
costs affect selection through marginal-preserving gradient
contrasts. The finite-change prediction remains subject to the
remainder bound in Eq.~\eqref{eq:app-taylor-remainder}.
On complete support, all $2\times2$ cost contrasts vanish if and
only if $C_{ij}=u_i+v_j$: setting one row and column as anchors
gives $C_{ij}=C_{i1}+C_{1j}-C_{11}$.
Restricted supports may require longer alternating cycles to
describe all marginal-preserving contrasts.

\paragraph{Gradient components that affect coupling changes.}
Consider any decomposition
\begin{equation}
\label{eq:app_gradient_decomposition}
\mathbf g_{ij}=\mathbf u_i^g+\mathbf v_j^g+\mathbf r_{ij},
\end{equation}
where $\mathbf u_i^g$ and $\mathbf v_j^g$ depend only on the
source and target, respectively, and $\mathbf r_{ij}$ is the
remaining component. This decomposition need not be unique.
For $\boldsymbol\pi\in\Pi(\mathbf a,\mathbf b)$,
\begin{equation}
\label{eq:app_gradient_decomposition_sum}
\mathbf g_{\boldsymbol\pi}
=\sum_i a_i\mathbf u_i^g+\sum_j b_j\mathbf v_j^g
+\sum_{i,j}\pi_{ij}\mathbf r_{ij}.
\end{equation}
Consequently,
\begin{equation}
\label{eq:app_residual_gradient_difference}
\mathbf g_{\boldsymbol\pi}-\mathbf g_0
=\sum_{i,j}(\pi_{ij}-\pi_{0,ij})\mathbf r_{ij}.
\end{equation}
Source-only and target-only gradient components cancel in
coupling differences, but remain part of the baseline update.
The shared sensitivity $\mathbf h_{\mathrm{GOT}}$ must therefore
still be computed from the full baseline gradient $\mathbf g_0$.
If two couplings induce the same aggregate gradient, their
deterministic one-step optimizer updates and reference losses
are identical under the fixed evaluation conditions. Equal
linearized GOT scores alone do not imply equal finite-change
learning utilities.

\paragraph{Learning signal beyond geometry and prediction error.}
Let $G_{ij}=\|\mathbf y_j-\mathbf z_i\|_2^2$ be the geometric
cost and $M_{ij}=\ell_{ij}(\boldsymbol\theta)$ the current FM
prediction-error cost under a fixed time-evaluation rule for the
per-pair loss in Eq.~\eqref{eq:time-averaged-fm-loss}. Here, $\mathbf M$ is a
theoretical comparison criterion that need not coincide with the
two-endpoint proposal cost $\mathbf E_{\boldsymbol\theta}$ defined
in Eq.~\eqref{eq:app-endpoint-error-proposal}.
Define the linear space of perturbations preserving both
marginals and these two costs:
\begin{equation}
\label{eq:app_cost_preserving_space}
\mathcal L_{G,M}
=\left\{\mathbf D\in\mathbb R^{n\times m}:\begin{aligned}
&\mathbf D\mathbf1_m=\mathbf0,\quad
\mathbf D^\top\mathbf1_n=\mathbf0,\\
&\langle\mathbf G,\mathbf D\rangle_F=0,\\
&\langle\mathbf M,\mathbf D\rangle_F=0
\end{aligned}\right\}.
\end{equation}
For restricted support, also require $D_{ij}=0$ on disallowed
edges. Let $P_{\mathcal L}$ denote orthogonal projection onto a
linear space $\mathcal L$ under the Frobenius inner product, and set
\begin{equation}
\label{eq:app_residual_got_cost}
\mathbf C_{\mathrm{res}}
=P_{\mathcal L_{G,M}}\mathbf C^{\mathrm{GOT}}.
\end{equation}
This projection characterizes the cost for analysis and is not an
additional step in GOT-FM's selection procedure.

\noindent\textbf{Proposition 1 (Additional first-order learning signal).}
Suppose the differentiability assumptions of
Appendix~\ref{sec:sup_theory_1} hold. Then
$\mathbf C_{\mathrm{res}}\neq\mathbf0$ if and only if some
$\mathbf D\in\mathcal L_{G,M}$ has
$D\widehat F_s(\boldsymbol\pi_0)[\mathbf D]\neq0$.
If $\boldsymbol\pi_0-\tau\mathbf C_{\mathrm{res}}$ is also
feasible for every sufficiently small $\tau>0$, this direction
strictly decreases the post-update reference loss for such steps.

\noindent\textit{Proof.}
For any $\mathbf D\in\mathcal L_{G,M}$, orthogonal projection gives
\begin{equation}
\begin{aligned}
D\widehat F_s(\boldsymbol\pi_0)[\mathbf D]
&=\langle\mathbf C^{\mathrm{GOT}},\mathbf D\rangle_F\\
&=\langle\mathbf C_{\mathrm{res}},\mathbf D\rangle_F.
\end{aligned}
\end{equation}
The derivative vanishes for all such directions exactly when
$\mathbf C_{\mathrm{res}}=\mathbf0$. Otherwise,
$\mathbf D=-\mathbf C_{\mathrm{res}}$ gives derivative
$-\|\mathbf C_{\mathrm{res}}\|_F^2<0$.
Differentiability implies a strict decrease along sufficiently
small positive steps whenever those steps are feasible.
\hfill$\square$

The equivalence concerns the equality constraints, not the full
feasible set. At a sparse Euclidean OT baseline, nonnegativity
can exclude the projected direction. A nonzero residual cost
therefore does not by itself guarantee a feasible improvement.

For a quantitative finite-change bound, define the gradient map
\begin{equation}
\label{eq:app_gradient_map}
\mathcal G(\mathbf D)=\sum_{i,j}D_{ij}\mathbf g_{ij}.
\end{equation}
Using $q_s$ and the $L_s$-smoothness assumption from
Appendix~\ref{sec:sup_theory_3}, Eq.~\eqref{eq:app-taylor-remainder}
implies, for a feasible step $\tau>0$,
\begin{equation}
\label{eq:app_residual_improvement_bound}
\begin{aligned}
&\widehat F_s(\boldsymbol\pi_0-\tau\mathbf C_{\mathrm{res}})
-\widehat F_s(\boldsymbol\pi_0)\\
&\quad\leq-\tau\|\mathbf C_{\mathrm{res}}\|_F^2
+\frac{L_s\tau^2}{2}\|\mathcal G(\mathbf C_{\mathrm{res}})\|_2^2.
\end{aligned}
\end{equation}
Thus, a nonzero residual yields a strict decrease when the step
is feasible and the linear term dominates the remainder.
For complete support with only the equalities in
Eq.~\eqref{eq:app_cost_preserving_space},
$\mathbf C_{\mathrm{res}}=\mathbf0$ if and only if
\begin{equation}
\label{eq:app_residual_cost_characterization}
\mathbf C^{\mathrm{GOT}}
=\mathbf u\mathbf1_m^\top+\mathbf1_n\mathbf v^\top
+\lambda\mathbf G+\mu\mathbf M
\end{equation}
for some $\mathbf u\in\mathbb R^n$, $\mathbf v\in\mathbb R^m$,
and $\lambda,\mu\in\mathbb R$. This follows because the
orthogonal complement of $\mathcal L_{G,M}$ is the span of the
row, column, and two cost-constraint matrices. The residual
therefore identifies first-order information beyond additive
endpoint terms and linear combinations of the comparison costs.

\paragraph{Optimizer response and selection equivalence.}
For a nonempty feasible comparison set $\mathcal K$, define
\begin{equation}
\begin{aligned}
\mathcal L_{\mathcal K}
&=\operatorname{span}\{\boldsymbol\pi-\boldsymbol\pi':
\boldsymbol\pi,\boldsymbol\pi'\in\mathcal K\},\\
\mathcal V_{\mathcal K}&=\mathcal G(\mathcal L_{\mathcal K}).
\end{aligned}
\end{equation}
The space $\mathcal V_{\mathcal K}$ is the span of accessible
gradient differences. For any two plans in $\mathcal K$,
\begin{equation}
\label{eq:app_feasible_gradient_comparison}
\begin{aligned}
&\langle\mathbf C^{\mathrm{GOT}},\boldsymbol\pi-\boldsymbol\pi'\rangle_F\\
&=\langle P_{\mathcal L_{\mathcal K}}\mathbf C^{\mathrm{GOT}},
\boldsymbol\pi-\boldsymbol\pi'\rangle_F\\
&=\mathbf h_{\mathrm{GOT}}^\top
(\mathbf g_{\boldsymbol\pi}-\mathbf g_{\boldsymbol\pi'}).
\end{aligned}
\end{equation}
Hence the linearized score is constant over $\mathcal K$ exactly
when $\mathbf h_{\mathrm{GOT}}\perp\mathcal V_{\mathcal K}$.

\noindent\textbf{Proposition 2 (Selection equivalence of response directions).}
Let $\mathbf h_1$ and $\mathbf h_2$ use the same pairwise
gradients and comparison set $\mathcal K$. If some $c>0$ satisfies
\begin{equation}
\label{eq:app_response_equivalence}
\mathbf h_1-c\mathbf h_2\perp\mathcal V_{\mathcal K},
\end{equation}
the two directions induce identical linearized rankings and ties
over $\mathcal K$.

\noindent\textit{Proof.}
Every difference $\mathbf g_{\boldsymbol\pi}
-\mathbf g_{\boldsymbol\pi'}$ lies in $\mathcal V_{\mathcal K}$, so
\begin{equation}
\begin{aligned}
&\mathbf h_1^\top(\mathbf g_{\boldsymbol\pi}
-\mathbf g_{\boldsymbol\pi'})\\
&\quad=c\mathbf h_2^\top(\mathbf g_{\boldsymbol\pi}
-\mathbf g_{\boldsymbol\pi'}).
\end{aligned}
\end{equation}
Positive scaling preserves the signs and zeros of all pairwise
score differences. \hfill$\square$

A different optimizer response can therefore preserve the same
ranking. For a finite pool $\mathcal P^+\subseteq\mathcal K$,
$\mathcal V_{\mathcal P^+}\subseteq\mathcal V_{\mathcal K}$, so
some response differences that matter over the larger feasible
set may have no effect within the retained pool.

\paragraph{Stability under approximate scoring.}
Let $\mathbf C$ be the exact linearized cost and
$\widetilde{\mathbf C}$ an approximation, evaluated on the same
nonempty finite pool $\mathcal P^+$. Define the maximum contrast error
\begin{equation}
\label{eq:app_contrast_error}
\epsilon_{\mathcal P^+}
=\max_{\boldsymbol\pi,\boldsymbol\pi'\in\mathcal P^+}
\big|\langle\widetilde{\mathbf C}-\mathbf C,
\boldsymbol\pi-\boldsymbol\pi'\rangle_F\big|.
\end{equation}

\noindent\textbf{Proposition 3 (Linear selection regret).}
Let $\boldsymbol\pi^\star$ and $\widetilde{\boldsymbol\pi}$
minimize the costs $\mathbf C$ and $\widetilde{\mathbf C}$,
respectively, over $\mathcal P^+$. Then
\begin{equation}
\label{eq:app_linear_selection_regret}
0\leq\langle\mathbf C,
\widetilde{\boldsymbol\pi}-\boldsymbol\pi^\star\rangle_F
\leq\epsilon_{\mathcal P^+}.
\end{equation}

\noindent\textit{Proof.}
Optimality under $\widetilde{\mathbf C}$ gives
$\langle\widetilde{\mathbf C},
\widetilde{\boldsymbol\pi}-\boldsymbol\pi^\star\rangle_F\leq0$.
Therefore,
\begin{equation}
\begin{aligned}
&\langle\mathbf C,
\widetilde{\boldsymbol\pi}-\boldsymbol\pi^\star\rangle_F\\
&\quad\leq\langle\mathbf C-\widetilde{\mathbf C},
\widetilde{\boldsymbol\pi}-\boldsymbol\pi^\star\rangle_F
\leq\epsilon_{\mathcal P^+}.
\end{aligned}
\end{equation}
Optimality of $\boldsymbol\pi^\star$ under $\mathbf C$ gives
the lower bound. \hfill$\square$

If $\mathcal P^+$ has at least two distinct plans and the exact
minimizer is unique, define its score gap
\begin{equation}
\gamma_{\mathcal P^+}
=\min_{\substack{\boldsymbol\pi\in\mathcal P^+\\
\boldsymbol\pi\neq\boldsymbol\pi^\star}}
\langle\mathbf C,\boldsymbol\pi-\boldsymbol\pi^\star\rangle_F.
\end{equation}
Then $\gamma_{\mathcal P^+}>\epsilon_{\mathcal P^+}$ implies
$\widetilde{\boldsymbol\pi}=\boldsymbol\pi^\star$.
The contrast error captures only cost errors that affect plan
comparisons because additive endpoint errors cancel. This bound concerns
the linearized selection objective, not the actual nonlinear
post-update reference loss.

\paragraph{Exact reuse of shared scoring contributions.}
For the baseline $\boldsymbol\pi_0$ and a fixed candidate pool
$\mathcal P^+$, define
\begin{equation}
\label{eq:app_changed_edge_support}
\mathcal E_\Delta
=\{(i,j):\exists\boldsymbol\pi\in\mathcal P^+,
\ \pi_{ij}\neq\pi_{0,ij}\}.
\end{equation}
For every candidate,
\begin{equation}
\label{eq:app_changed_edge_score}
\begin{aligned}
&\langle\mathbf C^{\mathrm{GOT}},
\boldsymbol\pi-\boldsymbol\pi_0\rangle_F\\
&\quad=\sum_{(i,j)\in\mathcal E_\Delta}
(\pi_{ij}-\pi_{0,ij})C_{ij}^{\mathrm{GOT}}.
\end{aligned}
\end{equation}
Edges whose weights are unchanged across all candidates need no
additional directional-derivative scoring for these relative
scores. The full baseline gradient and shared optimizer-conditioned
sensitivity are still required. This identity suggests an
implementation opportunity but does not establish an end-to-end
runtime improvement for the reported experiments.

%\paragraph{Scope and limitations.}
%The residual-cost argument requires differentiability, smoothness,
%and feasibility assumptions that need not hold for hard
%assignments, restricted finite pools, or exact movement constraints.
%We do not establish numerical smoothness constants for the image
%experiments. A decrease relative to the baseline coupling's
%virtual update does not imply improvement over the pre-update
%model, subsequent training updates, or generation FID.
%The contrast and regret results describe the linear scores used
%to select the final coupling in
%Appendix~\ref{sec:sup_coupling_3}. They introduce no candidate-wise
%nonlinear evaluation or acceptance step.

\subsection{One-Step Utility and Finite-Horizon Limitations}
\label{sec:sup_theory_4}
The optimizer-conditioned transport cost is defined for one update
at a specified model and pre-update optimizer state. The scoring
gradient may be computed with fixed time nodes or dropout, whereas
the actual model update can use newly sampled training randomness.
In general,
\begin{equation}
\label{eq:app-stochastic-update-mismatch}
\begin{gathered}
\widehat J_{\mathrm{ref}}
\bigl(\mathcal A_s(\boldsymbol\theta,
\mathbb E_\xi[\mathbf g_{\boldsymbol{\pi}}(\xi)])\bigr)\\
\neq\mathbb E_\xi\!\left[
\widehat J_{\mathrm{ref}}
\bigl(\mathcal A_s(\boldsymbol\theta,
\mathbf g_{\boldsymbol{\pi}}(\xi))\bigr)\right]
\end{gathered}
\end{equation}
because the optimizer update and reference evaluation
are nonlinear in the gradient. Accordingly, linearized prediction,
exact virtual $J_\text{ref}$ gain, executed-update gain, and independent evaluation
gain should not be treated as the same quantity.

For repeated training, the full state consists of the model
parameters and optimizer state. After two different first-step
couplings, both states can differ even if training later reverts
to Euclidean OT. Subsequent gradients and optimizer responses are
therefore evaluated along different trajectories. A favorable
one-step reference improvement need not persist for multiple updates.
In particular, a repeated sparse policy cannot be evaluated by
multiplying the effect of one intervention by the number of
interventions. Generation quality, including FID at a given NFE, is
an additional downstream measurement and is not implied by the
one-step bound in Eq.~\eqref{eq:app-one-step-improvement-bound}.

\section{Coupling Construction and Selection}
\label{sec:sup_coupling}
\subsection{Candidate Generation and Proposal Costs}
\label{sec:sup_coupling_1}
We use the Euclidean OT coupling of the current minibatch as the
baseline $\boldsymbol{\pi}_0$, computed using SciPy's
\texttt{linear\_sum\_assignment}. For the sampled endpoints, define the
geometric edge cost $G_{ij}=\|\mathbf y_j-\mathbf z_i\|_2^2$. For the
model-dependent proposal, a fixed endpoint prediction-error cost
can be written as
\begin{equation}
\label{eq:app-endpoint-error-proposal}
\begin{gathered}
(\mathbf E_{\boldsymbol\theta})_{ij}
=\tfrac12[e_{ij}(0)+e_{ij}(1)],\\
e_{ij}(t)=\tfrac1d\left\|
\mathbf v_{\boldsymbol\theta}(\mathbf x_{ij}^t,t)
-\mathbf u_{ij}\right\|_2^2.
\end{gathered}
\end{equation}
This definition averages the FM prediction MSE at the two time
endpoints. It is a proposal criterion, not the optimizer-conditioned
cost $C_{ij}^\text{GOT}$ used for final selection.

In the default eight-alternative configuration, we construct four
geometry/endpoint-based (G) proposals and four random-cost (R) proposals.
The baseline $\boldsymbol\pi_0$ is computed separately
and is added to the final selection pool, but is not counted
among these eight alternatives.
For each proposal cost matrix $\mathbf Q$, we solve a linear program
over the same feasible set:
\begin{equation}
\label{eq:app-proposal-objective}
\boldsymbol{\pi}^{(\mathbf Q)}\in
\operatorname*{arg\,min}_{\boldsymbol{\pi}\in\mathcal K_{\mathrm{proposal}}}
\langle\mathbf Q,\boldsymbol{\pi}\rangle,
\end{equation}
where $\mathcal K_{\mathrm{proposal}}$ denotes the feasible set for
proposal generation, including the prescribed marginals and the
two-sided budgets in Eq.~\eqref{eq:app-budget-constraints}.
We solve these LPs using SciPy's
\texttt{linprog(method='highs')}, with both primal and dual feasibility
tolerances set to $10^{-8}$.

We construct an ordered geometry/endpoint-based proposal library,
starting with the endpoint-cost solution obtained using
$\mathbf Q=\mathbf E_{\boldsymbol\theta}$ and then solving LPs with
the following sequence of cost matrices:
\begin{equation}
\label{eq:app-geometry-endpoint-proposals}
\begin{gathered}
\mathbf Q_k=\cos\phi_k\,\frac{\mathbf G}{s_G}
+\sin\phi_k\,\frac{\mathbf E_{\boldsymbol\theta}}{s_E},\\
\phi_k=\frac{2\pi k}{32},\qquad k=0,\ldots,31,
\end{gathered}
\end{equation}
where $s_G$ and $s_E$ are the fixed calibration scales described below.
We remove duplicate couplings while preserving their order and use
the first four distinct feasible solutions. In the evaluated
$4\mathrm{G}+4\mathrm{R}$ configuration, these comprise one endpoint-cost
proposal, one geometry-cost proposal, and two proposals based on linear
combinations of the normalized geometry and endpoint costs.

Each of the four random-cost proposals uses an independently sampled
cost matrix $\mathbf Q$ whose entries follow $\mathcal N(0,1)$ independently.
Proposal generation determines which feasible plans are considered.
Gradient-aware scoring is applied after these plans have been constructed.

Let $\mathcal P^+$ denote the final selection pool, comprising
the baseline and the distinct, feasible proposals:
\begin{equation}
\label{eq:app-finite-pool}
\mathcal P^+=\{\boldsymbol{\pi}_0\}\cup
\{\boldsymbol{\pi}^{(1)},\ldots,\boldsymbol{\pi}^{(K)}\}\subseteq\mathcal K.
\end{equation}
The configured proposal count excludes the baseline. Generating the
proposals may require additional LP solves because different objectives
can yield duplicate couplings. The retained proposal count $K$ can be
smaller if fewer distinct feasible proposals are available, and
$|\mathcal P^+|\le K+1$ because a proposal may coincide with the baseline.
The baseline remains available even if no alternative is retained.

\subsection{Marginal and Cost Constraints}
\label{sec:sup_coupling_2}
Every retained candidate satisfies the prescribed empirical
marginals. The geometry and endpoint-error budgets use fixed
calibration scales. At initialization, we construct baseline Euclidean
OT couplings on three calibration minibatches and evaluate their
geometric and endpoint-error costs using the initial model parameters.
We fix the medians of these costs as $s_G$ and $s_E$, respectively,
and keep both scales unchanged throughout training.
For each subsequent minibatch, every alternative coupling
$\boldsymbol\pi$ must satisfy
\begin{equation}
\label{eq:app-budget-constraints}
\begin{gathered}
\left|\left\langle\boldsymbol\pi-\boldsymbol\pi_0,
\mathbf G\right\rangle\right|\le 0.01s_G,\\
\left|\left\langle\boldsymbol\pi-\boldsymbol\pi_0,
\mathbf E_{\boldsymbol\theta}\right\rangle\right|\le 0.01s_E.
\end{gathered}
\end{equation}
Here, $\boldsymbol\pi_0$, $\mathbf G$, and
$\mathbf E_{\boldsymbol\theta}$ are evaluated on the current minibatch.
The baseline itself satisfies both budgets with zero deviation and
preserves the same marginals and admissible edges.
In exact arithmetic, $\boldsymbol\pi_0$ minimizes geometric cost
under the same marginals and admissible edges, so the geometric
cost cannot decrease below the baseline. Its budget therefore
bounds the excess cost above this optimum. The endpoint-error
constraint is two-sided and permits either an increase or a decrease
relative to the baseline within its calibrated budget.
The reported $1\%$ budgets therefore refer to $s_G$ and $s_E$,
not to the current minibatch's baseline costs.
We validate each generated proposal against the budget inequalities with an
additional absolute numerical tolerance of $10^{-7}$ on each
right-hand side. The LP feasibility tolerances and this candidate-validation
tolerance are distinct.

\subsection{Candidate Selection}
\label{sec:sup_coupling_3}
For a finite feasible pool $\mathcal P^+$, GOT-FM scores each plan
with the local optimizer-conditioned transport cost. The selected
coupling is
\begin{equation}
\label{eq:app-pool-selection}
\boldsymbol{\pi}_{\mathrm{sel}}\in
\operatorname*{arg\,min}_{\boldsymbol{\pi}\in\mathcal P^+}
\langle\mathbf{C}^\text{GOT},\boldsymbol{\pi}\rangle.
\end{equation}
Because $\langle\mathbf{C}^\text{GOT},\boldsymbol{\pi}_0\rangle$ is independent of the
candidate, this is equivalent to minimizing the relative score
\begin{equation}
\label{eq:app-relative-score}
S_\text{GOT}(\boldsymbol{\pi})
=\langle\mathbf{C}^\text{GOT},\boldsymbol{\pi}-\boldsymbol{\pi}_0\rangle.
\end{equation}
A negative $S_\text{GOT}(\boldsymbol{\pi})$ indicates a predicted improvement in the
fixed one-step reference evaluation over the baseline update.
Since $S_\text{GOT}(\boldsymbol{\pi}_0)=0$ and the baseline is included,
\begin{equation}
\label{eq:app-baseline-nonnegative-gain}
\widehat\Delta_\text{GOT}
(\boldsymbol{\pi}_{\mathrm{sel}};\boldsymbol{\pi}_0)
=-S_\text{GOT}(\boldsymbol{\pi}_{\mathrm{sel}})\ge0.
\end{equation}
We break ties in favor of $\boldsymbol{\pi}_0$. Thus, if no alternative
has a strictly negative relative score, the baseline is selected.
This guarantee concerns the linearized score only; actual one-step
improvement still requires controlling the approximation remainder
in Eq.~\eqref{eq:app-one-step-improvement-bound}.

At each designated selection step, GOT-FM selects the feasible
candidate with the lowest linearized transport cost and uses that
coupling for the model update, including a baseline update when
the baseline is selected. Candidate-wise nonlinear
virtual-update loss comparisons are used only for the learning-utility
diagnostics and oracle comparisons in
Appendices~\ref{sec:sup_ablation_utility} and~\ref{sec:sup_objective_3}.

For sparse training, at update index $r$ the method uses
$\boldsymbol{\pi}_{\mathrm{sel},r}$ at designated selection steps and the
current minibatch's $\boldsymbol{\pi}_{0,r}$ otherwise. The actual FM update
starts from the original current model and optimizer state. The
virtual scoring update is not committed. The selected coupling is
not carried over to the next minibatch.

\paragraph{Weighted support-edge updates.}
For Rectified Flow and conditional ImageNet, a fractional plan
is applied by evaluating the weighted loss over its support:
\begin{equation}
\label{eq:app-weighted-support-update}
\widehat{\mathcal L}(\boldsymbol\theta;\boldsymbol\pi)
=\sum_{(i,j):\,\pi_{ij}>10^{-14}}
\pi_{ij}\,\ell_{ij}(\boldsymbol\theta,t_i).
\end{equation}
Here, $\ell_{ij}(\boldsymbol\theta,t_i)$ is the instantaneous
FM loss, including the target class condition for ImageNet.
We draw a fresh uniform time $t_i$ for each source and share it
across all support edges connected to that source.
Support edges are processed in microbatches, and their weighted
loss gradients are accumulated. After all support microbatches
have contributed, we apply gradient clipping and perform exactly
one optimizer step. We neither resample pairs from the plan nor
round it to a permutation. Marginal preservation refers to the
coupling weights, up to the numerical support threshold.
Conditional on the sampled endpoints, update randomness arises
from the fresh times and the model's training randomness.

\subsection{Relationship to Direct OT Optimization}
\label{sec:sup_relation_to_direct}
\label{sec:sup_coupling_4}
The local gradient-aware cost also defines a direct linearized transport
problem over the full feasible set. To distinguish three search
spaces, let
\begin{equation}
\label{eq:app-three-search-spaces}
\begin{gathered}
\boldsymbol{\pi}_D\in\operatorname*{arg\,min}_{\boldsymbol{\pi}\in\mathcal K}
\langle\mathbf{C}^\text{GOT},\boldsymbol{\pi}\rangle,\\
\boldsymbol{\pi}_S\in\operatorname*{arg\,min}_{\boldsymbol{\pi}\in\mathcal K_{\Omega_{\mathcal P^+}}}
\langle\mathbf{C}^\text{GOT},\boldsymbol{\pi}\rangle,\\
\boldsymbol{\pi}_{\mathrm{sel}}\in\operatorname*{arg\,min}_{\boldsymbol{\pi}\in\mathcal P^+}
\langle\mathbf{C}^\text{GOT},\boldsymbol{\pi}\rangle,
\end{gathered}
\end{equation}
where
$\Omega_{\mathcal P^+}=\bigcup_{\boldsymbol{\pi}\in\mathcal P^+}\operatorname{supp}(\boldsymbol{\pi})$
and
$\mathcal K_{\Omega_{\mathcal P^+}}=
\{\boldsymbol{\pi}\in\mathcal K:\operatorname{supp}(\boldsymbol{\pi})
\subseteq\Omega_{\mathcal P^+}\}$.
The inclusions
$\mathcal P^+\subseteq\mathcal K_{\Omega_{\mathcal P^+}}
\subseteq\mathcal K$ give
\begin{equation}
\label{eq:app-search-order}
\begin{aligned}
\min_{\boldsymbol{\pi}\in\mathcal K}\langle\mathbf{C}^\text{GOT},\boldsymbol{\pi}\rangle
&\le\min_{\boldsymbol{\pi}\in\mathcal K_{\Omega_{\mathcal P^+}}}
\langle\mathbf{C}^\text{GOT},\boldsymbol{\pi}\rangle\\
&\le\min_{\boldsymbol{\pi}\in\mathcal P^+}
\langle\mathbf{C}^\text{GOT},\boldsymbol{\pi}\rangle.
\end{aligned}
\end{equation}
Direct optimization can construct a new feasible coupling over the
full allowed graph. A support-restricted LP can recombine edges from
the candidate support. Finite-pool selection returns the supplied plan
$\boldsymbol\pi_{\mathrm{sel}}$ with the lowest linearized cost.
Evaluating costs on the support union is
not the same operation as resolving an LP on that graph.

The difference between finite-pool selection and direct linearized
optimization can be expressed as a candidate-coverage gap,
\begin{equation}
\label{eq:app-pool-coverage}
\begin{aligned}
\delta_{\mathrm{cov}}
&=\min_{\boldsymbol{\pi}\in\mathcal P^+}\langle\mathbf{C}^\text{GOT},\boldsymbol{\pi}\rangle\\
&\quad{}-\min_{\boldsymbol{\pi}\in\mathcal K}\langle\mathbf{C}^\text{GOT},\boldsymbol{\pi}\rangle
\ge0.
\end{aligned}
\end{equation}
The two searches have equal optimal linearized objective values
only if the pool contains at least one direct optimizer. Equal
objective values do not require identical plans when minimizers are
not unique. No small-coverage-gap guarantee follows from fixed
marginals or from scoring all candidate edges. This analysis
concerns the linearized objective, not the global optimum of the
nonlinear one-step reference loss.

\section{Extension to MeanFlow}
\label{sec:sup_meanflow}

We further extend GOT-FM to MeanFlow~\cite{geng2026mean},
which learns average velocity over a time interval
rather than instantaneous velocity.
The coupling-selection framework remains unchanged,
but we adapt the pairwise training gradients and
reference objective to the MeanFlow loss.

\paragraph{MeanFlow training objective.}
Let $\mathbf n_i$ and $\mathbf x_j$ denote a noise
sample and a data sample, respectively.
Unlike our Rectified Flow setup, MeanFlow uses
the data-to-noise interpolation
\begin{equation}
\begin{gathered}
\mathbf z_{ij}^t
=(1-t)\mathbf x_j+t\mathbf n_i,\\
\mathbf v_{ij}=\mathbf n_i-\mathbf x_j.
\end{gathered}
\end{equation}
The model
$\mathbf u_{\boldsymbol\theta}
(\mathbf z_{ij}^t,r,t)$
predicts the average velocity over $[r,t]$,
where $0\leq r\leq t\leq1$.

Following MeanFlow, we construct the training
target using a Jacobian-vector product (JVP):
\begin{equation}
\bar{\mathbf y}_{ij}
=\operatorname{sg}\Big[
\mathbf v_{ij}
-(t-r)\big(
\partial_{\mathbf z}\mathbf u_{\boldsymbol\theta}
\,\mathbf v_{ij}
+\partial_t\mathbf u_{\boldsymbol\theta}
\big)
\Big],
\end{equation}
where the derivatives are evaluated at
$(\mathbf z_{ij}^t,r,t)$, holding $r$ fixed.
The operator $\operatorname{sg}$ denotes
stop-gradient.

We use the adaptive MeanFlow loss
\begin{equation}
\begin{gathered}
\ell_{ij}^{\mathrm{MF}}(\boldsymbol\theta)
=\bar w_{ij}
\left\|
\mathbf u_{\boldsymbol\theta}
(\mathbf z_{ij}^t,r,t)
-\bar{\mathbf y}_{ij}
\right\|_2^2,\\
\bar w_{ij}
=\operatorname{sg}\left[
\left(
\left\|
\mathbf u_{\boldsymbol\theta}
-\bar{\mathbf y}_{ij}
\right\|_2^2+10^{-3}
\right)^{-0.75}
\right].
\end{gathered}
\end{equation}
Both the target and adaptive weight are
detached during gradient computation.
When $r=t$, the JVP correction vanishes
and the target reduces to the instantaneous
velocity $\mathbf v_{ij}$.

\paragraph{Optimizer-aware coupling cost.}
We retain the Euclidean OT baseline
$\boldsymbol\pi_0$ and the finite-pool
coupling-selection procedure of GOT-FM.
The pairwise gradients are now computed
from the stopped-target MeanFlow loss:
\begin{equation}
\begin{gathered}
\mathbf g_{ij}^{\mathrm{MF}}
=\nabla_{\boldsymbol\theta}
\ell_{ij}^{\mathrm{MF}}(\boldsymbol\theta),\\
\mathbf g_{\boldsymbol\pi}^{\mathrm{MF}}
=\sum_{i,j}\pi_{ij}
\mathbf g_{ij}^{\mathrm{MF}}.
\end{gathered}
\end{equation}
The target and weight are fixed when
evaluating these gradients.

As in the main method, we first compute
a virtual optimizer update from the baseline
coupling:
\begin{equation}
\boldsymbol\theta_0^+
=
\mathcal A_s
(\boldsymbol\theta,
\mathbf g_{\boldsymbol\pi_0}^{\mathrm{MF}}).
\end{equation}
Using the optimizer response $\mathbf R_0$
and the gradient of a shared fixed-target
reference objective, we obtain
\begin{equation}
\begin{gathered}
\mathbf h_{\mathrm{GOT}}^{\mathrm{MF}}
=\mathbf R_0^\top
\mathbf g_{\mathrm{ref}}^{+,\mathrm{MF}},\\
C_{ij}^{\mathrm{GOT,MF}}
=
D_{\boldsymbol\theta}
\ell_{ij}^{\mathrm{MF}}
[\mathbf h_{\mathrm{GOT}}^{\mathrm{MF}}].
\end{gathered}
\end{equation}
The candidate's predicted gain relative
to the baseline is
\begin{equation}
\widehat\Delta_{\mathrm{GOT}}^{\mathrm{MF}}
(\boldsymbol\pi;\boldsymbol\pi_0)
=
-\left\langle
\boldsymbol\pi-\boldsymbol\pi_0,
\mathbf C^{\mathrm{GOT,MF}}
\right\rangle_F.
\end{equation}
Thus, GOT-FM scores candidate couplings
using their estimated effect on the
post-update MeanFlow reference objective.
The selected coupling follows Eq.~\eqref{eq:app-pool-selection}
with the MeanFlow cost $\mathbf C^{\mathrm{GOT,MF}}$.

This computation involves two distinct JVPs.
The first constructs the MeanFlow target
through derivatives with respect to model inputs.
The second evaluates directional derivatives
of the stopped-target pairwise loss with respect
to model parameters.
We do not differentiate through the target-construction JVP
when computing the coupling cost.

\paragraph{Fixed-target reference objective.}
Unlike the fixed velocity target in Rectified Flow,
the MeanFlow target and adaptive weight depend
on the model parameters.
To compute the shared reference gradient,
we use a fixed-target surrogate
within each coupling-selection event.

Training targets and weights are generated
at the current parameters $\boldsymbol\theta$
and shared across candidate couplings.
For reference-gradient evaluation, we instead generate
the targets and weights at the baseline
virtual-update parameters
$\boldsymbol\theta_0^+$ and keep them fixed
when differentiating $J_\text{ref}$.
For diagnostic virtual-update comparisons, the same fixed-target
reference objective is used across candidate updates.

Importantly, the resulting $J_\text{ref}$ objective
is a fixed-target MeanFlow surrogate.
It does not differentiate through targets
regenerated at each candidate's updated
parameters and does not directly optimize
the final generation quality.

\paragraph{Time sampling and candidate construction.}
MeanFlow uses joint sampling of $(r,t)$
instead of a single flow time.
We sort two logit-normal samples to obtain
$r\leq t$, using off-diagonal intervals
for 75\% of the samples and setting $r=t$
for the remaining 25\%.
During coupling scoring, three common
$(r,t)$ samples per noise sample are used
across candidates.
The actual model update uses newly
sampled times.

We also adapt the endpoint-error proposal
to MeanFlow's full-interval prediction:
\begin{equation}
E_{ij}^{\mathrm{MF}}
=
\frac1d
\left\|
\mathbf n_i
-\mathbf u_{\boldsymbol\theta}
(\mathbf n_i,0,1)
-\mathbf x_j
\right\|_2^2.
\end{equation}
The corresponding endpoint-error budget
is recalibrated for this cost.
Candidate generation continues to preserve
both empirical noise and data marginals.

\paragraph{Implementation details.}
We initialize from the publicly released
MeanFlow~\cite{geng2026mean} checkpoint distributed as a training state
with the official PyTorch implementation (\texttt{py-meanflow}),
restoring the raw model parameters,
Adam optimizer state, and EMA weights.
We restore the random-number-generator
state so that target construction and
the prediction forward pass use consistent
dropout masks.
Activation checkpointing is applied
only to the prediction forward pass.
We apply gradient-aware selection once
every four training updates and use
Euclidean OT at the remaining updates.
The selection frequency and training
schedule are experimental settings,
rather than requirements of the
MeanFlow extension.

\section{Extension to Conditional Generation}
\label{sec:sup_conditional}

We extend GOT-FM to class-conditional image generation
using the SiT-B/4 architecture~\cite{ma2024sit}.
The FM objective and optimizer-aware coupling criterion
remain unchanged, but class conditioning is incorporated
throughout coupling construction and evaluation.
In particular, we restrict transport to samples
of the same class and evaluate candidate updates
using a shared class-conditional reference objective.

\paragraph{Class-Conditional FM Objective.}
Let $\mathbf n_i$ denote a noise sample,
$\mathbf z_j$ a data latent, and $c_j$ its class label.
We use the standard linear FM interpolation,
\begin{equation}
\begin{gathered}
\mathbf x_{ij}^t
=(1-t)\mathbf n_i+t\mathbf z_j,\\
\mathbf u_{ij}=\mathbf z_j-\mathbf n_i.
\end{gathered}
\end{equation}
The class-conditioned velocity model
$\mathbf v_{\boldsymbol\theta}
(\mathbf x_{ij}^t,t,c_j)$
predicts the target velocity.
The corresponding per-pair FM loss is
\begin{equation}
\ell_{ij}^{\mathrm{cond}}(\boldsymbol\theta)
=
\mathbb E_{t\sim\mathcal U[0,1]}
\Bigg[
\frac{1}{d}
\Big\|
\mathbf v_{\boldsymbol\theta}
(\mathbf x_{ij}^t,t,c_j)
-\mathbf u_{ij}
\Big\|_2^2
\Bigg].
\end{equation}
Thus, the underlying FM loss retains the same
form as in the unconditional setting, with
class conditioning added to the velocity predictor.

\paragraph{Class-Preserving Couplings.}
In class-conditional generation, unrestricted
transport may induce a dependence between
the source noise and the target class.
This can create a mismatch between the source
distributions used during training and inference.
We therefore assign class labels to independently
sampled Gaussian noise before constructing
class-preserving couplings.
Specifically, we sample noise independently of the class label,
with $\mathbf n_i\sim\mathcal N(\mathbf 0,\mathbf I)$,
and assign each noise sample a class $c_i$.
We restrict feasible couplings to
\begin{equation}
\begin{aligned}
\Pi_{\mathrm{cond}}(\mathbf a,\mathbf b)
&=\Big\{\boldsymbol\pi\in\Pi(\mathbf a,\mathbf b):\\
&\quad\pi_{ij}=0\quad\text{if }c_i\neq c_j\Big\}.
\end{aligned}
\end{equation}
Within each class block, the row and column
marginal constraints remain unchanged.
We apply these restrictions consistently
to the baseline Euclidean OT coupling,
all GOT-FM candidates, and the final
training coupling.
Consequently, coupling selection changes
source--target assignments only within
the same class.

To provide multiple feasible assignments
within each class, we use class-grouped
minibatches consisting of 16 class groups
with 16 samples per group, giving a total
batch size of 256.
Class groups are sampled according to
their empirical frequencies, while
corresponding noise samples are drawn
independently.

\paragraph{Conditional Gradient-Aware Selection.}
The optimizer-aware scoring procedure follows
the unconditional GOT-FM formulation.
For each admissible source--target pair,
we compute
\begin{equation}
\begin{gathered}
\mathbf g_{ij}^{\mathrm{cond}}
=\nabla_{\boldsymbol\theta}
\ell_{ij}^{\mathrm{cond}}(\boldsymbol\theta),\\
\mathbf g_{\boldsymbol\pi}^{\mathrm{cond}}
=\sum_{i,j}\pi_{ij}
\mathbf g_{ij}^{\mathrm{cond}}.
\end{gathered}
\end{equation}
Starting from the class-preserving Euclidean
OT baseline $\boldsymbol\pi_0$, we construct
the virtual optimizer update
\begin{equation}
\boldsymbol\theta_0^+
=
\mathcal A_s
(\boldsymbol\theta,
\mathbf g_{\boldsymbol\pi_0}^{\mathrm{cond}}).
\end{equation}
We then evaluate the gradient of a shared
conditional reference objective at
$\boldsymbol\theta_0^+$.
The resulting gradient-aware edge cost is
\begin{equation}
\begin{gathered}
C_{ij}^{\mathrm{GOT,cond}}
=(\mathbf g_{ij}^{\mathrm{cond}})^\top
\mathbf R_0^\top\mathbf g_{\mathrm{ref}}^{+,\mathrm{cond}},\\
c_i=c_j,
\end{gathered}
\end{equation}
where $\mathbf R_0$ denotes the optimizer
response at the baseline update.

Only within-class edges are scored, and
candidate selection is restricted to
class-preserving coupling plans.
$J_\text{ref}$ evaluation uses class-consistent
reference batches, so candidate updates
are compared using conditional FM losses.
We disable label dropout during training
and gradient-aware evaluation.

\paragraph{Latent-Space Implementation.}
For ImageNet~\cite{deng2009imagenet} $256\times256$ generation,
we encode images using a fixed VAE,
obtaining latents of size $4\times32\times32$.
All coupling-related computations are
performed in this latent space, including
the FM loss, Euclidean OT geometry,
endpoint prediction-error costs,
and gradient-aware $J_\text{ref}$ evaluation.

In particular, the geometric transport
cost is computed from squared Euclidean
distances between noise and data latents,
while endpoint-error proposals use the
class-conditioned velocity predictor.
The geometric and endpoint-error budgets
are calibrated to the corresponding
latent-space cost scales.

The class-preserving constraints and
conditional reference evaluation allow
GOT-FM to select learning-aware couplings
without modifying the SiT architecture
or introducing additional inference
operations.

\section{Implementation Details}
\label{sec:sup_implementation}
\subsection{Datasets, Architectures, and Initialization}
\label{sec:sup_impl_arch}

\noindent\textbf{CIFAR-10 Rectified Flow.}
We use the 50,000 training images of CIFAR-10~\cite{krizhevsky2009learning},
normalize pixel values to $[-1,1]$, and apply random horizontal flips.
The TorchCFM-style U-Net~\cite{dhariwal2021diffusion,tong2023improving} has 128 base channels, channel multipliers
$(1,2,2,2)$, four residual blocks per resolution level, attention at
resolution 16, and dropout probability 0.15, totaling 58,687,747
parameters. In the checkpoint-matched continuation protocol, we initialize
from an independent-coupling Rectified Flow~\cite{liu2022flow} checkpoint at training step
234,000. We restore the raw model, the saved Adam moments and step
counters, and the evaluation exponential moving average (EMA).
Thus, these comparisons evaluate additional training from a shared state,
rather than independent from-scratch runs.

\noindent\textbf{Class-conditional ImageNet.}
We use the ImageNet~\cite{deng2009imagenet} training split (1,281,167 images and 1,000 classes)
with $256\times256$ center crops and random horizontal flips.
A fixed SD-VAE encodes each image by sampling from its posterior and
multiplying the latent by 0.18125, yielding a $4\times32\times32$
representation. The class-conditioned backbone is SiT-B/4~\cite{ma2024sit}.
Continuation training starts from a publicly released ImageNet
checkpoint, restoring the raw model and provided optimizer state.
The VAE is kept frozen. Class-preserving coupling constraints and
the class-grouped sampler are detailed in
Appendix~\ref{sec:sup_conditional}.

\noindent\textbf{MeanFlow.}
For CIFAR-10 MeanFlow~\cite{geng2026mean}, we use a publicly released
MeanFlow-family SongUNet containing 55,932,419 parameters, of which
55,929,347 remain trainable when the unused augmentation mapping is
frozen. We restore the raw weights, Adam state, and EMA from the
MeanFlow checkpoint distributed with the official PyTorch implementation
(\texttt{py-meanflow}), as specified in Appendix~\ref{sec:sup_meanflow}.
All MeanFlow comparisons continue from this shared checkpoint.
The mean-velocity training objective and its
GOT-FM adaptation are described in Appendix~\ref{sec:sup_meanflow}.

\noindent\textbf{Two-dimensional experiments.}
We use a velocity MLP of
widths $3\!\rightarrow\!96\!\rightarrow\!96\!\rightarrow\!96\!\rightarrow\!2$,
with Swish activations and the flow time concatenated to the 2D input.
We compare coupling methods by continuing from a shared state after
500 epochs of Rectified Flow warm-up.

\subsection{Training Configurations}
\label{sec:sup_impl_training}

Table~\ref{tab:sup_training_configs} summarizes the continuation
training configurations. Unless noted otherwise, we restore the optimizer
state from the corresponding parent checkpoint rather than resetting its
moments or update counter. The MeanFlow and conditional ImageNet objectives
are not identical to the Rectified Flow training objective.

\begin{table*}[t]
\centering
\caption{\textbf{Continuation training configurations.}
The CIFAR-10 Rectified Flow settings refer to checkpoint-matched runs.
Reported optimizer states are resumed from the parent checkpoint.
The corresponding result must be identified by its run configuration.
The exact RF Adam betas are inherited from the selected parent checkpoint.}
\label{tab:sup_training_configs}
\footnotesize
\setlength{\tabcolsep}{7pt}
\renewcommand{\arraystretch}{1.12}
\begin{tabular}{@{}llll@{}}
\toprule
Setting & CIFAR-10 Rectified Flow~\cite{krizhevsky2009learning,liu2022flow} & CIFAR-10 MeanFlow~\cite{krizhevsky2009learning,geng2026mean} & Conditional ImageNet~\cite{deng2009imagenet} \\
\midrule
Training representation & RGB $3\times32\times32$ & RGB $3\times32\times32$ & Latent $4\times32\times32$ \\
Backbone & TorchCFM U-Net~\cite{dhariwal2021diffusion,tong2023improving} & SongUNet & SiT-B/4~\cite{ma2024sit} \\
Global batch size & 128 & 128 & 256 \\
Microbatch size & 128 (protocol-dependent) & 128 & 4 \\
Optimizer & Adam~\cite{kingma2014adam} (resumed) & Adam (resumed) & Adam (resumed) \\
Learning rate & $2\times10^{-4}$ & $10^{-4}$ & $10^{-4}$ \\
Learning-rate schedule & Constant & Constant & Constant \\
Adam $\beta_1,\beta_2$ & $(0.9,0.999)$ & $(0.9,0.999)$ & $(0.9,0.95)$ \\
Weight decay & 0 & 0 & 0 \\
Global gradient clip & 1 & 1 & 1 \\
Evaluation EMA decay & 0.999999 & 0.99999 & 0.9999 \\
EMA update period & Every update & Every 16 updates & Every update \\
GOT-FM selection period & Every 8 updates & Every 4 updates & Every 8 updates \\
Precision & FP32 & FP32 & FP32 \\
\bottomrule
\end{tabular}
\vspace{-1mm}
\end{table*}

\noindent\textbf{Continuation budgets.}
For CIFAR-10 with global batch size 128, one epoch-equivalent comprises
$\lfloor50{,}000/128\rfloor=390$ optimizer updates. Thus, the
140- and 150-epoch-equivalent continuations correspond to 54,600 and
58,500 additional updates, respectively. The MeanFlow experiment uses
54,600 additional updates. Conditional ImageNet uses 30 epoch-equivalents,
corresponding to $30\lfloor1{,}281{,}167/256\rfloor=150{,}120$
additional updates. Because ImageNet uses class-grouped sampling, an
epoch-equivalent specifies the number of sampled examples rather than
an exhaustive pass through the dataset. Matched-update and matched-GPU-hour
protocols are reported separately because the same update count does not imply
the same computational cost.

\subsection{GOT-FM Implementation}
\label{sec:sup_impl_got}

\noindent\textbf{Candidate construction.}
At each designated selection update, we construct a balanced geometric
minibatch OT plan $\boldsymbol\pi_0$ and generate alternative coupling
proposals that preserve both empirical marginals. The proposals are
filtered by the two-sided geometric-cost and endpoint-error
constraints in Eq.~\eqref{eq:app-budget-constraints}. Their $1\%$ budgets
use fixed median baseline-cost scales from three calibration
minibatches at initialization. We compute the baseline assignment using
SciPy's \texttt{linear\_sum\_assignment} and solve the proposal LPs
for their respective cost matrices using \texttt{linprog(method='highs')}. The solver
and candidate-validation tolerances are specified in
Appendix~\ref{sec:sup_coupling}.
All source--target edges are admissible in the CIFAR-10 settings,
whereas conditional ImageNet permits only within-class edges and
preserves each class block's empirical marginals. The configured
alternative-proposal counts are eight for Rectified Flow and MeanFlow and three
for ImageNet. The default eight-alternative configuration comprises four
geometry/endpoint-based proposals from the ordered library in
Appendix~\ref{sec:sup_coupling_1} and four random-cost proposals with
independent $\mathcal N(0,1)$ edge costs. The baseline
$\boldsymbol\pi_0$ is added to every final selection pool and is not
counted among the alternative proposals. The final pools therefore
contain at most nine plans for Rectified Flow and MeanFlow and four
for ImageNet.
These values are upper targets: duplicate or infeasible
proposals may reduce the number of distinct candidates actually scored.

Table~\ref{tab:sup_got_configs} lists the coupling-selection settings.
Here, ``samples per law'' refers to each of the IID and OT components
of the shared reference objective.

\begin{table*}[t]
\centering
\caption{\textbf{GOT-FM coupling-selection settings.}
Configured alternatives exclude the baseline, which is always included
in the final selection pool. Duplicate or infeasible proposals can
reduce the actual pool size below the stated maximum.
Samples/law counts reference pairs per IID or OT component in each
scoring replicate, separately from the OT matching size $B_{\mathrm{ref}}$.}
\label{tab:sup_got_configs}
\footnotesize
\setlength{\tabcolsep}{7pt}
\renewcommand{\arraystretch}{1.12}
\begin{tabular}{@{}llll@{}}
\toprule
Setting & CIFAR-10 Rectified Flow~\cite{krizhevsky2009learning,liu2022flow} & CIFAR-10 MeanFlow~\cite{krizhevsky2009learning,geng2026mean} & Conditional ImageNet~\cite{deng2009imagenet} \\
\midrule
Configured alternatives & 8 & 8 & 3 \\
Maximum pool size (with baseline) & 9 & 9 & 4 \\
Geometry budget & 1\% & 1\% & 1\% \\
Endpoint-error budget & 1\% & 1\% & 1\% \\
Selection bank: samples/law & 32 & 32 & 4 \\
Selection bank refresh & 32 updates & 32 updates & 32 updates \\
\bottomrule
\end{tabular}
\end{table*}

% Declare the baseline table early so it can appear at the top of the next page.
\begin{table*}[t]
\centering
\caption{\textbf{Baseline algorithms and implementation choices.}
The matched continuation setup is distinct from each method's
original training and evaluation protocol.}
\label{tab:sup_baseline_details}
\footnotesize
\setlength{\tabcolsep}{5pt}
\renewcommand{\arraystretch}{1.13}
\begin{tabular}{@{}p{0.12\textwidth}p{0.37\textwidth}p{0.43\textwidth}@{}}
\toprule
Method & Retained algorithmic elements & Adaptation or implementation detail \\
\midrule
Rectified Flow~\cite{liu2022flow} & Independent noise--data coupling and standard FM loss & Shared backbone, parent checkpoint, optimizer state, and continuation budget \\
OT-CFM~\cite{tong2023improving} & Balanced minibatch OT with squared Euclidean cost & Same RF training loss with geometric coupling substituted for independent coupling \\
LOOM-CFM~\cite{davtyan2025faster} & Cached Gaussian seeds per augmented image identity, Hungarian reassignment, and cache refresh & Six seeds per image in an in-memory table instead of a disk cache, with shared backbone and optimizer \\
AlignFlow~\cite{kong2026alignflow} & Global semi-discrete OT assignment, target-quota rebalancing, and bank refresh & Public SDOT dual, 1M sources and 100K augmented targets, and bank refresh every 7,812 updates \\
MAC-full~\cite{lin2026beyond} & Full endpoint-EMA cost matrix, entropic OT, and maximum-mass Hungarian extraction & Regularization 0.5, separate scoring EMA (0.999) and evaluation EMA, and native continuous-time evaluation \\
OAT-FM~\cite{yue2025oat} & Reduced coupling cost and two-segment velocity/acceleration training objective & Paper-based reimplementation with Hungarian matching, $\alpha=2/3$, and boundary-target EMA (0.9999) \\
MeanFlow~\cite{geng2026mean} & Native average-velocity target, joint time sampler, and adaptive weighting & MeanFlow-specific objective and checkpoint continuation, rather than an RF loss substitution \\
\bottomrule
\end{tabular}
\end{table*}

\noindent\textbf{Reference banks and flow-time evaluation.}
The training bank is the current training minibatch used to construct
$\boldsymbol\pi_0$, generate candidate couplings, and perform the actual
optimizer update. In the CIFAR-10 Rectified Flow configuration,
this bank contains 128 training examples.

The reference bank defines the finite reference objective used to
compute $\mathbf g_{\mathrm{ref}}^+$ and the shared optimizer-aware
direction $\mathbf h_{\mathrm{GOT}}$. These quantities determine
the linear edge costs used to rank candidate couplings.
For the same CIFAR-10 Rectified Flow configuration, it contains two
replicates, each comprising 32 IID and 32 OT reference samples.
Thus, the selection bank contains $2(32+32)=128$ reference pairs
across both laws and replicates. The samples/law count specifies
the scoring bank size, whereas $B_{\mathrm{ref}}$ in
Appendix~\ref{sec:sup_ablation_batch} specifies the minibatch size
used to construct an OT reference matching.
We refresh the bank inputs every 32 optimizer updates and hold them
fixed across candidates within each selection event.
The cached quantities are the bank inputs. Model-dependent reference
losses and $\mathbf h_{\mathrm{GOT}}$ are recomputed at each selection event.

We form the selection reference objective by equally weighting losses
under independent (IID) and geometric OT couplings. The conditional
ImageNet reference uses within-class IID and OT laws. For the Rectified
Flow and ImageNet scoring objectives, we use three-point Gauss--Legendre
quadrature with nodes $(0.1127017,0.5,0.8872983)$ and weights
$(5/18,8/18,5/18)$. The actual model update instead draws fresh
uniform flow times. MeanFlow uses its joint $(r,t)$ sampler, sharing
three interval samples per source across candidate scores. Within a
MeanFlow selection event, the stopped targets and adaptive weights are
fixed for pairwise-gradient and reference-gradient evaluation,
as described in Appendix~\ref{sec:sup_meanflow}.

\noindent\textbf{Optimizer response and edge scoring.}
We read the current model parameters, Adam moments, and update counter
without modifying the live training state. A functional virtual update
under $\boldsymbol\pi_0$ provides the baseline updated parameters.
We evaluate the shared reference gradient at this virtual update and form
$\mathbf h_{\mathrm{GOT}}=\mathbf R_0^\top\mathbf g_{\mathrm{ref}}^+$,
where $\mathbf R_0$ is obtained by differentiating through the
implemented Adam update, including global gradient clipping where active.
We obtain pairwise costs from directional derivatives
$C_{ij}^{\mathrm{GOT}}=
D_{\boldsymbol\theta}\ell_{ij}[\mathbf h_{\mathrm{GOT}}]$,
without materializing a full parameter-gradient vector for each scored
edge. All candidates are scored at the same current model and optimizer
state. The baseline virtual update changes neither the live model
nor its optimizer moments or counters. The candidate with the lowest
linearized cost in the baseline-inclusive pool $\mathcal P^+$ is used
directly for the model update at each designated selection step, with
ties resolved in favor of $\boldsymbol\pi_0$. The weighted support-edge training update
is specified in Appendix~\ref{sec:sup_coupling_3}. Only this model
update commits a state change.

\subsection{Baseline Implementations}
\label{sec:sup_impl_baselines}

Our baseline comparisons adapt the respective algorithms to the
reported shared architectures and continuation protocols. We do not
claim to reproduce every original from-scratch training recipe.
Table~\ref{tab:sup_baseline_details} distinguishes the retained
algorithmic components from implementation-specific changes.

\noindent\textbf{OT-CFM \cite{tong2023improving}.}
At each update, we draw 128 independent Gaussian sources and 128
augmented training images, construct the full $128\times128$
pixel-space squared Euclidean cost matrix in float64, and solve
the exact minibatch assignment using the Hungarian algorithm.
Costs are divided by the image dimensionality, which does not change
the optimal assignment. We use all 128 assigned pairs exactly once
with straight interpolation ($\sigma=0$), independently sampled
uniform times, and the standard velocity-regression loss.
Unlike the public TorchCFM sampler, which samples pairs with
replacement from the OT plan, this implementation uses the complete
assignment. The two procedures have the same expected gradient
under the plan but differ in sampling variance and individual updates.

\noindent\textbf{LOOM-CFM \cite{davtyan2025faster}.}
We adapt the official implementation, retaining six cached Gaussian
seeds per augmented image identity and the original Euclidean-distance
Hungarian reassignment. At each update, one seed is selected for each
image in the minibatch, the corresponding noise samples are reassigned,
and the selected cache entries are updated with the matched seeds.
Original and horizontally flipped images have separate identities,
and we retain the 1000-epoch seed-resampling schedule.
An in-memory seed table replaces per-image disk writes and is
checkpointed together with the optimizer and random states.
The architecture, batch size, interpolation noise, optimizer schedule,
and EMA follow the shared continuation protocol.

\noindent\textbf{OAT-FM \cite{yue2025oat}.}
We use a paper-based reimplementation because runnable official
repository code was not available in the examined version. The
implementation follows the reduced coupling objective reported as
Eq.~(39) in the original paper. We do not assert equivalence to an
independently implemented full objective. Since OAT-FM modifies the
training objective as well as coupling construction, the
comparison should not be described as using an identical loss for
all baselines.

\noindent\textbf{AlignFlow \cite{kong2026alignflow}.}
We adapt the official semi-discrete OT coupling to the shared
RF backbone using the publicly released CIFAR-10 SDOT dual.
The target set contains all 50,000 training images and their
horizontal flips in the ordering expected by the dual.
Each bank contains one million independent Gaussian sources,
assigned by minimizing the pixel-space squared Euclidean cost
minus the target dual weight. The official quota-rebalancing
procedure then gives each augmented target exactly ten sources.
We shuffle the resulting pairs and refresh the bank every
7,812 updates. A stable-sort implementation replaces the original
rebalancing scans and is tested for tensor-exact agreement with
the upstream procedure. The model uses the shared
velocity-regression objective. Bank construction and refreshes
are charged to the continuation budget, while training of the released
SDOT dual is excluded, as specified below.

\noindent\textbf{MAC-full \cite{lin2026beyond}.}
The implementation evaluates all $128\times128$ minibatch edges, applies
$z$-score normalization to their costs, and solves entropic OT with
regularization 0.5. A maximum-mass Hungarian assignment is then
extracted from the resulting soft plan. We align target endpoint
predictions with the target axis and use pairwise matrix products and
float64 log-domain Sinkhorn. The scoring EMA (0.999) is maintained
separately from the evaluation EMA.

\subsection{Evaluation Protocol}
\label{sec:sup_impl_eval}
For the matched-compute CIFAR-10 Rectified Flow comparisons in
Table~\ref{tab:gpu_hour_matched}, we evaluate the budget-end checkpoints
using EMA weights and generate 50,000 samples per method with
the Euler solver at 1, 4, 8, 16, and 50 NFEs.
All methods in this comparison use the same fixed Gaussian
input tensors, image quantization procedure, Inception weights,
and FID implementation in the same PRO6000 evaluation environment.
FID~\cite{heusel2017gans} is computed against fixed CIFAR-10~\cite{krizhevsky2009learning}
test-set reference statistics. Generation starts from independent standard Gaussian
noise without coupling, rematching, guidance, or sample selection
at inference time. Evaluation is excluded from the attributed
training budget and does not determine training settings or
checkpoint selection. We verify checkpoint hashes, generated sample
counts, and the shared input-noise hash before aggregating results.

The main comparisons use the following model-specific evaluation protocols.
For Table~\ref{tab:ot_main_cifar}, both Rectified Flow and MeanFlow
use EMA checkpoints and CIFAR-10 test-set FID reference statistics.
Rectified Flow uses Euler integration, whereas MeanFlow uses its
native average-velocity sampler with one function evaluation.
Tables~\ref{tab:ot_main_cifar} and~\ref{tab:gpu_hour_matched}
both evaluate EMA checkpoints against the same CIFAR-10 test-set
FID reference statistics.

For the class-conditional ImageNet results in
Table~\ref{tab:imagenet_cond}, we evaluate EMA checkpoints.
For a sampling budget of $N$ NFEs, we use fixed-step explicit Euler
with $t_i=i/N$, $i=0,\ldots,N-1$, and $\Delta t=1/N$,
integrating from noise at $t=0$ to data at $t=1$ with one
velocity-field evaluation per step. The classifier-free guidance
(CFG) scale is 1.0, using class conditioning without additional
guidance. For each model and NFE budget, we generate 50,000 samples.
Class labels are drawn independently and uniformly from the
1,000 classes, so the number of generated samples per class is not fixed.
Latent samples are decoded with the fixed VAE described in
Appendix~\ref{sec:sup_conditional}.
FID is computed using 2,048-dimensional Inception features against
reference statistics from all 50,000 ImageNet validation images,
with ADM center-crop preprocessing at $256\times256$ resolution.

\begin{table}[t]
    \centering
    \caption{\textbf{Training costs of GOT-FM on CIFAR-10.}
    We report the total training time and peak allocated CUDA
    memory for 58,500 continuation updates on a
    single NVIDIA RTX PRO 6000 GPU.
    Evaluation costs are excluded.}
    \label{tab:got_overhead}
    \vspace{-0.5em}
    \footnotesize
    \setlength{\tabcolsep}{6pt}
    \renewcommand{\arraystretch}{1.15}
    \begin{tabular}{@{}lr@{}}
        \toprule
        Metric & Value \\
        \midrule
        GPU & NVIDIA RTX PRO 6000 \\
        Number of GPUs & 1 \\
        Training updates & 58,500 \\
        Total GPU-hours & 16.965 \\
        Wall-clock time (h) & 16.965 \\
        Peak CUDA allocated memory (GB) & 15.692 \\
        \bottomrule
    \end{tabular}
    \vspace{-0.5em}
\end{table}

\section{Computational Costs}
\label{sec:sup_computational}

\subsection{Training Costs}
\label{sec:sup_computational_2}
Table~\ref{tab:got_overhead} reports the training cost of GOT-FM over
58,500 continuation updates on a single NVIDIA RTX PRO 6000 GPU.
The update count, 16.965 GPU-hours, and 16.965-hour wall-clock time
were recorded in the same run.
Peak memory is recorded with
\texttt{torch.cuda.max\_memory\_allocated()}.

\subsection{Comparison under Matched Training Compute}
\label{sec:sup_computational_1}
GOT-FM maintains competitive generation quality under matched
training compute, including the additional overhead of candidate
construction, optimizer-aware scoring, and coupling selection.
We compare methods under the same additional GPU-hour budget
to assess generation quality after accounting for this overhead.

All methods are evaluated using the CIFAR-10
continuation protocol described in
Appendix~\ref{sec:sup_implementation}.
The measured training cost includes the
method-specific computations required for
each update, such as transport matching,
candidate scoring, and reference evaluation.
Each method is trained within the same
additional GPU-hour budget, although the
number of completed optimizer updates may differ.

Table~\ref{tab:gpu_hour_matched} shows that GOT-FM achieves the
lowest FID at 4, 8, 16, and 50 NFEs under a shared budget of
8.613 GPU-hours. GOT-FM completes 27,776 optimizer updates,
compared with 58,500--118,848 updates for the baselines.

This comparison evaluates generation quality
under a shared continuation-training compute
budget, rather than an equal number of updates.
The reported GPU-hours exclude shared pretraining
and the cost of training the publicly available
AlignFlow dual model.
Therefore, the results should be interpreted
as compute-matched continuation comparisons,
not as end-to-end training-cost comparisons.

\begin{figure*}[!t]
    \centering

    \begin{subfigure}[b]{0.32\textwidth}
        \centering
        \includegraphics[width=\linewidth]
        {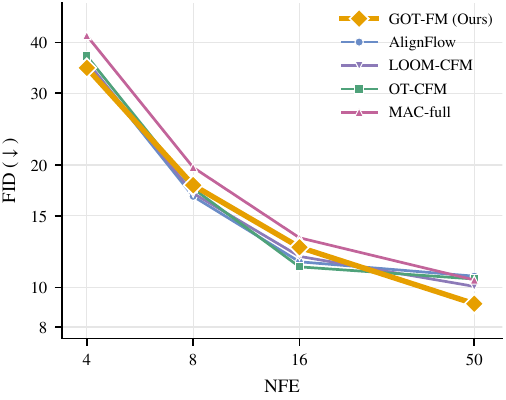}
        \caption{2.5 GPU-hours}
        \label{fig:compute_2_5h}
    \end{subfigure}
    \hfill
    \begin{subfigure}[b]{0.32\textwidth}
        \centering
        \includegraphics[width=\linewidth]
        {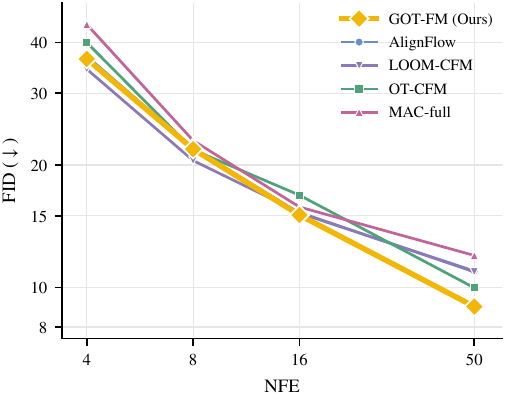}
        \caption{4 GPU-hours}
        \label{fig:compute_4h}
    \end{subfigure}
    \hfill
    \begin{subfigure}[b]{0.32\textwidth}
        \centering
        \includegraphics[width=\linewidth]
        {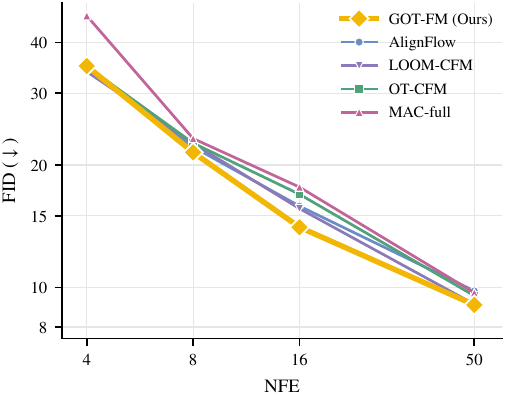}
        \caption{6 GPU-hours}
        \label{fig:compute_6h}
    \end{subfigure}

    \vspace{-1mm}
    \caption{\textbf{Generation quality under matched
    training-compute budgets on CIFAR-10~\cite{krizhevsky2009learning}.}
    We compare FID~\cite{heusel2017gans} across different numbers of function
    evaluations (NFEs) after training each method for
    (a) 2.5, (b) 4, and (c) 6 GPU-hours.
    All comparisons include method-specific training
    overhead under the same additional compute budget.
    Lower FID is better.}
    \label{fig:compute_matched}
    \vspace{-2mm}
\end{figure*}

Figure~\ref{fig:compute_matched} extends this comparison to
2.5, 4, and 6 GPU-hours, with method-specific training overhead
included in each budget. GOT-FM achieves the lowest FID at
50 NFEs in all three cases.
Together, these results show that GOT-FM retains competitive
generation quality under matched continuation-training compute,
even after accounting for its coupling-selection overhead.

\begin{table*}[!t]
    \centering
    \caption{\textbf{Generation quality under matched
    training compute on CIFAR-10~\cite{krizhevsky2009learning}.}
    All methods are allocated the same additional
    GPU-hour budget, including method-specific
    training overhead.
    We use EMA checkpoints and CIFAR-10 test-set FID statistics.
    We report the number of completed optimizer
    updates and FID~\cite{heusel2017gans} at different numbers of function
    evaluations (NFEs). Lower FID is better.}
    \label{tab:gpu_hour_matched}
    \vspace{-0.5em}
    \footnotesize
    \setlength{\tabcolsep}{7pt}
    \renewcommand{\arraystretch}{1.12}
    \begin{tabular}{@{}lccccccc@{}}
        \toprule
        Method
        & Updates
        & GPU-hours
        & FID$_1$($\downarrow$)
        & FID$_4$($\downarrow$)
        & FID$_8$($\downarrow$)
        & FID$_{16}$($\downarrow$)
        & FID$_{50}$($\downarrow$) \\
        \midrule
        Rectified Flow~\cite{liu2022flow}
        & 120,450 & 8.6130 & 353.31 & 58.23 & 23.67 & \underline{15.36} & 11.49 \\
        OT-CFM~\cite{tong2023improving}
        & 118,848 & 8.6130 & 245.79 & \underline{37.17} & 22.29 & 15.89 & 11.03 \\
        LOOM-CFM~\cite{davtyan2025faster}
        & 116,116 & 8.6129 & \textbf{226.22} & 38.16 & 21.97 & 15.45 & \underline{10.65} \\
        OAT-FM~\cite{yue2025oat}
        & 78,824 & 8.6129 & 314.07 & 188.85 & 172.32 & 168.37 & 167.63 \\
        AlignFlow~\cite{kong2026alignflow}
        & 113,789 & 8.6129 & \underline{229.31} & 37.64 & \underline{21.52} & 16.21 & 11.20 \\
        MAC-full~\cite{lin2026beyond}
        & 77,485 &  8.6130 & 318.12 & 53.82 & 27.21 & 18.23 & 12.68 \\
        \midrule
        % Match the highlight overhang to the table's zero outer padding.
        \rowcolor{purple!8}
        \multicolumn{1}{@{}>{\columncolor{purple!8}[0pt][\tabcolsep]}l}{\textbf{GOT-FM (Ours)}}
        & 27,776 & 8.6130 & 232.20 & \textbf{35.49} & \textbf{20.60} & \textbf{14.65}
        & \multicolumn{1}{>{\columncolor{purple!8}[\tabcolsep][0pt]}c@{}}{\textbf{9.51}} \\
        \bottomrule
    \end{tabular}
\end{table*}

\section{Additional Qualitative Results}
\label{sec:sup_qualitative}

\begin{figure*}[t]
    \centering

    \begin{subfigure}[b]{0.32\textwidth}
        \centering
        \includegraphics[width=\linewidth]{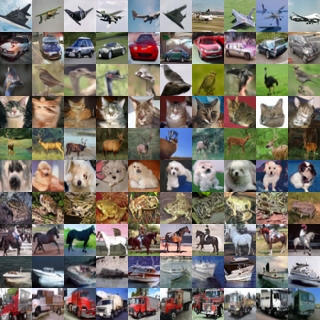}
        \caption{Rectified Flow~\cite{liu2022flow}.}
        \label{fig:qual_a}
    \end{subfigure}
    \hfill
    \begin{subfigure}[b]{0.32\textwidth}
        \centering
        \includegraphics[width=\linewidth]{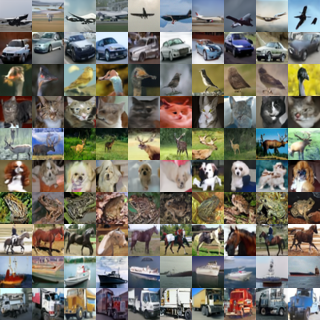}
        \caption{OT-CFM~\cite{tong2023improving}.}
        \label{fig:qual_b}
    \end{subfigure}
    \hfill
    \begin{subfigure}[b]{0.32\textwidth}
        \centering
        \includegraphics[width=\linewidth]{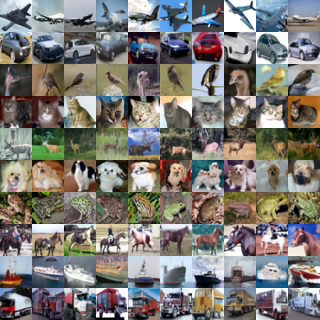}
        \caption{GOT-FM (Ours).}
        \label{fig:qual_c}
    \end{subfigure}

    \vspace{-1mm}
    \caption{\textbf{Additional qualitative results on CIFAR-10~\cite{krizhevsky2009learning}.}
    Samples from Rectified Flow, OT-CFM, and GOT-FM at 50 function
    evaluations (NFE=50).}
    \label{fig:qualitative_cifar}
    \vspace{-2mm}
\end{figure*}

\begin{figure*}[t]
    \centering

    \begin{subfigure}[b]{0.32\textwidth}
        \centering
        \includegraphics[width=\linewidth]{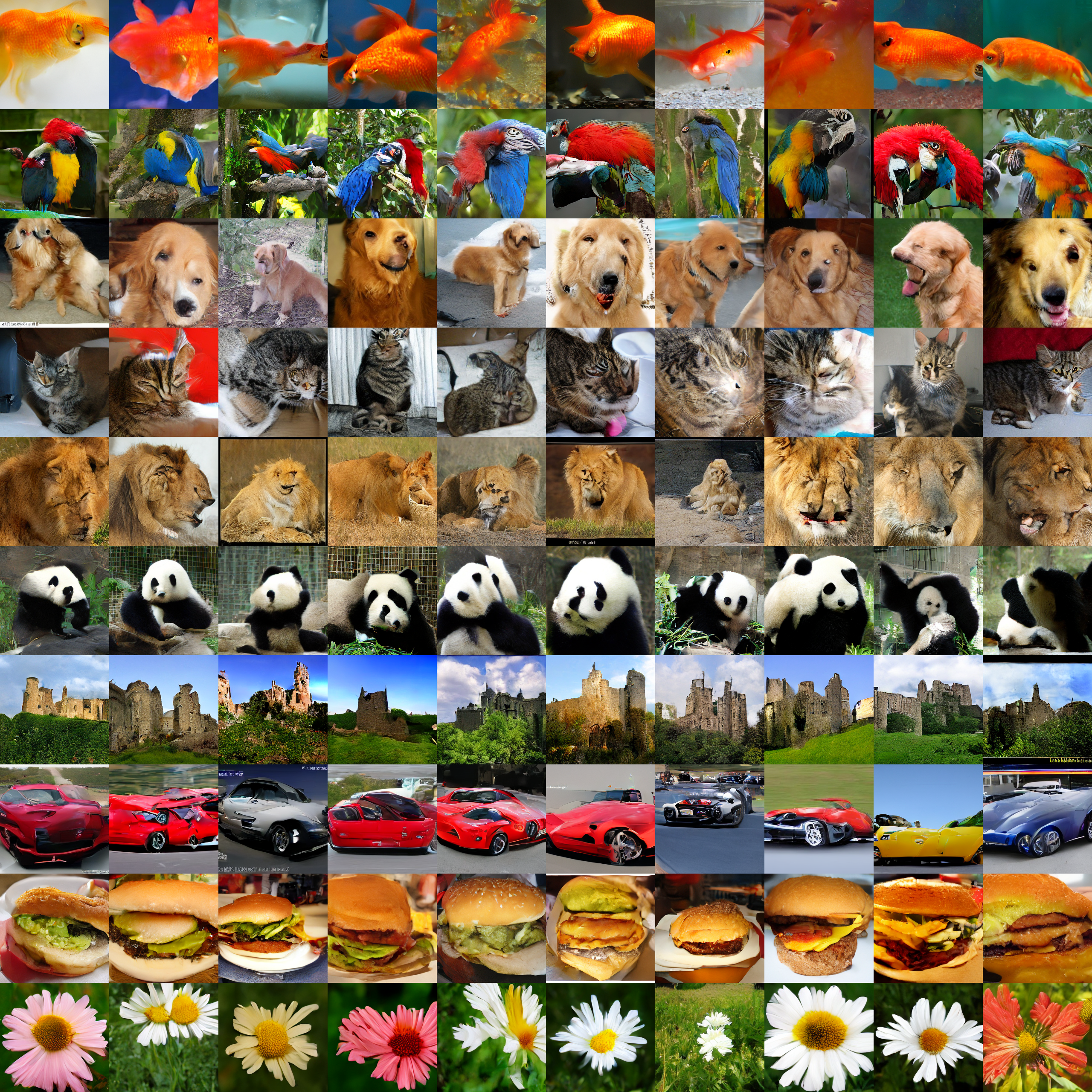}
        \caption{Rectified Flow~\cite{liu2022flow}.}
        \label{fig:qual_imagenet_a}
    \end{subfigure}
    \hfill
    \begin{subfigure}[b]{0.32\textwidth}
        \centering
        \includegraphics[width=\linewidth]{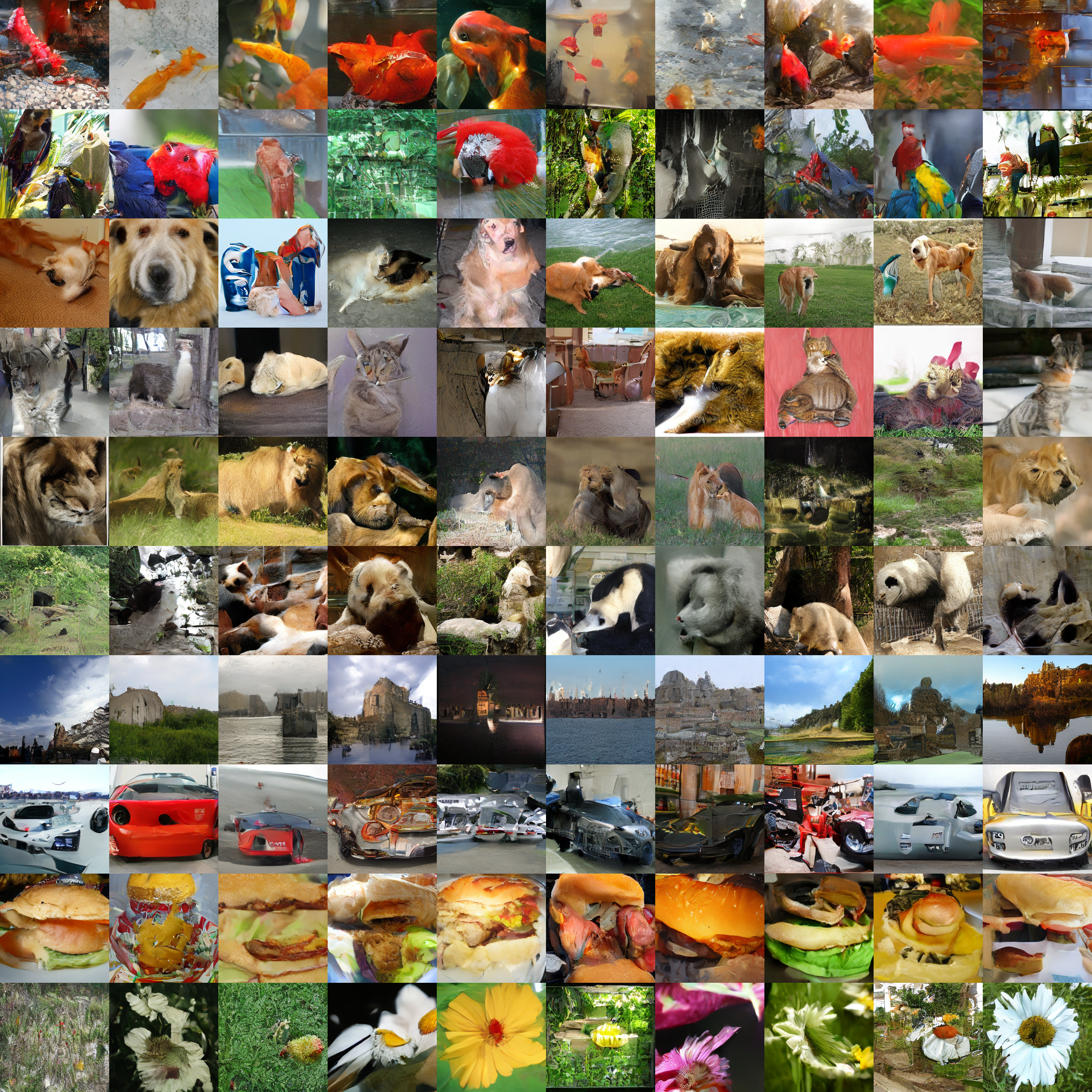}
        \caption{OT-CFM~\cite{tong2023improving}.}
        \label{fig:qual_imagenet_b}
    \end{subfigure}
    \hfill
    \begin{subfigure}[b]{0.32\textwidth}
        \centering
        \includegraphics[width=\linewidth]{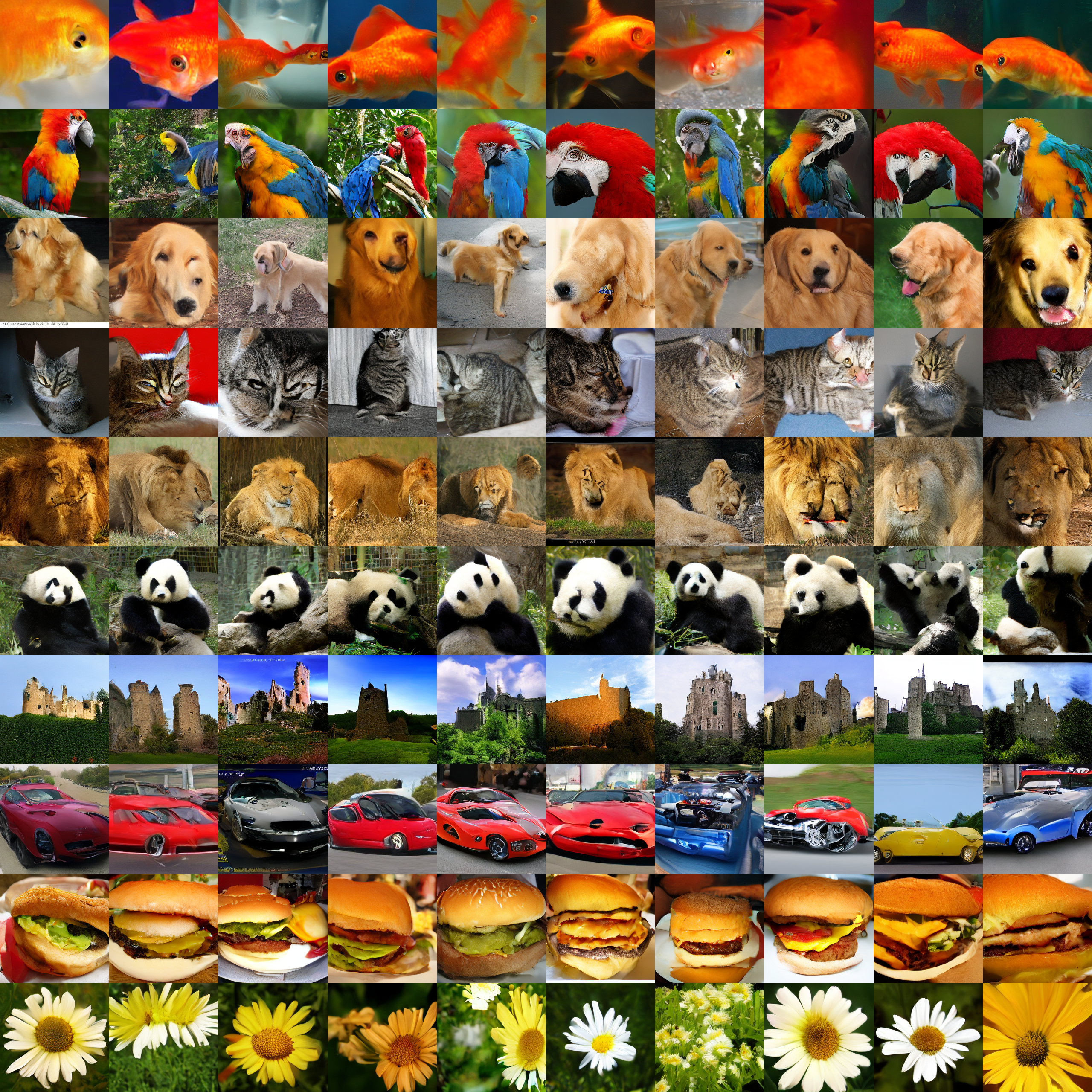}
        \caption{GOT-FM (Ours).}
        \label{fig:qual_imagenet_c}
    \end{subfigure}

    \vspace{-1mm}
    \caption{\textbf{Additional qualitative results on ImageNet~\cite{deng2009imagenet}} $\mathbf{256\times256}$.
    Samples from Rectified Flow, OT-CFM, and GOT-FM at 50 function
    evaluations (NFE=50).}
    \label{fig:qualitative_imagenet}
    \vspace{-2mm}
\end{figure*}

Figures~\ref{fig:qualitative_cifar} and \ref{fig:qualitative_imagenet} present additional generated samples to complement the quantitative FID results. We use the same model checkpoints and sampling configurations as in the corresponding experiments, providing a visual comparison of generation quality across methods.

\section{Discussion}
\label{sec:sup_discussion}

\subsection{Implications and Contributions}
\label{sec:sup_discussion_1}
Our work establishes a learning-oriented perspective on optimal transport for Flow Matching. While existing gradient-based data selection and reweighting methods primarily evaluate the utility of predefined training examples, coupling selection involves a structurally different decision problem: changing the joint assignment of source and target samples while preserving both endpoint marginals.

Under fixed-marginal constraints, the relevant learning signal is not determined by individual pairwise gradients alone, but by gradient contrasts between alternative feasible assignments. Our analysis shows that source-only and target-only gradient components cancel in coupling comparisons, identifying the coupling-dependent component of the training gradient that can influence learning through reassignment.

More importantly, we characterize when gradient-aware coupling provides additional first-order learning information beyond geometric transport costs and prediction errors. Proposition 1 identifies this information through the residual gradient-aware cost projected onto marginal- and cost-preserving perturbations. When this residual is nonzero, couplings that are indistinguishable under existing criteria can exhibit different local learning effects, provided the corresponding perturbations are feasible.

This characterization distinguishes GOT-FM from a direct application of gradient-based data selection to Flow Matching. Rather than merely assigning learning-utility scores to training examples, GOT-FM exploits the structure of feasible transport contrasts to identify learning-relevant changes in source–target assignments.

Our theoretical counterexamples and empirical analysis of cost-matched couplings support this distinction. GOT-FM then translates this additional learning signal into an optimizer-conditioned transport cost that enables efficient coupling selection without explicitly storing per-pair parameter gradients.

The resulting framework extends the role of optimal transport in Flow Matching from constructing geometrically favorable training paths to identifying assignments that are beneficial for model optimization. Its applicability to Rectified Flow, MeanFlow, and class-conditional generation further demonstrates the flexibility of this perspective without requiring architectural modifications or additional inference operations.

\subsection{Limitations}
\label{sec:sup_discussion_2}
GOT-FM relies on a first-order approximation of
the post-update reference loss around a baseline
coupling. Although the resulting cost captures
local optimizer-conditioned sensitivity, it does
not guarantee improvement for finite coupling
changes or over multiple training updates.
Furthermore, selecting from a finite candidate
pool limits the available coupling choices,
and constructing and scoring these candidates
introduces additional training overhead.

The effectiveness of GOT-FM also depends on
the reference objective used to evaluate learning
utility. As shown in our ablations, different
reference objectives can lead to different
coupling decisions and generation performance.
Therefore, the proposed criterion does not
guarantee uniformly better performance across
reference choices or training settings.

\subsection{Future Directions}
\label{sec:sup_discussion_3}
An important direction is to develop gradient-aware
coupling criteria that account for multiple optimizer
updates rather than a single local update.
Another direction is to improve computational
efficiency through adaptive candidate generation
and selection schedules.
Finally, exploring more robust reference objectives
and evaluating the approach across broader model
architectures and training budgets could further
clarify when optimizer-aware coupling selection
is most effective.

% Place the qualitative figures on the page after the quantitative results.
\clearpage

\end{document}